\documentclass[11pt,a4paper]{article}
\usepackage[utf8]{inputenc}

\usepackage{mathrsfs,amsthm,xcolor,verbatim,bbm,amsmath,amsfonts,amssymb,nicefrac,hyperref,bm,mathtools,xparse,etoolbox}
\usepackage[inline,shortlabels]{enumitem}
\usepackage[capitalise,sort]{cleveref} %

\crefname{enumi}{item}{items}
\usepackage[margin=1in]{geometry}
\usepackage[capitalise]{cleveref} 

\usepackage{stmaryrd} 

\crefname{enumi}{item}{items}
\crefname{equation}{}{}
\crefname{subsection}{Subsection}{Subsections}
\usepackage{xurl}

\hypersetup{
	colorlinks,
	linkcolor={red!80!black},
	citecolor={green},
	urlcolor={blue!80!black},
}

\theoremstyle{plain}
\newtheorem{theorem}{Theorem}[section]

\newtheorem{prop}[theorem]{Proposition}

\newtheorem{setting}[theorem]{Setting}

\theoremstyle{definition}
\newtheorem{definition}[theorem]{Definition}

\DeclareMathAlphabet{\mathpzc}{OT1}{pzc}{m}{it}

\DeclareFontEncoding{LS1}{}{}
\DeclareFontSubstitution{LS1}{stix}{m}{n}
\DeclareMathAlphabet{\mathscr}{LS1}{stixscr}{m}{n}

\newcommand{\E}{\mathbb{E}}
\renewcommand{\P}{\mathbb{P}}

\newcommand{\R}{\mathbb{R}}
\newcommand{\N}{\mathbb{N}}
\newcommand{\Z}{\mathbb{Z}}
\newcommand{\scre}{\mathscr{e}}

\newcommand{\bbL}{\mathbb{L}}

\newcommand{\Dens}{\mathscr{p}}

\newcommand{\inact}{\cI}

\renewcommand{\d}{ \mathrm{d}}

\renewcommand{\c}[1]{\mathfrak{c}^{#1}}

\newcommand{\ssum}{\textstyle\sum}

\newcommand{\cLnri}[2]{\mathcal{L}_{#1}^{#2}}
\newcommand{\g}{\mathscr{g}}

\newcommand{\bftheta}{\eta}
\newcommand{\bfvartheta}{\zeta}

\newcommand{\free}{W}

\newcommand{\cB}{\mathcal{B}}

\newcommand{\cF}{\mathcal{F}}
\newcommand{\cG}{\mathcal{G}}

\newcommand{\cI}{\mathcal{I}}
\newcommand{\cJ}{\mathcal{J}}

\newcommand{\cL}{\mathcal{L}}

\newcommand{\cP}{\mathcal{P}}

\newcommand{\cX}{\mathcal{X}}
\newcommand{\cY}{\mathcal{Y}}

\newcommand{\set}{B}

\newcommand{\bfc}{\mathbf{c}}
\newcommand{\bfd}{\mathbf{d}}
\newcommand{\bfe}{\mathbf{e}}

\newcommand{\bfw}{\mathbf{w}}

\newcommand{\bfy}{\mathbf{y}}
\newcommand{\bfz}{\mathbf{z}}
\newcommand{\scrv}{\mathscr{v}}

\newcommand{\bbA}{\mathbb{A}}
\newcommand{\bfA}{\mathbf{A}}

\newcommand{\bfI}{\mathbf{I}}

\newcommand{\bfN}{\mathbf{N}}

\newcommand{\bfX}{\mathbf{X}}

\newcommand{\scrM}{\mathscr{M}}
\newcommand{\scrN}{\mathscr{N}}

\newcommand{\fL}{\mathfrak{L}}

\newcommand{\fU}{\mathfrak{U}}

\newcommand{\fb}{\mathfrak{b}}

\newcommand{\fd}{\mathfrak{d}}

\newcommand{\fp}{\mathfrak{p}}

\newcommand{\fw}{\mathfrak{w}}

\newcommand{\cki}{c}

\def\mN{\mathcal N}

\newcommand{\InActL}[1]{\mathfrak{I}_{#1}}
\newcommand{\NNel}{\theta}

\newcommand{\NNelll}{\Theta}

\newcommand{\W}{w}

\newcommand{\grad}{\mathscr{a}}
\newcommand{\alphaw}{\mathscr{v}}
\newcommand{\betaw}{\mathscr{w}}

\DeclarePairedDelimiter{\abs}{\lvert}{\rvert}

\DeclarePairedDelimiter{\spro}{\langle}{\rangle}

\newcommand{\qandq}{\quad \text{and} \quad }
\newcommand{\qqandqq}{\qquad\text{and}\qquad}

\newcommand{\indicator}[1]{\mathbbm{1}_{\smash{#1}}}

\newcommand{\xTheta}[2]{\Theta_{#1}^{#2}}
\newcommand{\xxell}[2]{\ell_{#2}^{#1}}
\newcommand{\xell}[1]{\ell_{#1}}

\newcommand{\muu}[2]{\mu(#1,#2)}
\newcommand{\muuu}[3]{\mu(#1,#2,#3)}

\usepackage{cleveref}

\ExplSyntaxOn

\seq_new:N \g_cflist_loaded
\seq_new:N \g_cflist_pending

\NewDocumentCommand{\cfadd} { m } {
  \seq_if_in:NnF \g_cflist_loaded { #1 } {
    \seq_if_in:NnF \g_cflist_pending { #1 } {
      \seq_gput_right:Nn \g_cflist_pending { #1 }
    }
  }
}

\NewDocumentCommand{\cfconsiderloaded} { m } {
  \seq_gput_right:Nn \g_cflist_loaded {#1}
}

\NewDocumentCommand{\cfremove} { m } {
  \seq_gremove_all:Nn \g_cflist_pending { #1 }
}

\NewDocumentCommand{\cfload} { o } {
  \seq_if_empty:NTF \g_cflist_pending {
    \IfValueTF{#1}{\ignorespaces}{\unskip}
  } {
    (cf.\ \cref{\seq_use:Nn \g_cflist_pending {,}})\IfValueTF{#1}{#1~}{\unskip}
    \seq_gconcat:NNN \g_cflist_loaded \g_cflist_loaded \g_cflist_pending
    \seq_gclear:N \g_cflist_pending
    \IfValueT{#1}{\ignorespaces}
  }
}

\NewDocumentCommand{\cfclear} {} {
  \seq_gclear:N \g_cflist_loaded
  \seq_gclear:N \g_cflist_pending
}

\NewDocumentCommand{\cfout} { o } {
  \seq_if_empty:NTF \g_cflist_pending {\unskip\IfValueT{#1}{\ignorespaces}} {
    (cf.\ \cref{\seq_use:Nn \g_cflist_pending {,}})\IfValueTF{#1}{#1~}{\unskip}
    \seq_gclear:N \g_cflist_pending
    \IfValueT{#1}{\ignorespaces}
  }
}

\NewDocumentCommand{\ifnocf} { m } {
  \seq_if_empty:NT \g_cflist_pending { #1 }
}

\ExplSyntaxOff

\ExplSyntaxOn

\bool_new:N \g_noteobserve

\NewDocumentCommand{\setnote}{}{
  \bool_gset_true:N \g_noteobserve
}

\NewDocumentCommand{\setobserve}{}{
  \bool_gset_false:N \g_noteobserve
}

\NewDocumentCommand{\nobs}{ o }{
  \IfValueT{#1}{
    \str_if_eq:noTF {note} {#1} {
      \bool_gset_true:N \g_noteobserve
    } {
      \str_if_eq:noTF {Note} {#1} {
        \bool_gset_true:N \g_noteobserve
      } {
        \bool_gset_false:N \g_noteobserve
      }
    }
  }
  \bool_if:nTF { \g_noteobserve } {
    \bool_gset_false:N \g_noteobserve
    note
  } {
    \bool_gset_true:N \g_noteobserve
    observe
  }
  \IfValueF{#1}{~}
}

\NewDocumentCommand{\Nobs}{ o }{
  \IfValueT{#1}{
    \str_if_eq:noTF {note} {#1} {
      \bool_gset_true:N \g_noteobserve
    } {
      \str_if_eq:noTF {Note} {#1} {
        \bool_gset_true:N \g_noteobserve
      } {
        \bool_gset_false:N \g_noteobserve
      }
    }
  }
  \bool_if:nTF { \g_noteobserve } {
    \bool_gset_false:N \g_noteobserve
    Note
  } {
    \bool_gset_true:N \g_noteobserve
    Observe
  }
  \IfValueF{#1}{~}
}

\ExplSyntaxOff

\ExplSyntaxOn

\bool_new:N \g_hencetherefore

\NewDocumentCommand{\hence}{ o }{
  \IfValueT{#1}{
    \str_if_eq:noTF {hence} {#1} {
      \bool_gset_true:N \g_hencetherefore
    } {
      \str_if_eq:noTF {Hence} {#1} {
        \bool_gset_true:N \g_hencetherefore
      } {
        \bool_gset_false:N \g_hencetherefore
      }
    }
  }
  \bool_if:nTF { \g_hencetherefore } {
    \bool_gset_false:N \g_hencetherefore
    hence
  } {
    \bool_gset_true:N \g_hencetherefore
    therefore
  }
  \IfValueF{#1}{~}
}

\NewDocumentCommand{\Hence}{ o }{
  \IfValueT{#1}{
    \str_if_eq:noTF {hence} {#1} {
      \bool_gset_true:N \g_hencetherefore
    } {
      \str_if_eq:noTF {Hence} {#1} {
        \bool_gset_true:N \g_hencetherefore
      } {
        \bool_gset_false:N \g_hencetherefore
      }
    }
  }
  \bool_if:nTF { \g_hencetherefore } {
    \bool_gset_false:N \g_hencetherefore
    Hence,~we~obtain
  } {
    \bool_gset_true:N \g_hencetherefore
    Therefore,~we~obtain
  }
  \IfValueF{#1}{~}
}

\ExplSyntaxOff

\ExplSyntaxOn

\seq_const_from_clist:Nn \g_prove_mru {
  establish,
  demonstrate,
  prove,
  show,
  imply,
  ensure
}

\prop_new:N \l__verbs
\prop_put:Nnn \l__verbs {show} {shows}
\prop_put:Nnn \l__verbs {imply} {implies}
\prop_put:Nnn \l__verbs {demonstrate} {demonstrates}
\prop_put:Nnn \l__verbs {prove} {proves}
\prop_put:Nnn \l__verbs {establish} {establishes}
\prop_put:Nnn \l__verbs {ensure} {ensures}
\prop_put:Nnn \l__verbs {assure} {assures}

\tl_new:N \g_wordtmp
\seq_new:N \l_mytmps

\cs_generate_variant:Nn \str_if_in:nnTF { nVTF }
\cs_generate_variant:Nn \str_if_in:nnTF { xVTF }

\NewDocumentCommand{\prove}{ o }{
  \IfValueTF{#1}{
    \seq_clear:N \l_mytmps
    \seq_map_inline:Nn \g_prove_mru {
      \str_if_eq:nnTF {##1} {ensure} {
        \str_set:Nn \l_temps {n}
      } {
        \str_set:Nx \l_temps {\str_head_ignore_spaces:n {##1}}
      }
      \str_if_in:xVTF {#1} \l_temps {
        \seq_put_right:Nn \l_mytmps {##1}
      } { }
    }
    \seq_get_right:NN \l_mytmps \g_wordtmp
  } {
    \seq_get_right:NN \g_prove_mru \g_wordtmp
  }
  \tl_use:N \g_wordtmp
  \IfValueTF{#1}{}{~}
  \seq_gput_left:NV \g_prove_mru \g_wordtmp
  \seq_gremove_duplicates:N \g_prove_mru
}

\NewDocumentCommand{\proves}{ o }{
  \IfValueTF{#1}{
    \seq_clear:N \l_mytmps
    \seq_map_inline:Nn \g_prove_mru {
      \str_if_eq:nnTF {##1} {ensure} {
        \str_set:Nn \l_temps {n}
      } {
        \str_set:Nx \l_temps {\str_head_ignore_spaces:n {##1}}
      }
      \str_if_in:xVTF {#1} \l_temps {
        \seq_put_right:Nn \l_mytmps {##1}
      } { }
    }
    \seq_get_right:NN \l_mytmps \g_wordtmp
  } {
    \seq_get_right:NN \g_prove_mru \g_wordtmp
  }
  \str_set:NV \l_tmpa_str \g_wordtmp
  \prop_get:NVN \l__verbs \l_tmpa_str \l_tmpa_tl
  \tl_use:N \l_tmpa_tl
  \IfValueTF{#1}{}{~}
  \seq_gput_left:NV \g_prove_mru \g_wordtmp
  \seq_gremove_duplicates:N \g_prove_mru
}

\newcommand{\llabel}[1]{\savelabel{#1}\label{\loc.#1}\ignorespaces}

\clist_new:N \l_localreflist
\clist_new:N \l_reflist

\NewDocumentCommand{\lref} { m } {
  \clist_set:No \l_localreflist {#1}
  \clist_clear:N \l_reflist
  \clist_map_inline:Nn \l_localreflist { \clist_put_right:Nn \l_reflist {\loc.##1} }
  \cref{\l_reflist}
}

\NewDocumentCommand{\Lref} { m } {
  \clist_set:No \l_localreflist {#1}
  \clist_clear:N \l_reflist
  \clist_map_inline:Nn \l_localreflist { \clist_put_right:Nn \l_reflist {\loc.##1} }
  \Cref{\l_reflist}
}

\NewDocumentCommand{\itref}{ m m }{
  \clist_set:No \l_localreflist {#2}
  \clist_clear:N \l_reflist
  \clist_map_inline:Nn \l_localreflist { \clist_put_right:Nn \l_reflist {#1.##1} }
  \cref{\l_reflist}~in~\cref{#1}
}

\seq_new:N \l_enum_seq
\int_new:N \l_num_items

\bool_new:N \g_commaused_bool

\providecommand{\comma}{}

\cs_new:Nn \enum_it:nn {
  \int_case:nnF {\l_num_items - #1} {
    {0} {
      \renewcommand{\comma}{}
      #2\space
    }
    {1} {
      \bool_gset_false:N \g_commaused_bool
      \renewcommand{\comma}{,~\bool_gset_true:N \g_commaused_bool}
      #2
      \bool_if:NTF \g_commaused_bool {} {,~}
      and~
    }
  } {
    \bool_gset_false:N \g_commaused_bool
    \renewcommand{\comma}{,~\bool_gset_true:N \g_commaused_bool}
    #2
    \bool_if:NTF \g_commaused_bool {} {,~}
  }
}

\cs_new:Nn \enum_it_U:nn {
  \int_case:nnF {\l_num_items - #1} {
    {0} {
      \renewcommand{\comma}{}
      #2
      \space
    }
    {1} {
      \bool_gset_false:N \g_commaused_bool
      \renewcommand{\comma}{,~\bool_gset_true:N \g_commaused_bool}
      #2
      \bool_if:NTF \g_commaused_bool {} {,~}
      and~
    }
  } {
    \bool_gset_false:N \g_commaused_bool
    \renewcommand{\comma}{,~\bool_gset_true:N \g_commaused_bool}
    \int_compare:nTF {#1=1} {
      \text_titlecase_first:n {#2}
    } {
      #2
    }
    \bool_if:NTF \g_commaused_bool {} {,~}
  }
}

\cs_generate_variant:Nn \tl_if_eq:nnTF {onTF}

\cs_new:Nn \enum:nnnn {
  \seq_set_split:Nnn \l_enum_seq ; {#1}
  \seq_remove_all:Nn \l_enum_seq { }
  \seq_remove_all:Nn \l_enum_seq {#2}
  \seq_log:N \l_enum_seq
  \int_set:Nn \l_num_items {\seq_count:N \l_enum_seq}
  \int_log:N \l_num_items
  \int_case:nnF {\l_num_items} {
    { 0 } { 0 }
    { 1 } {
      \IfBooleanTF{#4} {
        \tl_set:Nn \l_text_case_exclude_arg_tl {\cref}
        \bool_set_false:N \l_text_titlecase_check_letter_bool
        \text_titlecase_first:n {\seq_use:Nn \l_enum_seq {}}
      } {
        \seq_use:Nn \l_enum_seq {}
      }
      \space
      \tl_if_eq:onTF{#3}{-}{}{
        \bool_if:NTF \l_plural_bool {
          \prove[#3]~
        } {
          \proves[#3]~
        }
      }
    }
    { 2 } {
      \IfBooleanTF{#4} {
        \tl_set:Nn \l_text_case_exclude_arg_tl {\cref}
        \bool_set_false:N \l_text_titlecase_check_letter_bool
        \text_titlecase_first:n {\seq_item:Nn \l_enum_seq {1}}
        {} ~and~
        \seq_item:Nn \l_enum_seq {2}
      } {
        \seq_use:Nn \l_enum_seq {~and~}
      }
      \space
      \tl_if_eq:onTF{#3}{-}{}{
        \prove[#3]~
      }
    }
  } {
    \IfBooleanTF{#4} {
      \tl_set:Nn \l_text_case_exclude_arg_tl {\cref}
      \seq_indexed_map_function:NN \l_enum_seq \enum_it_U:nn
    } {
      \seq_indexed_map_function:NN \l_enum_seq \enum_it:nn
    }
    \tl_if_eq:onTF{#3}{-}{}{
      \prove[#3]~
    }
  }
}

\cs_generate_variant:Nn \enum:nnnn {nxnn}
\cs_generate_variant:Nn \enum:nnnn {nxxn}

\NewDocumentCommand{\enum}{O{} m O{-} s}{
  \IfBooleanTF{#4}{
    \enum:nxnn {#2} {#1} {sindep} \BooleanFalse
  } {
    \enum:nxxn {#2} {#1} {#3} \BooleanFalse
  }
}

\NewDocumentCommand{\dott}{}{\ifnocf{.}\space}

\bool_new:N \g_arg_start_bool
\bool_gset_true:N \g_arg_start_bool

\NewDocumentCommand{\startnewargseq}{}{\bool_gset_true:N \g_arg_start_bool \tl_set:Nn \g_label_tl {}}

\cs_generate_variant:Nn \seq_if_in:NnTF {NxTF}
\cs_generate_variant:Nn \seq_remove_all:Nn {Nx}

\int_new:N \l_random_int

\cs_generate_variant:Nn \tl_if_head_eq_catcode:nNTF {oNTF}
\cs_generate_variant:Nn \tl_if_head_eq_catcode:nNTF {VNTF}

\cs_generate_variant:Nn \tl_if_head_eq_catcode:nNTF {eNTF}
\cs_generate_variant:Nn \tl_log:n {o}
\cs_generate_variant:Nn \tl_log:n {f}
\cs_generate_variant:Nn \tl_log:n {x}
\cs_generate_variant:Nn \tl_log:n {e}

\cs_generate_variant:Nn \tl_if_in:nnTF {onTF}
\cs_generate_variant:Nn \tl_if_in:NnTF {NeTF}

\cs_generate_variant:Nn \tl_if_head_eq_meaning:nNTF {VNTF}

\seq_const_from_clist:Nn \g_arg_mru_this {
  Ahpr,
  Tapr,
  Ctapr,
  H
}

\seq_const_from_clist:Nn \g_arg_mru_nothis {
  Ia,
  Nwc,
  N,
  Itns,
  Fm,
  Itnswc,
  Mo
}

\seq_new:N \l_arg_seq
\tl_new:N \l_cons_tl
\tl_new:N \l_dummy_tl

\bool_new:N \g_debug_bool
\bool_gset_false:N \g_debug_bool

\bool_new:N \l_insidearg_bool

\bool_new:N \g_firstargletter_bool

\sys_gset_rand_seed:n {0903}

\bool_new:N \l_plural_bool
\tl_new:N \l_arg_verbs_tl

\NewDocumentCommand{\argument}{mom}{
\color{blue}
  \bool_set_false:N \l_plural_bool
  \tl_set:Nn \l_arg_verbs_tl {sindep}
  \keys_define:nn { benno/argument } {
    plural .value_forbidden:n = true,
    plural .code:n = {\bool_set_true:N \l_plural_bool},
    verbs .value_required:n = false,
    verbs .tl_set:N = \l_arg_verbs_tl,
  }
  \IfValueT{#2}{
    \keys_set:nn { benno/argument } {#2}
  }
  \bool_log:N \l_plural_bool
  \bool_gset_true:N \l_insidearg_bool
  \seq_set_split:Nnn \l_arg_seq ; {#1}
  \seq_remove_all:Nn \l_arg_seq { }
  \seq_log:N \l_arg_seq
  \tl_set:Nn \l_cons_tl {#3}
  \tl_trim_spaces:N \l_cons_tl
  \seq_if_in:NxTF \l_arg_seq {\lref{\g_label_tl}} {
    \seq_remove_all:Nx \l_arg_seq {\lref{\g_label_tl}}
    \seq_get_left:NNTF \l_arg_seq \l_dummy_tl {
      \tl_trim_spaces:N \l_dummy_tl
      \bool_gset_false:N \g_firstargletter_bool
      \tl_if_head_eq_catcode:VNTF \l_dummy_tl a {
        \bool_gset_true:N \g_firstargletter_bool
      } {
        \tl_if_head_eq_meaning:VNTF \l_dummy_tl {\cref} {
          \tl_set:Nx \l_tmpa_tl {\tl_tail:N \l_dummy_tl}
          \tl_set:Nx \l_tmpb_tl {\tl_head:N \l_tmpa_tl}
          \bool_gset_true:N \g_firstargletter_bool
          \tl_if_in:NeTF \l_tmpb_tl {lem\c_colon_str} {} {
            \tl_if_in:NeTF \l_tmpb_tl {thm\c_colon_str} {} {
              \tl_if_in:NeTF \l_tmpb_tl {prop\c_colon_str} {} {
                \tl_if_in:NeTF \l_tmpb_tl {cor\c_colon_str} {} {
                  \bool_gset_false:N \g_firstargletter_bool
                }
              }
            }
          }
        } {
        }
      }
      \bool_if:NTF \g_firstargletter_bool {
        \seq_set_eq:NN \l_tmpa_seq \g_arg_mru_this
        \seq_remove_all:Nn \l_tmpa_seq {H}
        \seq_get_right:NN \l_tmpa_seq \l_tmpa_tl
        \int_case:nnF {\seq_count:N \l_arg_seq} {
          {1} {
            \str_case:VnF {\l_tmpa_tl} {
              {Ahpr} {
                \bool_if:NT \g_debug_bool {C1.1}
                \seq_gput_left:Nn \g_arg_mru_this {Ahpr}
                \seq_gremove_duplicates:N \g_arg_mru_this
                \enum:nxnn {#1} {\lref{\g_label_tl}} {-} {\BooleanTrue}
                \hence~
                \bool_if:NTF \l_plural_bool {
                  \prove[\l_arg_verbs_tl]~\ignorespaces #3
                } {
                  \proves[\l_arg_verbs_tl]~\ignorespaces #3
                }
              }
              {Tapr} {
                \bool_if:NT \g_debug_bool {C1.2}
                \seq_gput_left:Nn \g_arg_mru_this {Tapr}
                \seq_gremove_duplicates:N \g_arg_mru_this
                \enum[\lref{\g_label_tl}]{
                  This;
                  #1
                }[\l_arg_verbs_tl]\ignorespaces #3
              }
              {Ctapr} {
                \bool_if:NT \g_debug_bool {C1.3}
                \seq_gput_left:Nn \g_arg_mru_this {Ctapr}
                \seq_gremove_duplicates:N \g_arg_mru_this
                Combining~
                \enum[\lref{\g_label_tl}]{
                  this;
                  #1
                } \proves[\l_arg_verbs_tl]~\ignorespaces #3
              }
            } {}
          }
        } {
          \str_case:VnF {\l_tmpa_tl} {
             {Ahpr} {
              \bool_if:NT \g_debug_bool {C2.1}
              \seq_gput_left:Nn \g_arg_mru_this {Ahpr}
              \seq_gremove_duplicates:N \g_arg_mru_this
              \enum:nxnn {#1} {\lref{\g_label_tl}} {-} {\BooleanTrue}
              \hence~
              \prove[\l_arg_verbs_tl]~\ignorespaces #3
            }
            {Tapr} {
              \bool_if:NT \g_debug_bool {C2.2}
              \seq_gput_left:Nn \g_arg_mru_this {Tapr}
              \seq_gremove_duplicates:N \g_arg_mru_this
              \enum[\lref{\g_label_tl}]{
                This;
                #1
              }[\l_arg_verbs_tl]\ignorespaces #3
            }
            {Ctapr} {
              \int_case:nn {\int_rand:nn {0} {1}} {
                {0} {
                  \bool_if:NT \g_debug_bool {C2.3}
                  \seq_gput_left:Nn \g_arg_mru_this {Ctapr}
                  \seq_gremove_duplicates:N \g_arg_mru_this
                  Combining~
                  \enum[\lref{\g_label_tl}]{
                    this;
                    #1
                  } \proves[\l_arg_verbs_tl]~\ignorespaces #3
                }
                {1} {
                  \bool_if:NT \g_debug_bool {C2.4}
                  \seq_gput_left:Nn \g_arg_mru_this {Ctapr}
                  \seq_gremove_duplicates:N \g_arg_mru_this
                  Combining~
                  \enum:nxnn {#1} {\lref{\g_label_tl}} {-} {\BooleanFalse}
                  \hence~
                  \proves[\l_arg_verbs_tl]~\ignorespaces #3
                }
              }
            }
          } {}
        }
      } {
        \seq_set_eq:NN \l_tmpa_seq \g_arg_mru_this
        \seq_remove_all:Nn \l_tmpa_seq {H}
        \seq_remove_all:Nn \l_tmpa_seq {Ahpr}
        \seq_get_right:NN \l_tmpa_seq \l_tmpa_tl
        \int_case:nnF {\seq_count:N \l_arg_seq} {
          {1} {
            \str_case:VnF {\l_tmpa_tl} {
              {Tapr} {
                \bool_if:NT \g_debug_bool {C3.1}
                \seq_gput_left:Nn \g_arg_mru_this {Tapr}
                \seq_gremove_duplicates:N \g_arg_mru_this
                \enum[\lref{\g_label_tl}]{
                  This;
                  #1
                }[\l_arg_verbs_tl]\ignorespaces #3
              }
              {Ctapr} {
                \bool_if:NT \g_debug_bool {C3.2}
                \seq_gput_left:Nn \g_arg_mru_this {Ctapr}
                \seq_gremove_duplicates:N \g_arg_mru_this
                Combining~
                \enum[\lref{\g_label_tl}]{
                  this;
                  #1
                } \proves[\l_arg_verbs_tl]~\ignorespaces #3
              }
            } {}
          }
        } {
          \str_case:VnF {\l_tmpa_tl} {
            {Tapr} {
              \bool_if:NT \g_debug_bool {C4.1}
              \seq_gput_left:Nn \g_arg_mru_this {Tapr}
              \seq_gremove_duplicates:N \g_arg_mru_this
              \enum[\lref{\g_label_tl}]{
                This;
                #1
              }[\l_arg_verbs_tl]\ignorespaces #3		
            }
            {Ctapr} {
              \int_case:nn {\int_rand:nn {0} {1}} {
                {0} {
                  \bool_if:NT \g_debug_bool {C4.2}
                  \seq_gput_left:Nn \g_arg_mru_this {Ctapr}
                  \seq_gremove_duplicates:N \g_arg_mru_this
                  Combining~
                  \enum[\lref{\g_label_tl}]{
                    this;
                    #1
                  } \proves[\l_arg_verbs_tl]~\ignorespaces #3		
                }
                {1} {
                  \bool_if:NT \g_debug_bool {C4.3}
                  \seq_gput_left:Nn \g_arg_mru_this {Ctapr}
                  \seq_gremove_duplicates:N \g_arg_mru_this
                  Combining~
                  \enum:nxnn {#1} {\lref{\g_label_tl}} {-} {\BooleanFalse}
                  \hence~
                  \proves[\l_arg_verbs_tl]~\ignorespaces #3    
                }
              }
            }
          } {}
        }
      }
    } {
      \tl_if_head_eq_catcode:oNTF \l_cons_tl a {
        \seq_set_eq:NN \l_tmpa_seq \g_arg_mru_this
        \seq_remove_all:Nn \l_tmpa_seq {Ctapr}
        \seq_remove_all:Nn \l_tmpa_seq {Ahpr}
        \seq_get_right:NN \l_tmpa_seq \l_tmpa_tl
        \str_case:VnF {\l_tmpa_tl} {
          {H} {
            \bool_if:NT \g_debug_bool {C5.1}
            \seq_gput_left:Nn \g_arg_mru_this {H}
            \seq_gremove_duplicates:N \g_arg_mru_this
            Hence,~we~obtain~\ignorespaces #3
          }
          {Tapr} {
            \bool_if:NT \g_debug_bool {C5.2}
            \seq_gput_left:Nn \g_arg_mru_this {Tapr}
            \seq_gremove_duplicates:N \g_arg_mru_this
            This~\proves[\l_arg_verbs_tl]~\ignorespaces #3
          }
        } {}
      } {
        \bool_if:NT \g_debug_bool {C6.1}
        \seq_gput_left:Nn \g_arg_mru_this {Tapr}
        \seq_gremove_duplicates:N \g_arg_mru_this
        This~\proves[\l_arg_verbs_tl]~\ignorespaces #3
      }
    } 
  } {
    \int_compare:nNnTF {\seq_count:N \l_arg_seq} = {0} {
      \bool_if:NTF \g_arg_start_bool {
        \bool_if:NT \g_debug_bool {C7.1}
        \Nobs\unskip
        #3
      } {
        \bool_if:NT \g_debug_bool {C7.2}
        \Moreover~
        #3
      }
    } {
      \bool_if:NTF \g_arg_start_bool {
        \bool_if:NT \g_debug_bool {C8.1}
        \tl_log:N \l_arg_verbs_tl
        \Nobs~that~
        \enum{
          #1
        }[\l_arg_verbs_tl]\ignorespaces #3
      } {
        \int_compare:nNnTF {\seq_count:N \l_arg_seq} = {1} {
          \seq_set_eq:NN \l_tmpa_seq \g_arg_mru_nothis
          \seq_remove_all:Nn \l_tmpa_seq {Nwc}
          \seq_remove_all:Nn \l_tmpa_seq {Itnswc}
          \seq_get_right:NN \l_tmpa_seq \l_tmpa_tl
        } {
          \seq_get_right:NN \g_arg_mru_nothis \l_tmpa_tl
        }
        \str_case:VnF {\l_tmpa_tl} {
          {Mo} {
            \bool_if:NT \g_debug_bool {C9.1}
            \seq_gput_left:Nn \g_arg_mru_nothis {Mo}
            \seq_gremove_duplicates:N \g_arg_mru_nothis
            Moreover,~\nobs~that~
            \enum{
              #1
            }[\l_arg_verbs_tl]\ignorespaces #3		
          }
          {Fm} {
            \bool_if:NT \g_debug_bool {C9.2}
            \seq_gput_left:Nn \g_arg_mru_nothis {Fm}
            \seq_gremove_duplicates:N \g_arg_mru_nothis
            Furthermore,~\nobs~that~
            \enum{
              #1
            }[\l_arg_verbs_tl]\ignorespaces #3		
          }
          {Ia} {
            \bool_if:NT \g_debug_bool {C9.3}
            \seq_gput_left:Nn \g_arg_mru_nothis {Ia}
            \seq_gremove_duplicates:N \g_arg_mru_nothis
            In~addition,~\nobs~that~
            \enum{
              #1
            }[\l_arg_verbs_tl]\ignorespaces #3		
          }
          {N} {
            \bool_if:NT \g_debug_bool {C9.4}
            \seq_gput_left:Nn \g_arg_mru_nothis {N}
            \seq_gremove_duplicates:N \g_arg_mru_nothis
            Next,~\nobs~that~
            \enum{
              #1
            }[\l_arg_verbs_tl]\ignorespaces #3		
          }
          {Itns} {
            \bool_if:NT \g_debug_bool {C9.5}
            \seq_gput_left:Nn \g_arg_mru_nothis {Itnswc}
            \seq_gput_left:Nn \g_arg_mru_nothis {Itns}
            \seq_gremove_duplicates:N \g_arg_mru_nothis
            In~the~next~step~we~\nobs~that~
            \enum{
              #1
            }[\l_arg_verbs_tl]\ignorespaces #3		
          }
          {Nwc} {
            \bool_if:NT \g_debug_bool {C9.6}
            \seq_gput_left:Nn \g_arg_mru_nothis {Nwc}
            \seq_gremove_duplicates:N \g_arg_mru_nothis
            Next~we~combine~
            \enum{
              #1
            }to~obtain~\ignorespaces #3
          }
          {Itnswc} {
            \bool_if:NT \g_debug_bool {C9.7}
            \seq_gput_left:Nn \g_arg_mru_nothis {Itns}
            \seq_gput_left:Nn \g_arg_mru_nothis {Itnswc}
            \seq_gremove_duplicates:N \g_arg_mru_nothis
            In~the~next~step~we~combine~
            \enum{
              #1
            }to~obtain~\ignorespaces #3
          }
        } {}
      }
    }
  }
  \bool_gset_false:N \g_arg_start_bool
  \bool_gset_false:N \l_insidearg_bool
  \cfload[.]
  \color{black}
}

\tl_new:N \g_label_tl
\tl_gset:Nn \g_label_tl { }

\NewDocumentCommand{\savelabel}{m}{
  \bool_if:NTF \l_insidearg_bool {
    \tl_gset:Nn \g_label_tl {#1}
  } {
    \tl_gset:Nn \g_label_tl { }
  }
}

\ExplSyntaxOff

\ExplSyntaxOn

\NewDocumentEnvironment {athm} {m m o} {
\str_if_eq:noTF {example} {#1} {
  \bool_gset_true:N \g_example_bool
} {
  \bool_gset_false:N \g_example_bool
}
\cfclear
\IfNoValueTF{#3}{
\begin{#1}\label{#2}\global\def\loc{#2}
}{
\begin{#1}[#3]\label{#2}\global\def\loc{#2}
}
}{
\end{#1}
}

\NewDocumentEnvironment {adef} {m} {
\begin{definition}\label{#1}\global\def\loc{#1}
}{
\end{definition}
}

\NewDocumentEnvironment{aproof} {} {
\bool_if:NTF \g_example_bool {
  \bool_gset_true:N \g_arg_start_bool
  \begin{proof}[Proof~for~\cref{\loc}]
} {
  \bool_gset_true:N \g_arg_start_bool
  \begin{proof}[Proof~of~\cref{\loc}]
}
\bool_gset_false:N \g_finishproof_bool
}{
\bool_if:NTF \g_finishproof_bool {}
{\finishproofthus}
\end{proof}
}

\NewDocumentCommand{\finishproofthus} {} {
  \bool_gset_true:N \g_finishproof_bool 
  \bool_if:NTF \g_example_bool {
    The~proof~for~\cref{\loc}~is~thus~complete.
  } {
    The~proof~of~\cref{\loc}~is~thus~complete.
  }
}
\NewDocumentCommand{\finishproofthis} {} {
  \bool_gset_true:N \g_finishproof_bool 
  \bool_if:NTF \g_example_bool {
    This~completes~the~proof~for~\cref{\loc}.
  } {
    This~completes~the~proof~of~\cref{\loc}.
  }
}

\ExplSyntaxOff

\NewDocumentEnvironment{cproof}{m}
{\begin{proof}[Proof of \cref{#1}]}%
{\noindent The proof of \cref{#1} is thus complete.
\end{proof}}

\NewDocumentEnvironment{cproof2}{m}
{\begin{proof}[Proof of \cref{#1}]}%
{\noindent This completes the proof of \cref{#1}.
\end{proof}}

\hypersetup{
    colorlinks,
    linkcolor={blue!80!black},
    citecolor={green},
    urlcolor={blue!80!black}
}

\makeatletter
\ExplSyntaxOn
\seq_new:N \g_abbrs
\prop_new:N \g_abbr_counts
\tl_new:N \l_abbr_count_tl

\ExplSyntaxOn

\bool_new:N \g_forexample

\NewDocumentCommand{\eg}{ o }{
	\IfValueT{#1}{
		\str_if_eq:noTF {fe} {#1} {
			\bool_gset_true:N \g_forexample
		} {\bool_gset_false:N \g_forexample}
	}
	\bool_if:nTF { \g_forexample } {
		\bool_gset_false:N \g_forexample
		for~example
	}{
		\bool_gset_true:N \g_forexample
		for~instance
	}
}

\NewDocumentCommand{\abbr}{m m O{#1} m m O{#4} m}{
	\expandafter\newcommand\csname#3\endcsname[1][]{
		\seq_if_in:NnTF \g_abbrs {#1} {
			\prop_get:NnN \g_abbr_counts {#1} \l_abbr_count_tl
			\prop_gput:Nnx \g_abbr_counts {#1} {\int_eval:n {\l_abbr_count_tl + 1}}
			\hyperref[#1]{#7}
		} {
			\seq_gput_left:Nn \g_abbrs {#1}
			\prop_gput:Nnn \g_abbr_counts {#1} {1}
			\expandafter\gdef\csname#1@def\endcsname{#2}
			\phantomsection\label{#1}
			\str_if_eq:nnTF{##1}{}{\emph{#2}}{##1}~(\hyperref[#1]{#7})
		}
	}
	\expandafter\newcommand\csname#6\endcsname[1][]{
		\seq_if_in:NnTF \g_abbrs {#1} {
			\prop_get:NnN \g_abbr_counts {#1} \l_abbr_count_tl
			\prop_gput:Nnx \g_abbr_counts {#1} {\int_eval:n {\l_abbr_count_tl + 1}}
			\hyperref[#1]{#4}
		} {
			\expandafter\gdef\csname#1@def\endcsname{#5}
			\seq_gput_left:Nn \g_abbrs {#1}
			\prop_gput:Nnn \g_abbr_counts {#1} {1}
			\phantomsection\label{#1}
			\str_if_eq:nnTF{##1}{}{\emph{#5}}{##1}~(\hyperref[#1]{#4})
		}
	}
}

\ExplSyntaxOff
		\makeatother
		\abbr{Adamax}{adaptive moment estimation maximum SGD}{Adamaxs}{Adaptive moment estimation maximum SGD}{Adamax}
		\abbr{Adagrad}{adaptive gradient SGD}{Adagrads}{adaptive gradient SGD}{Adagrad}
		\abbr{Nadam}{Nesterov accelerated adaptive moment estimation SGD}{Nadams}{Nesterov accelerated adaptive moment estimation SGD}{Nadam}
		\abbr{RMSprop}{root mean square propagation SGD}{RMSprops}{the root mean square propagation}{RMSprop}
		\abbr{ANN}{artificial neural network}{ANNs}{artificial neural networks}{ANN}
\abbr{DNN}{deep artificial neural network}{DNNs}{deep artificial neural networks}{DNN}
		\abbr{SGD}{Stochastic gradient descent}{SGDs}{Stochastic gradient descent}{SGD}
        \abbr{Adam}{adaptive moment estimation \SGD}{Adams}{adaptive moment estimation \SGD}{Adam}
		\abbr{iid}{independent and identically distributed}{i.i.d}{independent and identically distributed}{i.i.d.}
		\abbr{DL}{Deep learning}{DLs}{Deep learning}{DL}
        \abbr{RePUs}{rectified power unit}{RePU}{rectified power unit}{RePU}
		\abbr{AI}{artificial intelligence}{AIs}{artificial intelligence}{AI}
		\abbr{LLM}{large language model}{LLMs}{large language models}{LLM}
		\abbr{PDE}{partial differential equation}{PDEs}{partial differential equations}{PDE}
		\abbr{as}{almost surely}{ass}{almost surely}{a.s.}
		\abbr{pas}{$\P$-almost surely}{pass}{almost surely}{$\P$-a.s.}
		\abbr{ReLUs}{rectified linear unit}{ReLU}{rectified linear unit}
		\makeatother

		\title{Non-convergence to the optimal risk
for Adam\\ and stochastic gradient descent optimization\\
in the training of deep neural networks}
\author{Thang Do$^{1,2}$, Arnulf Jentzen$^{3,4}$, and Adrian Riekert$^{5}$
	\bigskip
	\\
    	\small{$^1$ School of Data Science, The Chinese University of Hong Kong, Shenzhen}
	\vspace{-0.1cm}\\
	\small{ (CUHK-Shenzhen), China, e-mail: \texttt{minhthangdo@link.cuhk.edu.cn}}
 \smallskip
	\\
 \small{$^2$ Deparment of Probability and Statistic, Institute of Mathematics,}
	\vspace{-0.1cm}\\
	\small{Vietnam Academy of Science and Technology, Vietnam, e-mail: \texttt{dmthang@math.ac.vn}}
	\smallskip
	\\
	\small{$^3$ School of Data Science and Shenzhen Research Institute of Big Data, The Chinese University}
	\vspace{-0.1cm}\\
	\small{of Hong Kong, Shenzhen (CUHK-Shenzhen), China, e-mail: \texttt{ajentzen@cuhk.edu.cn}}
	\smallskip
	\\
 \small{$^4$ Applied Mathematics: Institute for Analysis and Numerics, Faculty of Mathematics and}
	\vspace{-0.1cm}\\
	\small{Computer Science, University of M{\"u}nster, Germany, e-mail: \texttt{ajentzen@uni-muenster.de}}
	\smallskip
	\\
    \small{$^5$ Applied Mathematics: Institute for Analysis and Numerics, Faculty of Mathematics and}
	\vspace{-0.1cm}\\
	\small{Computer Science, University of M{\"u}nster, Germany, e-mail: \texttt{ariekert@uni-muenster.de}}
	\smallskip
	\\
}

		\date{\today}
		\allowdisplaybreaks 

        \newcommand{\reli}[4]{\mathcal{N}^{#1,#2,#3}_{{#4}}\cfadd{definition: ANN}}
\newcommand{\relii}[5]{\mathcal{N}^{#1,#2,#3}_{#4,#5}\cfadd{definition: ANN}}
\newcommand{\rel}[3]{\mathbf{N}^{#1,#2}_{#3}\cfadd{definition: ANN}}
\newcommand{\rell}[4]{\mathscr{N}^{#1,#2}_{#3,#4}\cfadd{definition: ANN}}
\newcommand{\ffd}{\mathfrak{d}\cfadd{definition: ANN}}
		
\begin{document}
			\maketitle
            \begin{abstract}
                Despite the omnipresent use of stochastic gradient descent (SGD) optimization methods in the training of deep neural networks (DNNs), it remains, in basically all practically relevant scenarios, a fundamental open problem to provide a rigorous theoretical explanation for the success (and the limitations) of SGD optimization methods in deep learning. In particular, it remains an open question to prove or disprove convergence of the true risk of SGD optimization methods to the optimal true risk value in the training of DNNs. In one of the main results of this work we reveal for a general class of activations, loss functions, random initializations, and SGD optimization methods (including, for example, standard SGD, momentum SGD, Nesterov
accelerated SGD, Adagrad, RMSprop, Adadelta, Adam, Adamax, Nadam, Nadamax, and AMSGrad) that in the training of any arbitrary fully-connected feedforward DNN it does not hold that the true risk of the considered optimizer converges in probability to the optimal true risk value. Nonetheless, the true risk of the considered SGD optimization method may very well converge to a strictly suboptimal true risk value.

            \end{abstract}
			\tableofcontents
			\section{Introduction}
			\newcommand{\scrd}{\mathscr{d}}
			\newcommand\restr[2]{{
					\left.\kern-\nulldelimiterspace 
					#1 
					\vphantom{|} 
					\right|_{#2} 
			}}
			\newcommand{\smalll}{\mathfrak{l}}
			\newcommand{\sml}{l}
            \SGD\ optimization methods are the method of choice to train deep \ANNs\ in data-driven learning problems (see, \eg, \cite{geminiteam2024geminifamilyhighlycapable,Brownlanguagemodel2020,Yihengsummary2023,ramesh2021zeroshottexttoimagegeneration,Rombachstablediffusion2021,Sahariaimagen2022} and the references therein) as well as scientific computing problems (see, \eg, \cite{Beck_2023,WhatsNext,MR4356985,Gemain21,MR4795589,Ruf19} and the references therein). However, often not the plain vanilla standard \SGD\ optimization method is the employed optimizer but instead more sophisticated accelerated and adaptive variants of the standard \SGD\ method such as the \Adam\ optimizer (see \cite{KingmaBa2024_Adam}) are used in practically relevant deep \ANN\ training problems. We also refer, \eg, to \cite{bach2024learning,WeChaoSteLei2020,Goodfellow-et-al-2016,ArBePhi2024,PhilipJa2024,ruder2017overviewgradientdescentoptimization} for monographs and surveys treating \SGD\ optimization methods for the training of \ANNs. 


The considered \SGD\ optimization method is used with the aim to minimize the true risk function (the objective function) of the considered \ANN\ learning problem so that, roughly speaking, the realization function of the deep \ANN\ minimizing the true risk function approximates as best as possible the output data given the input data. Despite the omnipresent use of \SGD\ optimization methods in the training of \ANNs, it remains, in basically all practically relevant scenarios, a fundamental open problem to provide a rigorous theoretical description and explanation for the convergence (and non-convergence) properties of \SGD\ optimization methods in deep learning. In particular, it remains an open question to prove or disprove convergence of the true risk of SGD optimization methods to the optimal/infimal true risk value in the training of deep \ANNs\ (cf., \eg, \cite{CHERIDITO2021101540,HannibalJentzenThang2024,ArAd2024,Lu_2020} and the literature review in \cref{subsec: literature review} below). In this work we contribute to this open problem of research in two aspects. 

\begin{enumerate}[label=(\roman*)]
   
 \item \label{item 1: introduction} In our main results in \cref{main cor: scientific square mean} and \cref{main cor: scientific general loss very deep} in \cref{subsec: positive probability} below (see also \cref{main cor: scientific square mean self-contained} in \cref{subsec: positive probability}) we reveal for a large class of activations, loss functions, random initializations, and \SGD\ optimization methods (including, for example, the \Adam\ and the standard \SGD\ optimizers) that in the training of any arbitrary deep fully-connected feedforward \ANN\ we have \emph{with strictly positive probability} that the true risk of the considered \SGD\ method does \emph{not} converge to the optimal/infimal true risk value. To briefly sketch the contributions of \cref{main cor: scientific square mean} and \cref{main cor: scientific general loss very deep} within this introductory section, we present in \cref{main theorem1} in \cref{subsec: main theorem 1} below a special case of \cref{main cor: scientific square mean} in which we restrict ourselves, among other simplifications and restrictions, to the plain vanilla standard \SGD\ optimization method and we refer to \cref{main cor: scientific square mean} for the more general non-convergence result that applies to a general class of \SGD\ optimization methods including, for instance, the \Adam\ optimizer.

\item \label{item 2: introduction} In our further main results in \cref{main theo: scientific square mean error very deep} and \cref{main theo: scientific other error very deep} in \cref{subsec: high probability} below (see also \cref{main theo: scientific square mean error very deep self-contained} in \cref{subsec: high probability}) we prove for a large class of activations, loss functions, random initializations, and \SGD\ optimization methods (also covering, for example, the \Adam\ and the standard \SGD\ optimizers) that in the training of \emph{sufficiently deep} fully-connected feedforward \ANNs\ we have \emph{with high probability} that the true risk of the considered \SGD\ method does \emph{not} converge to the optimal/infimal true risk value. In \cref{main theorem2} in \cref{subsec: main theorem 2} within this introductory section we present a special case of \cref{main theo: scientific square mean error very deep} in which we also restrict us to the standard \SGD\ optimizer and we refer to \cref{main theo: scientific square mean error very deep} for the more general non-convergence result that is, \eg, also applicable to the \Adam\ optimizer.
\end{enumerate}
In our precise formulation of \cref{main theorem1} and \cref{main theorem2} below 
we employ realization functions of fully-connected feedforward \ANNs. In \cref{definition: ANN} in \cref{subsec: definition ANN} we briefly recall the mathematical description of realization functions of such \ANNs\ in a vectorized format (cf., \eg, \cite{HannibalJentzenThang2024,ArBePhi2024} and the references therein) and, thereafter, we present in \cref{subsec: main theorem 1,subsec: main theorem 2} below the two above sketched special cases (see \cref{item 1: introduction,item 2: introduction} above) of some of the main results of this works. 
                \subsection{Deep artificial neural networks (ANNs)}\label{subsec: definition ANN}
                The realization function of a standard fully-connected feedforward \ANN\ is typically given as multiple compositions of affine functions and multidimensional versions of a certain fixed one-dimensional nonlinear function, usually referred to as activation function.
In \cref{definition: ANN} we present for every number of affine functions $L \in \N = \{ 1, 2, 3, \dots \}$, every \ANN\ architecture vector $\ell = ( \ell_0, \ell_1, \dots, \ell_L ) \in \N^{ L + 1 }$, every \ANN\ parameter vector 
$\theta = ( \theta_1,\dots, \theta_{ \sum_{i=1}^L\ell_i(\ell_{i-1}+1) } ) \in \R^{ \sum_{i=1}^L\ell_i(\ell_{i-1}+1)}$,
and every activation function $A \colon \R \to \R$ 
the realization function $\rel{\ell}{\theta}{A}\colon \R^{\ell_0}\to\R^{\ell_L}$ of the \ANN\ with architecture vector $\ell$, parameter vector $\theta$, and activation function $A$ (cf., \eg, \cite{HannibalJentzenThang2024,ArBePhi2024}). 
\begin{samepage}
            \begin{definition}\label{definition: ANN}
        For every $L\in \N$, $\ell=(\ell_0,\ell_1,\dots,\ell_L)\in \N^{L+1}$ we denote by $\ffd(\ell)\in \N$ the natural number which satisfies
        \begin{equation}\label{def: fd(ell)}
\ffd(\ell)=\textstyle\sum\limits_{i=1}^{L}\ell_i ( \ell_{ i - 1 } + 1 )
        \end{equation}
        and for every $L\in \N$, $\ell=(\ell_0,\ell_1,\dots,\ell_L)\in \N^{L+1}$, $\theta\in \R^{\ffd(\ell)}$ and every $A\colon \R\to\R$ we denote by $\rel{\ell}{\theta}{A}\colon \R^{\ell_0}\to\R^{\ell_L}$ and $\reli{\ell}{v}{\theta}{A}=(\relii{\ell}{v}{\theta}{A}{1},\dots,\relii{\ell}{v}{\theta}{A}{\ell_v})\colon \R^{\ell_0}\to\R^{\ell_v}$, $v\in \{0,1,\dots,L\}$, the functions which satisfy for all $v\in \{0,1,\dots,L-1\}$, $x=(x_1,\dots,x_{\ell_0})\in \R^{\ell_0}$, $i\in \{1,2,\dots,\ell_{v+1}\}$ that $\rel{\ell}{\theta}{A}(x)=\reli{\ell}{L}{\theta}{A}(x)$ and
        \begin{equation}\label{realization multi2}
        \begin{split}
					\relii{\ell}{v+1}{\theta}{A}{i}( x ) &=\theta_{\ell_{v+1}\ell_{v}+i+\sum_{h=1}^v\ell_h(\ell_{h-1}+1)}\\ &\quad+\textstyle\sum\limits_{j=1}^{\ell_{v}}\theta_{(i-1)\ell_{v}+j+\sum_{h=1}^v\ell_h(\ell_{h-1}+1)}\bigl(x_j\indicator{\{0\}}(v) 
					+A(\relii{\ell}{v}{\theta}{A}{j})\indicator{\N}(v) 
					\bigr).
                    \end{split}
				\end{equation}
    \end{definition}
    \end{samepage}
    In \cref{definition: ANN} we note that for every $L \in \N$, $\ell \in \N^L$
we have that $\fd( \ell )$ in \cref{def: fd(ell)} is the number of real parameters (the number of real numbers) used to described \ANNs\ with the architecture vector $\ell$ and we note that for every $L \in \N$, $\ell = (\ell_1,\dots,\ell_L) \in \N^L$, $\theta \in \R^{\fd(\ell)}$ and every $A \colon \R \to \R$ we have that $\rel{\ell}{\theta}{A}\colon \R^{\ell_0}\to\R^{\ell_L}$ is the realization function of the \ANN\ with $ L + 1 $ layers and $L-1$ hidden layers (with $\ell_0 \in \N$ neurons on the input layer, $\ell_1\in\N$ neurons on the $1\textsuperscript{st}$ hidden layer, $\ell_2\in \N$ neurons on the $2\textsuperscript{nd}$ hidden layer, $\dots$,  $\ell_{ L - 1 }\in\N$ neurons on the $(L-1)\textsuperscript{th}$ hidden layer, and $\ell_L\in\N$ neurons on the output layer), with the architecture vector $\ell$, with the parameter vector $\theta$, and with the activation $A$. 
    \subsection{Non-convergence for shallow and deep ANNs with strictly positive probability}\label{subsec: main theorem 1}
    After having presented realization functions of \ANNs\ in \cref{subsec: definition ANN} above, we are now ready to formulate in the following result, \cref{main theorem1} below, the above announced special case of our main results on non-convergence of \SGD\ optimization methods \emph{with strictly positive probability}; see \cref{main cor: scientific square mean} and \cref{main cor: scientific general loss very deep} in \cref{subsec: positive probability} below.
    \begin{samepage}
    \cfclear
            \begin{athm}{theorem}{main theorem1}
                Let $L,\scrd\in \N\backslash\{1\}$, $\ell =(\ell_0,\ell_1,\dots,\ell_{L}) \in \N^{L+1}$ satisfy $\scrd=\ffd(\ell)$,  let $ ( \Omega, \mathcal{F}, \P) $ be a probability space, let
				$ a \in \R $, 
				$ b \in [a, \infty)  $, for every $ m, n \in \N_0 $ 
				let 
				$ X^m_n \colon \Omega \to [a,b]^{\ell_0} $
				and 
				$ Y^m_n \colon \Omega \to \R^{\ell_{L}} $
				be random variables, let $S\subseteq \R$ be finite, for every $r\in \N_0$ let $\mathbb A_r\in C^{\min\{r,1\}}(\R,\R)$, let $\grad\colon \R\to\R$ satisfy for all $x\in \R$ that there exists $R\in \N$ such that $ \restr{\mathbb A_0}{\R \backslash S} \in C^1(\R \backslash S,\R)$, $\restr{\grad}{\R \backslash S}=(\restr{\mathbb A_0}{\R \backslash S})'$, and
				\begin{equation}\label{main theorem1: eq1}
					\textstyle
					\sum_{r =R}^\infty\bigl(|\mathbb A_r(x)-\mathbb A_0(x)|+|(\mathbb A_r)'(x)-\grad(x)|\bigr)=0,
				\end{equation}
				 let $\alphaw \in \R$, $\betaw \in (\alphaw,\infty)$ satisfy $\sup_{ x \in ( \alphaw, \betaw )\backslash S } | \grad(x)| = 0<\sup_{x\in\R\backslash S}|\grad(x)|$,
				 let $\fL\colon \R^{\scrd}\to \R$ satisfy for all $\theta\in \R^{\scrd}$ that 
				\begin{equation}\label{main theorem1: eq2}
					\fL(\theta)=\E\bigl[\|\bfN_{\bbA_0}^{\ell,\theta}(X^0_0)-Y^0_0\|^2\bigr],
				\end{equation}
                assume $\P(\E[Y_0^0|X_0^0]=\E[Y_0^0])<1$, 
				for every $n\in \N_0$ let $ M_n \in  \N $, for every $r,n\in \N_0$ let
				$ 
				\cLnri{n}{r} \colon \R^{ \scrd } \times \Omega \to \R 
				$
				satisfy for all 
				$ \theta \in \R^{ \scrd }$
				that
				\begin{equation}\label{maintheorem1: eq3}
					\displaystyle
					\cLnri{n}{r}( \theta) 
					= 
					\frac{ 1 }{ M_n } 
					\biggl[ \textstyle
					\sum\limits_{ m = 1 }^{ M_n} 
					\|\bfN_{\bbA_r}^{\ell,\theta}(X_n^m)-Y_n^m\|^2
					\biggr]
					,
				\end{equation}
				for every $n\in \N_0$ let $\gamma_n\in \R$, let
				$ 
				\cG_n  
				\colon \R^{ \scrd} \times \Omega \to \R^{ \scrd } 
				$ 
				satisfy for all $\omega\in \Omega$, $\theta\in \{\vartheta\in \R^{\scrd}\colon (\nabla_{\vartheta} \cLnri{n}{r}(\vartheta,\omega))_{r\in \N}$ is convergent$\}$
				that
				\begin{equation}\label{main theorem1: eq4}
					\cG_n( \theta,\omega) 
					= 
					\lim_{r\to\infty}\bigl[\nabla_\theta \cLnri{n}{r}(\theta,\omega)\bigr],
				\end{equation}
				and let 
				$
				\Theta_n 
				\colon \Omega  \to \R^{\scrd }
				$
				be a random variable,  
				assume for all $n\in \N$ that
					$\Theta_{ n  } 
					=  
					\Theta_{n-1}-\gamma_{n}
					\cG_n ( \Theta_{n-1} )$,
								 and let $\sigma,\mu\in \R$ satisfy that $\sigma\Theta_0+\mu$ is standard normal   \cfload.
				Then 
				\begin{equation}\label{main theorem1: conclude}
					\textstyle
					\P\Bigl(\inf\limits_{ n\in \N_0 }  \fL(\NNelll_n)>\inf\limits_{\theta\in \R^{\scrd}}\fL(\theta)\Bigr)>0 \ifnocf.
				\end{equation}
                \cfout[.]
            \end{athm}
            \end{samepage}
            \cref{main theorem1} is a direct consequence of the more general non-convergence result in \cref{main cor: scientific square mean self-contained} in \cref{subsec: positive probability} below. \cref{main cor: scientific square mean self-contained}, in turn, follows directly from \cref{main cor: scientific square mean}, which is one of the main results of this work.
            
            In \cref{main theorem1} the random variables $X^m_n \colon\Omega\to[a,b]^{\ell_0}$, $(m,n)\in (\N_0)^2$, describe the input data of the considered \ANN\ learning problem and the random variables $Y^{m }_{ n }\colon\Omega\to \R^{\ell_L}$, $(m,n)\in (\N_0)^2$, describe the associated output data of the considered \ANN\ learning problem.
\newcommand{\leakycons}{\lambda}
The function $\bbA_0 \colon \R\to\R$ describes the activation function of the considered class of realization functions of \ANNs. The activation function $\bbA_0 \colon \R \to \R$ may be not differentiable as, for example, the popular \ReLU\ activation function 
\begin{equation}\label{eq: def relu}
  \R \ni x \mapsto \max\{x,0\} \in \R
\end{equation}
or the clipping activation function $\R\ni x\mapsto \max\{u,\min\{x,v\}\}\in \R$
with clipping parameters $u\in\R$, $v\in (u,\infty)$ (cf., \eg, \cite[Subsections 1.2.3 and 1.2.4]{ArBePhi2024}). 
If the activation function $\bbA_0$ is not differentiable, then the realization functions of the associated \ANNs\ (see \cref{realization multi2} above) and the empirical risk functions in \cref{maintheorem1: eq3} are typically also not differentiable. The empirical risk functions are, in turn, used to specify the \SGD\ optimization process $( \Theta_n )_{ n \in \N_0 }$ (see below \cref{main theorem1: eq4}). However, as the empirical risk functions are, in general, not differentiable, standard gradients of the empirical risk functions can not be employed in the \SGD\ optimization process (as they do, in general, not exist) but instead suitably generalized gradients (see \cref{main theorem1: eq4} in \cref{main theorem1}) are used to recursively determine the \SGD\ optimization process. The generalized gradients are obtained as limits of standard gradients of regularized empirical risks with the continuously differentiable activation approximations $\bbA_r \colon \R \to \R$, $r \in \N$, of the possibly non-smooth activation function $\bbA_0 \colon \R\to \R$ (see \cref{main theorem1: eq1} in \cref{main theorem1}). 
With this approximation procedure one can describe in a concise way precisely those generalized gradients that are used in the common automatic differentiation frameworks {\sc Pytorch} and {\sc TensorFlow} in practical implementations. For more details on this \emph{approximation approach to describe generalized gradients in deep learning} we refer, \eg, to \cite{Dereichmathematical2024} (cf., \eg, also \cite{BolPau,BP2021,Cheriditoconvegence2021,MaArconvergenceproof,ArAdconvergence2021}).

 We also observe that the assumption that $ \sup_{ x \in \R \backslash S } | \grad(x) | > 0$ in \cref{main theorem1} is equivalent to the condition that the activation function $\bbA_0 \colon \R \to \R$ in \cref{main theorem1} is \emph{not a constant function}. More formally, we note that the assumption that $\sup_{ x \in \R \backslash S } | \grad(x) | > 0$ in \cref{main theorem1} is equivalent to the condition that
\begin{equation}\label{eq: NR3}
  \sup_{ x \in \R } | \bbA_0(x) - \bbA_0(0) | > 0 .
\end{equation}
Besides the \ReLU\ activation in \cref{eq: def relu} and the clipping activation function also, \eg, the \RePU\ activation function
\begin{equation}\label{eq: def RePU}
  \R \ni x \mapsto (\max\{x,0\})^p \in \R
\end{equation}
 with power parameter $p \in \N \backslash \{ 1 \}$ (cf., \eg, \cite[Subsection 1.2.13]{ArBePhi2024}) satisfies the assumptions of \cref{main theorem1}. In the case of 
\cref{eq: def RePU} where $\bbA_0 = ( \R \ni x\mapsto (\max\{x,0\})^p\allowbreak \in \R )$ and where $h=(\bbA_0)'$ we have that $\bbA_0$ is continuously differentiable and, therefore, we obtain that the empirical risk functions in \cref{maintheorem1: eq3} are continuously differentiable in the sense that for all $n \in \N_0$, $\omega \in \Omega$ it holds that $\R^{\scrd}\ni \theta \mapsto \cL_n^0( \theta, \omega ) \in \R$ is continuously differentiable. This, in turn, assures that the generalized gradients of the empirical risk functions in \cref{main theorem1: eq4} coincide with the standard gradients of the empirical risk functions in the sense that for all $n\in \N_0$, $\theta\in \R^\scrd$, $\omega\in \Omega$ it holds that $\cG_n(\theta,\omega)=\nabla_\theta \cLnri{n}{0}(\theta,\omega)$ (see \cite[Theorem 1.1]{Dereichmathematical2024}).

We note that \cref{main theorem1}, in particular, implies that the probability that the true risk in \cref{main theorem1: eq2} of the \SGD\ process $( \Theta_n )_{ n \in \N_0 }$ does \emph{not} converge to the optimal/infimal value of the true risk $\inf_{ \theta \in \R^{\fd(\ell)} } \fL( \theta )$ is strictly positive. In particular, we note that \cref{main theorem1: conclude} in \cref{main theorem1} implies that
\begin{equation}\label{NR}
  \P\biggl( \liminf_{ n \to \infty } \fL( \Theta_n ) = \inf_{ \theta\in \R^\scrd} \fL(\theta) \biggr) < 1 .
\end{equation} 
Furthermore, we observe that \cref{main theorem1} demonstrates that it does \emph{not} hold that the true risk in \cref{main theorem1: eq2} of the \SGD\ process $(\Theta_n )_{ n \in \N_0 }$ converges in probability to the optimal/infimal true risk value $\inf_{ \theta \in \R^{\scrd}}  \fL( \theta )$. More formally, we \nobs that \cref{main theorem1: conclude} in \cref{main theorem1} implies that for all $\delta \in (0,\infty]$ it holds that  
\begin{equation}\label{NR1}
   \liminf_{n\to\infty}\E\Bigl[\textstyle\min\Bigl\{\delta, | \fL( \Theta_n ) - \inf\limits_{ \theta\in \R^\scrd} \fL(\theta)|\Bigr\}\Bigr] \geq \inf\limits_{ n \in \N_0 } \textstyle\E\Bigl[ \min\ \!\Bigl\{ \delta, \fL( \Theta_n ) - \inf\limits_{ \theta \in \R^{\scrd} } \fL( \theta ) \Bigr\} \Bigr] > 0
\end{equation}
(cf.\ \cref{lem: equivalent high probability} in \cref{subsec: positive probability} below).
To keep \cref{main theorem1} within this introductory section as short as possible, in \cref{main theorem1: conclude} and \cref{main theorem1} above, respectively, we restrict ourselves to the situation of a simple (standard) normal random initialization (see above \cref{main theorem1: conclude}), we restrict ourselves to just the plain vanilla standard \SGD\ optimization instead of a general class of optimizers (see below \cref{main theorem1: eq4}), and we restrict ourselves to the standard mean squared error loss in the true risks in \cref{main theorem1: eq2} and the empirical risks in \cref{maintheorem1: eq3}, respectively. We refer to \cref{main cor: scientific square mean} and \cref{main cor: scientific general loss very deep} in \cref{subsec: positive probability} for generalized and extended variants of \cref{main theorem1} in which also a large class of \SGD\ methods, initializations, and loss functions are covered.

            \subsection{Non-convergence for very deep ANNs with high probability}\label{subsec: main theorem 2}

            \cref{main theorem1} and its generalizations and extensions in \cref{main cor: scientific square mean} and \cref{main cor: scientific general loss very deep} below, respectively, offer \emph{rather general results} that apply to realistic practical relevant \ANN\ training scenarios with general input-output data, with general deep fully-connected feedforward \ANNs\ with basically no restriction on the deep \ANN\ architecture, with a general class of activation functions, and with a general class of \SGD\ optimization methods covering many of the most popular optimizers. 
            
            However, it should also be pointed out that  \cref{main theorem1} and its extensions in \cref{main cor: scientific square mean} and \cref{main cor: scientific general loss very deep}, respectively, offer only a \emph{rather weak conclusion} in the sense that \cref{main theorem1} just ensures that the probability for non-convergence to the optimal true risk value is strictly positive but \cref{main theorem1} does \emph{not} provide us insights how large this probability actually is. In the special situation where the considered \ANNs\ are in a certain sense very deep \ANNs\ one can strengthen the conclusion of \cref{main theorem1} in the sense that this non-convergence to the optimal/infimal true risk value can then be shown to happen with a high probability that converges to one. This is precisely the subject of the next result.
            \begin{samepage}
             \cfclear
            \begin{athm}{theorem}{main theorem2}
                      Let $d,\delta\in \N$, for every $k\in \N_0$ let $l_k,L_k\in \N\backslash\{1\}$, $\xell{k} =(\xxell{0}{k},\xxell{1}{k},\dots,\xxell{L_k}{k}) \in \{d\}\times \{1,2,\dots,l_k\}^{L_k-1}\allowbreak\times \{\delta\}$, let $ ( \Omega, \mathcal{F}, \P) $ be a probability space, let
				$ a \in \R $, 
				$ b \in [a, \infty)  $, for every $ m, n \in \N_0 $ 
				let 
				$ X^m_{n} \colon \Omega \to [a,b]^{d} $
				and 
				$ Y^m_{n} \colon \Omega \to \R^{\delta} $
				be random variables, let $S\subseteq \R$ be finite, for every $r\in \N_0$ let $\mathbb A_r\in C^{\min\{r,1\}}(\R,\R)$, let $\grad\colon \R\to\R$ satisfy for all $x\in \R$ that there exists $R\in \N$ such that $ \restr{\mathbb A_0}{\R \backslash S} \in C^1(\R \backslash S,\R)$, $\restr{\grad}{\R \backslash S}=(\restr{\mathbb A_0}{\R \backslash S})'$, and
				\begin{equation}\label{main theorem2: eq1}
					\textstyle
					\sum_{r =R}^\infty\bigl(|\mathbb A_r(x)-\mathbb A_0(x)|+|(\mathbb A_r)'(x)-\grad(x)|\bigr)=0,
				\end{equation}
				 let  $\betaw\in \R$ satisfy $\sup_{ x \in ( -\infty, \betaw )\backslash S } | \grad(x)| = 0<\sup_{x\in\R\backslash S}|\grad(x)|$, 
				for every $k\in \N_0$ let $\fL_k\colon \R^{\ffd(\xell{k})}\to \R$ satisfy for all $\theta\in \R^{\ffd(\xell{k})}$ that 
				\begin{equation}\label{main theorem2: eq2}
			\fL_k(\theta)=\E\bigl[\|\rel{\xell{k}}{\theta}{\bbA_0}(X^0_{0})-Y^0_{0}\|^2\bigr],
				\end{equation}
                assume $\P(\E[Y_{0}^0|X_{0}^0]=\E[Y_{0}^0])<1$,
				for every $k,n\in \N_0$ let $ M_n ^k\in  \N $, for every $r,k,n\in \N_0$ let
				$ 
				\cL_{k,n}^r \colon \R^{ \ffd(\xell{k}) } \times \Omega \to \R 
				$
				satisfy for all 
				$ \theta \in \R^{ \ffd(\xell{k}) }$
				that
				\begin{equation}\label{main theorem2: eq3}
					\displaystyle
					\cL_{k,n}^r( \theta) 
					= 
					\frac{ 1 }{ M_n^k } 
					\biggl[ \textstyle
					\sum\limits_{ m = 1 }^{ M_n^k} 
					\|\rel{\xell{k}}{\theta}{\bbA_r}(X_{n}^m)-Y_{n}^m\|^2
					\biggr]
					,
				\end{equation}
				for every $k,n\in \N_0$ let $\gamma_n^k\in \R$, let
				$ 
				\cG_n ^k 
				\colon \R^{ \ffd(\xell{k})} \times \Omega \to \R^{ \ffd(\xell{k}) } 
				$ 
				satisfy for all $\omega\in \Omega$, $\theta\in \{\vartheta\in \R^{\ffd(\xell{k})}\colon (\nabla_{\vartheta} 
                \cL_{k,n}^r(\vartheta,\omega))_{r\in \N}$ is convergent$\}$
				that
				\begin{equation}\label{main theorem2: eq4}
					\cG_n^k( \theta,\omega) 
					= 
					\lim_{r\to\infty}\bigl[\nabla_\theta \cL_{k,n}^r(\theta,\omega)\bigr],
				\end{equation}
				and let 
				$
				\Theta_n^k 
				\colon \Omega  \to \R^{\ffd(\xell{k}) }
				$
				be a random variable, 
				assume for all $k,n\in \N$ that
					$\Theta_{ n  } ^k
					=  
					\Theta_{n-1}^k-
					\gamma_{n}^k\cG_n^k ( \Theta_{n-1}^k )
					$, let $\sigma,\mu\in \R$ satisfy for all $k\in \N$ that $\sigma\Theta_0^k+\mu$ is standard normal, 
assume $\inf_{x\in \R}\bbA_0(x)\geq 0$, and assume $\limsup_{p\searrow 0}\liminf_{k\to\infty}\bigl( p^{l_k(l_k+1)}L_k\bigr)=\infty$ \cfload. Then
                \begin{equation}\label{main theorem2: conclude}
                    \textstyle
\liminf\limits_{k\to\infty}\P\Bigl(\inf\limits_{ n\in \N_0 }  \fL_k(\NNelll_n^k)>\inf\limits_{\theta\in \R^{\ffd(\xell{k})}}\fL_k(\theta)\Bigr)=1\ifnocf.
                \end{equation}
                \cfout[.]
            \end{athm}
            \end{samepage}
             \cref{main theorem2} is a direct consequence of the more general non-convergence result in \cref{main theo: scientific square mean error very deep self-contained} in \cref{subsec: high probability} below. \cref{main theo: scientific square mean error very deep self-contained}, in turn, follows directly from \cref{main theo: scientific square mean error very deep}, which is one of the main results of this work.
             
             In \cref{main theorem2} we do not consider just one fixed \ANN\ architecture as in \cref{main theorem1} but instead in \cref{main theorem2} we consider a whole sequence/family of \ANN\ architectures indexed by the non-negative integer $k \in \N_0 = \{ 0, 1, 2, \dots \}$ and \cref{main theorem2: conclude} proves that the probability that the true risk of the \SGD\ optimization process $( \Theta^k_n )_{ n \in \N_0 }$ converges to the optimal/infimal true risk value $\inf_{ \theta \in \R^{\fd(\ell_k)} }\fL_k(\theta)$ converges to zero as $k$ goes to infinity. Specifically, we observe that \cref{main theorem2: conclude} implies that
\begin{equation}
  \limsup_{ k \to \infty } \P\biggl( \liminf_{ n \to \infty } \fL_k( \Theta^k_n ) = \inf_{ \theta \in \R^{\fd(\ell_k)} }\fL_k(\theta)\biggr) = 0 .
\end{equation}
As in \cref{main theorem1} we have that \cref{main theorem2} just applies to the plain vanilla standard \SGD\ method instead of a general class of \SGD\ methods (see below \cref{main theorem2: eq4}), just applies to the standard mean square error loss in the true risks in \cref{main theorem2: eq2} and the empirical risks in \cref{main theorem2: eq3}, and just considers
simple (standard) normal random initializations (see above \cref{main theorem2: conclude}) but we refer to \cref{main theo: scientific square mean error very deep} and \cref{main theo: scientific other error very deep} in \cref{subsec: high probability} for generalized and extended variants of \cref{main theorem2} in which also a large class of \SGD\ methods, random initializations, and loss functions are covered.

\subsection{Literature review}\label{subsec: literature review}
In this subsection we provide a short review of lower bounds and (non-)convergence results for \SGD\ optimization methods related to the main findings of this work. In the special situation of the \ReLU\ activation, of the standard mean squared error loss, of the standard \SGD\ optimization method, of \ANNs\ with one-dimensional output, and where $\P\bigl( Y^0_0 = \E[ Y^0_0 | X^0_0 ] \bigr) = 1$ we refer to \cite{CHERIDITO2021101540} (cf.\ also \cite{Cheriditononconvergencev1} and \cite{Lu_2020}) for lower bounds and non-convergence results
which, in the setup of \cref{main theorem1}, essentially imply that the true risk of the \SGD\ method does not converge in probability to zero in the sense that 
\begin{equation}
 \liminf_{n\to\infty}\E[\min\{1,\fL(\Theta_n)\}] > 0
\end{equation}
(cf., \eg, \cite[Proposition 5.2]{Cheriditononconvergencev1} and \cite{Lu_2020}).
We also note that both \cref{main theorem1} and \cref{main theorem2} above assume that the conditional expectation of the output datum $Y^0_0$ given the input datum $X^0_0$ does not $\P$-almost surely coincide with the expectation of the output datum $\E[ Y^0_0 ]$ in the sense that 
\begin{equation}\label{eq: NR1}
  \P(\E[Y_0^0|X_0^0]=\E[Y_0^0])<1.
\end{equation}
This mild assumption basically postulates that the factorization of the conditional expectation of the output datum given the input datum is not almost surely (with respect to the probability distribution of the input datum) a constant function but must depend in a non-constant way on the input datum. In our opinion it is highly convincing that this assumption is satisfied in practically relevant data-driven and model-driven supervised learning problems and it should also be pointed out that the conclusions of \cref{main theorem1} and \cref{main theorem2} are simply not valid in the case where this assumption is not satisfied. Indeed, the work \cite{MaArconvergenceproof} establishes in the training of deep fully-connected feedforward \ANNs\ with an arbitrary architecture that the probability that the true risk of the standard \SGD\ optimization process converges to the optimal/infimal value of the true risk is exactly equal to one (see \cite{MaArconvergenceproof} and cf., \eg, also \cite[Proposition 4.11]{HannibalJentzenThang2024}) in the case where the target function (the factorization of the conditional expectation of the output data given the input data) is a constant function. 

We also refer to the article \cite{HannibalJentzenThang2024,ArAd2024} for results that imply that the probability to converge to the optimal empirical/true risk is not only strictly positive but even converges to one as the width of the \ANNs\ increases to infinity (the work \cite{ArAd2024} treats shallow \ANNs\ and the true risk and the work \cite{HannibalJentzenThang2024} treats deep \ANNs\ and the empirical risk). While the conclusions of the works in \cite{HannibalJentzenThang2024} and \cite{ArAd2024} is much stronger than the conclusion of \cref{main theorem1} above, the articles \cite{HannibalJentzenThang2024,ArAd2024} essentially assume that the target function of the considered supervised learning  problem cannot be represented exactly by an \ANN. Even so this seems to be a weak assumption, it is not yet clear how this condition can be verified for a concrete supervised learning problem.

We also refer, \eg, to \cite{ReddiKale2019} to a proof that the \Adam\ optimizer might converge to a suboptimal limit even in the case of convex objective functions if certain requirements on the stochastic gradients do not hold.
On a related note, \cite{Dereichnonconvergence2024} demonstrates that certain \SGD\ optimization methods including the \Adam\ optimizer fail to converge to any possible random point in the optimization space (and, in particular, to any local or global minimizer) if the learning rates do not converge to zero (cf., \eg, also \cite[Corollary 7.2.10 and Lemma 7.2.11]{ArBePhi2024}). In contrast, in \cref{main theorem1} and \cref{main theorem2} above there is no assumption on the learning rates and, in particular, the learning rates may or may not converge to zero but in both cases the non-convergence statements in \cref{main theorem1: conclude} and \cref{main theorem2: conclude} can be concluded/established.

It should be noted that the results of this work rely on the notion of inactive or \emph{dead} neurons which have a constant output over the entire input domain (cf.~\cref{def: inact} below).
The existence of such neurons is due to the use of the \ReLU\ activation function, which has the constant value zero on $(-\infty , 0 ]$.
We refer to \cite{Lu_2020,ShinKarniadakis2020} for upper and lower bounds on the probability of existence of dead neurons.
Related results were established in \cite{CHERIDITO2021101540} to prove
in the situation of the \ReLU\ activation, the standard \SGD\ optimization method, and the standard mean square error loss that the risk values do not converge to zero in probability for sufficiently deep \ANNs\ (cf.\ also \cref{lem: equivalent high probability} below).
The results of this work are somewhat stronger since they show that these risk values do not even converge in probability to the global minimum of the risk, which in many practically relevant cases is strictly positive.

The non-convergence of \SGD\ methods to global minima is closely related to the existence of non-global, sometimes called \emph{spurious} or \emph{bad} local minima.
In \cite{ChristofKowalczyk2023} it has been shown in the case of many activation functions including the \ReLU\ that such minima frequently occur in the \ANN\ optimization landscape.
Similar results for activations such as sigmoid and softplus have been established in  \cite{DingLiSun2019}.
On the other hand, if no global or local minima exist in the optimization landscape then gradient methods might even diverge as is shown, \eg, in \cite{gallon2022blowphenomenagradientdescent}. Finally, we also refer to \cite{Philiptheorytopractice2021} for further theoretical lower bounds regarding the rate of convergence of certain optimization methods for the training of \ANNs.

\subsection{Structure of this article}

The remainder of this work is organized as follows. In the main result of \cref{sec: general analysis}, \cref{main theorem general} in \cref{subsec: general analysis final}, we establish a general lower bound for the probability that the true/empirical risk (cf.\ \cref{def: fL2} in \cref{setting: SGD2}) of a general \SGD\ optimization process (cf.\ \cref{recurrent2} in \cref{setting: SGD2}) does \emph{not} converges to the optimal/infimal value of the true/empirical risk (cf.\ \cref{conclude: most general} in \cref{main theorem general}) under the abstract assumption that \ANNs\ with constant realization functions can not reach the optimal/infimal value of the true/empirical risk (cf.\ the second line of \cref{main theorem general}). In the main result of \cref{sec: improve risk}, \cref{verify maintheorem 5} in \cref{sucsec: improve risk for deep ANN} (which is a key new contribution of this work), we provide a concrete sufficient condition (cf.\ \cref{assume1: a} in \cref{verify maintheorem 5}) which ensures that this abstract assumption, that \ANNs\ with constant realization functions can not reach the optimal/infimal value of the risk, is fulfilled (cf.\ \cref{conclude: main improve risk} in \cref{verify maintheorem 5}). In \cref{sec: application} we combine \cref{main theorem general} from \cref{sec: general analysis} with \cref{verify maintheorem 5} from \cref{sec: improve risk} to establish in \cref{main cor: scientific general square} in \cref{subsec: lower bound} a general lower bound for the probability that the true risk (cf.\ \cref{loss: scientific} in \cref{setting: scientific learning}) of a general \SGD\ optimization process (cf.\ \cref{SGD: scientific} in \cref{setting: scientific learning}) does \emph{not} converge to the optimal/infimal true risk value (cf.\ \cref{conclude: estimate1} in \cref{main cor: scientific general square}). Furthermore, in \cref{sec: application} we apply \cref{main cor: scientific general square} to establish the concrete non-convergence results in 
\begin{enumerate}[label=(\roman*)]
    \item 
\cref{main cor: scientific square mean} (non-convergence with strictly positive probability for the standard mean squared error loss; see \cref{def: bbLa} and \cref{conclude: squaremean1} in \cref{main cor: scientific square mean}),
\item \cref{main cor: scientific general loss very deep} (non-convergence with strictly positive probability for a class of general loss functions; see \cref{assume1a} and \cref{conclude: other loss1} in \cref{main cor: scientific general loss very deep}),
\item \cref{main theo: scientific square mean error very deep} (non-convergence with high probability for the standard mean squared error loss; see \cref{conclude: mean square loss2} in \cref{main theo: scientific square mean error very deep}), and 
\item \cref{main theo: scientific other error very deep} (non-convergence with high probability for a class of general loss functions; see \cref{def: bbLaa}, \cref{assume1aa}, and \cref{concludeaaa} in \cref{main theo: scientific other error very deep}).
\end{enumerate}
\cref{main theorem1} and \cref{main theorem2} in this introductory section are direct consequences of \cref{main cor: scientific square mean} and \cref{main theo: scientific square mean error very deep}, respectively.

			\section{Abstract lower bounds for the non-convergence probability}\label{sec: general analysis}
            In the main result of this section, \cref{main theorem general} in \cref{subsec: general analysis final} below, we establish a general lower bound for the probability that the true/empirical risk (cf.\ \cref{def: fL2} in \cref{setting: SGD2}) of a general \SGD\ optimization process (cf.\ \cref{recurrent2} in \cref{setting: SGD2}) does \emph{not} converge to the optimal/infimal value of the true/empirical risk (cf.\ \cref{conclude: most general} in \cref{main theorem general}) under the abstract assumption that \ANNs\ with constant realization functions can not reach the optimal/infimal value of the true/empirical risk (cf.\ the second line of \cref{main theorem general}). The arguments in this section are, \eg, inspired by the arguments in \cite[Section 5]{HannibalJentzenThang2024}, \cite[Section 4]{ArAd2024}, and \cite[Section 3]{CHERIDITO2021101540}.
            
			\subsection{Mathematical framework for the training of deep ANNs}\label{sucsec: setting general analysis}
            \cfclear
			\begin{setting}\label{setting: SGD2}
				Let $L\in \N$, $\ell=(\ell_0,\ell_1,\dots,\ell_L)\in \N^{L+1}$, 
				$ a \in \R $, 
				$ b \in [a, \infty)  $, for every $ m, n \in \N_0 $ 
				let 
				$ X^m_n \colon \Omega \to [a,b]^{\ell_0} $
				and 
				$ Y^m_n \colon \Omega \to \R^{\ell_{L}} $
				be random variables, let $S\subseteq \R$ be finite, for every $r\in \N_0$ let $\mathbb A_r\in C^{\min\{r,1\}}(\R,\R)$, let $\grad\colon \R\to\R$ satisfy for all $x\in \R$ that there exists $R\in \N$ such that $ \restr{\mathbb A_0}{\R \backslash S} \in C^1(\R \backslash S,\R)$, $\restr{\grad}{\R \backslash S}=(\restr{\mathbb A_0}{\R \backslash S})'$, and
				\begin{equation}\label{def: g_rsetting}
					\textstyle
					\sum_{r =R}^\infty\bigl(|\mathbb A_r(x)-\mathbb A_0(x)|+|(\mathbb A_r)'(x)-\grad(x)|\bigr)=0,
				\end{equation}
				let $\betaw \in \R$, $\alphaw \in [-\infty,\betaw)$ satisfy $\sup_{ x \in ( \alphaw, \betaw )\backslash S } | \grad(x)| = 0$, 
				let $\smalll\colon \R^{ \ell_L } \times \R^{ \ell_L }\to\R$ be measurable, let $\mu\colon \mathcal B([a,b]^{\ell_0}\times\R^{\ell_L})\times \Omega\to [0,\infty]$ satisfy for all $\omega\in \Omega$ that $\mathcal B([a,b]^{\ell_0}\times\R^{\ell_L})\ni \set\mapsto \muu{\set}{\omega}\in [0,\infty]$ is a measure, assume for all $\set\in \mathcal B([a,b]^{\ell_0}\times\R^{\ell_L})$ that $\Omega\ni \omega \mapsto \muu{\set}{\omega}\in [0,\infty]$ is measurable, let $\fL\colon \R^{\ffd(\ell)}\times \Omega\to \R$ satisfy for all $\theta\in \R^{\ffd(\ell)}$, $\omega\in \Omega$ that $ \int_{[a,b]^{\ell_0}\times \R^{\ell_L}}
				| \smalll(\rel{\ell}{\theta}{\bbA_0}(x),y)|\, \muuu {\mathrm{d}x}{\mathrm{d}y}{\omega}<\infty$ and
				\begin{equation}\label{def: fL2}
					\fL(\theta,\omega)=\textstyle 
					\int_{[a,b]^{\ell_0}\times \R^{\ell_L}}
					\smalll(\rel{\ell}{\theta}{\bbA_0}(x),y)\, \muuu{ \mathrm{d}x}{\mathrm{d}y}{\omega},
				\end{equation}
				  for every $n\in \N_0$ let $\bbL_n\in C^1(\R^{\ell_L}\times\R^{\ell_L},\R)$, $ M_n \in  \N $, for every $r,n\in \N_0$ let
				$ 
				\cLnri{n}{r} \colon \R^{ \ffd(\ell) } \times \Omega \to \R 
				$
				satisfy for all 
				$ \theta \in \R^{ \ffd(\ell) }$
				that
				\begin{equation}\label{loss2}
					\displaystyle
					\cLnri{n}{r}( \theta) 
					= 
					\frac{ 1 }{ M_n } 
					\biggl[ \textstyle
					\sum\limits_{ m = 1 }^{ M_n} 
					\bbL_n(\rel{\ell}{\theta}{\bbA_r}(X_n^m),Y_n^m)
					\biggr]
					,
				\end{equation}
				for every $n\in \N_0$ let 
				$ 
				\cG_n  
				\colon \R^{ \ffd(\ell)} \times \Omega \to \R^{ \ffd(\ell) } 
				$ 
				satisfy for all $\omega\in \Omega$, $\theta\in \{\vartheta\in \R^{\ffd(\ell)}\colon (\nabla_{\vartheta} \cLnri{n}{r}(\vartheta,\omega))_{r\in \N}$ is convergent$\}$
				that
				\begin{equation}\label{gradient2}
					\cG_n( \theta,\omega) 
					= 
					\lim_{r\to\infty}\bigl[\nabla_\theta \cLnri{n}{r}(\theta,\omega)\bigr]
				\end{equation}
				and let 
				$
				\Theta_n =(\xTheta{n}{1},\dots,\xTheta{n}{\ffd(\ell)})
				\colon \Omega  \to \R^{\ffd(\ell) }
				$
				be a random variable, for every $n\in \N$ let 
				$
				\Phi_n 
				= 
				( 
				\Phi^{ 1 }_n, \dots, 
				\Phi^{ \ffd(\ell) }_n 
				)
				\colon 
				\allowbreak
				( \R^{ \ffd(\ell)} )^{ 2n }
				\allowbreak
				\to 
				\R^{ \ffd(\ell) }
				$ 
				satisfy 
				for all 
				$
				g =
				( 
				( g_{ i, j } )_{ j \in \{ 1, 2, \dots, \ffd(\ell) \} }
				)_{
					i \in \{ 1, 2, \dots, 2n\}
				}
				\in 
				(
				\R^{ 
					\ffd(\ell)
				}
				)^{ 2n }
				$, 
				$ 
				j \in \{1,2,\dots,\ffd(\ell)\}  
				$
				with $
				\sum_{ i = 1 }^{2n}
				\abs{ g_{ i, j } -g_{1,j}\mathbbm 1_{[1,n]}(i)}
				= 0
				$
				that 
				$
				\Phi^{ j }_n( g ) = g_{n,j} 
				$ and
				assume 
				\begin{equation}
					\label{recurrent2}
					\Theta_{ n  } 
					=  
					\Phi_{n}\bigl(\Theta_0,\Theta_1,\dots,\Theta_{n-1},
					\cG_1( \Theta_0  ) ,
					\cG_2( \Theta_1  ) ,
					\dots ,
					\cG_n ( \Theta_{n-1} )
					\bigr),
				\end{equation} 
                and for every $k\in \{1,2,\dots,L\}$ let $\inact_k\subseteq \R^{\ffd(\ell)}$ satisfy
                \begin{equation}\label{def: inact}
                    \inact_k=\bigl\{\theta\in \R^{\ffd(\ell)}\colon \bigl(\forall \, i\in\{1,2,\dots,\ell_k\},\, x\in [a,b]^{\ell_0}\colon \relii{\ell}{k}{\theta}{\bbA_0}{i}(x)\in (\alphaw,\betaw)\backslash S\bigr)\bigr\} \ifnocf.
                \end{equation}
                \cfload[.]
			\end{setting}
			\subsection{Invariance of inactive neurons during the training procedure} \label{sucsec: general analysis 1}
             \cfclear
			\begin{athm}{lemma}{vanishing gradient: very deep2}
				Assume \cref{setting: SGD2} \cfload.
                Then
                it holds for all $k\in \{1,2,\dots L-1\}$, $\theta\in \inact_k$, $n\in\N_0$, $\omega\in \Omega$, $j\in \{1,2,\dots,\sum_{i=1}^k\ell_i(\ell_{i-1}+1)\}$ that
					\begin{equation}\llabel{conclude}
                    \cG^{j}_{n}(\theta,\omega)=0 \ifnocf.
                    \end{equation}
                    \cfout[.]
			\end{athm}
			\begin{aproof}
				Throughout this proof for every $f\colon \R\to\R$ and $g\colon \R\to\R$ with $f|_{\R\backslash S}\in C^1(\R\backslash S,\R)$ let $\scrd_g f\colon \R\to\R$ satisfy for all $x\in \R$ that
    \begin{equation}\llabel{def: scrd}
        (\scrd_\grad f)(x)=\begin{cases}
            f'(x)&\colon x\in \R\backslash S\\
            g(x)&\colon x\in S,
        \end{cases}
    \end{equation} 
    for every $m,n\in \N_0$ let $\bfX_n^m=(\bfX_n^{m,1},\dots,\bfX_n^{m,\ell_0})\colon\Omega\allowbreak\to \R^{\ell_0}$ satisfy $\bfX_n^m=X_n^m$, for every $\theta\in \ffd(\ell)$ let $\scrN^{\ell,\theta}_{\bbA_0}=(\rell{\ell}{\theta}{\bbA_0}{1},\dots,\rell{\ell}{\theta}{\bbA_0}{v_L})\colon \R^{\ell_0}\to \R^{\ell_L}$ satisfy for all $x\in \R^{\ell_0}$ that $\scrN^{\ell,\theta}_{\bbA_0}(x)=\rel{\ell}{\theta}{\bbA_0}(x)$, for every $k\in\{1,2,\dots,L\}$, $\theta=(\theta_1,\dots,\theta_{\ffd(\ell)})\in \R^{\ffd(\ell)}$, $i\in \{1,2,\dots,\ell_k\}$
				let 
				$  
				\fb_i^{ k, \theta }, \fw^{ k, \theta }_{ i, 1 } ,\fw^{ k, \theta }_{ i, 2 },\dots, \fw^{ k, \theta }_{ i, \ell_{k-1} }
				\in \R $ 
				satisfy for all 
				$ j \in \{ 1,2, \ldots, \ell_{ k - 1 } \} $ 
				that
				\begin{equation}
					\llabel{wb}
					\fb^{ k, \theta }_i 
					=
					\theta_{ \ell_k \ell_{ k - 1 } + i 
						+ 
						\sum_{h=1}^{k-1}\ell_h(\ell_{h-1}+1)}
					\qqandqq
					\fw^{ k, \theta }_{ i, j }
					= 
					\theta_{ ( i - 1 ) \ell_{ k - 1 } + j 
						+ 
						\sum_{h=1}^{k-1}\ell_h(\ell_{h-1}+1) },
				\end{equation}
				and for every $n\in \N_0$, $\omega\in \Omega$ let $\nu_{n,\omega}\colon \mathcal B([a,b]^{\ell_0}\times\R)\to [0,\infty]$ satisfy for all $\set\in \mathcal B([a,b]^{\ell_0}\times\R)$ that 
				\begin{equation}\llabel{def: mu}
					\nu_{n,\omega}(\set)=\frac{1}{M_n^\ell}\biggl[\textstyle\sum\limits_{m=1}^{M_n^\ell}\mathbbm 1_{A}\bigl((\bfX_n^m(\omega),Y_n^m(\omega))\bigr)\biggr]\ifnocf.
				\end{equation}
                \cfload[.]
				\argument{\cref{loss2};\lref{def: mu}}{that for all $r\in \N_0$, $\theta\in \R^{\ffd(\ell)}$, $n\in \N_0$, $\omega\in \Omega$ it holds that
					\begin{equation}\llabel{eq0.5}
						\cLnri{n}{r}(\theta,\omega)=\int_{[a,b]^{\ell_0}\times\R^{\ell_L}}\bbL_n(\rel{\ell}{\theta}{\bbA_r}(x)-y)\,\nu_{n,\omega}(\d x,\d y)\dott
				\end{equation}}
                				\argument{the assumption that for all $R\in \N$, $\omega\in \Omega$ it holds that 
                                \begin{equation}
                                \sup_{r \in \N}\sup_{x\in[-R,R]}\allowbreak\bigl[(|\bbA_r(x)|+|(\bbA_r)'(x)|)\mathbbm 1_{\{\infty\}}(|\mathrm{supp}(\mu(\omega))|)\bigr]<\infty;
                                \end{equation}
                                \cite [items (ii) and (iii) in Proposition 2.14]{Dereichmathematical2024} (applied for every $n\in \N_0$, $k\in \{1,2,\dots,L\}$, $\omega\in \Omega$ with $(\ell_k)_{k\in \{0,1,\dots,L\}}\curvearrowleft (\ell_k)_{k\in \{0,1,\dots,L\}}$, $L\curvearrowleft L$, $d\curvearrowleft\ffd(\ell)$,
					$((\fb^{k,\theta}_i)_{ i\in\{1,2,\dots,\ell_k\}}\allowbreak)_{(k,\theta)\in \{1,2,\dots,L\}\times \R^{\ffd(\ell)}}\allowbreak\curvearrowleft\allowbreak(\allowbreak(\fb^{k,\theta}_i\allowbreak)_{ i\in\{1,2,\dots,\ell_k\}}\allowbreak)_{(k,\theta)\in \{1,2,\dots,L\}\times \R^{\ffd(\ell)}}$, $((\fw_{i,j}^{k,\theta}\allowbreak)_{(i,j)\in \{1,2,\dots,\ell_k\}\times\{1,2,\dots,\ell_{k-1}\}}\allowbreak\allowbreak)_{(k,\theta)\in \{1,2,\dots,L\}\times \R^{\ffd(\ell)}}\allowbreak\curvearrowleft \allowbreak((\fw_{i,j}^{k,\theta}\allowbreak\allowbreak)_{(i,j)\in \{1,2,\dots,\ell_k\}\times\{1,2,\dots,\ell_{k-1}\}}\allowbreak)_{(k,\theta)\in \{1,2,\dots,L\}\times \R^{\ffd(\ell)}}$, $\{y_1,y_2,\dots,y_r\}\curvearrowleft S$, 
					$(A_n)_{n\in \N_0}\curvearrowleft (\mathbb A_r)_{r\in \N_0},
					\allowbreak(\mN^{k,\vartheta}_n)_{(n,\vartheta,k)\in \N_0\times \R^{d}\times \{0,1,\dots,L\}}\allowbreak\curvearrowleft (\reli{\ell}{k}{\vartheta}{\bbA_r})_{(r,\vartheta,k)\in \N_0\times \R^{\ffd(\ell)}\times \{0,1,\dots,L\}}$, 
					$\mu\curvearrowleft \nu_
					{n,\omega}$, $H\curvearrowleft (\R^{\ell_L}\times\R^{\ffd(\ell)}\times \R^{\ell_L}\ni (x,\theta,y)\mapsto \bbL_n(x,y)\in \R)$, $(\cL_n)_{n\in \N_0}\curvearrowleft (\cL_r)_{r\in \N_0}$,
					$\cG\curvearrowleft \cG_n$ in the notation of \cite [Propposition 2.14]{Dereichmathematical2024});\lref{eq0.5};\lref{def: scrd};\lref{def: mu}}{that for all $n\in \N_0$, $k\in \{1,2,\dots,L\}$,  $\theta\in \R^{\ffd(\ell)}$,
					$ i \in \{ 1,2, \ldots, \ell_{k} \} $, 
					$ j \in \{ 1,2, \ldots, \ell_{k-1} \} $, $\omega\in \Omega$
					it holds that
					\begin{align}
							&\cG^{ 
								( i - 1 ) \ell_{k-1} +j+\sum_{h=1}^{k-1}\ell_h(\ell_{h-1}+1)
							}_n( \theta,\omega ) 
							\\&=\frac{1}{M_n}\Biggl[
							\textstyle \sum\limits_{m=1}^{M_n}\biggl[\sum\limits_{g=1}^{\ell_L}
							\sum\limits_{ 
								\substack{
									v_k,v_{k+1}, \ldots, v_L \in \N, 
									\\
									\forall \, w \in \N \cap [k,L] \colon 
									v_w \leq \ell_w
								}
							}
							2 \Bigl(\bigl(\mathbb A_0\bigl(
							\relii{\ell}{k-1}{\theta}{\bbA_0}{j}( \bfX_n^m (\omega))\bigr)\bigr)\mathbbm 1_{(1,L]}(k)+\bfX_n^{m,j}(\omega)\mathbbm 1_{\{1\}}(k)\Bigr)\nonumber\\ 
							&\cdot \mathbbm 1_{\{i\}}(v_k)\textstyle(\frac{\partial}{\partial x_g}\bbL_n)\bigl(
					\rell{\ell}{\theta}{\bbA_0}{v_{L}}( \bfX^m_n(\omega) ) 
							,
							Y^m_n(\omega) 
							\bigr)
							\biggl(
							\textstyle
							\prod\limits_{ p= k+1}^L
							\bigl[\fw^{ p, \theta }_{ v_p, v_{ p - 1 } }
							(\scrd_h \bbA_0)
							\bigl(\relii{\ell}{p-1}{\theta}{\bbA_0}{v_{p-1}} (\bfX_n^m (\omega))\bigr)
							\bigr] \biggr)
							\biggr]\Biggr] \nonumber
					\end{align}
					and
					\begin{equation}
						\llabel{G_b}
						\begin{split}
							& \cG^{ 
								\ell_{k}\ell_{k-1}+i+\sum_{h=1}^{k-1}\ell_h(\ell_{h-1}+1)
							}_n( \theta,\omega )\\
							&
							=\frac{1}{M_n}\Biggl[
							\textstyle \sum\limits_{m=1}^{M_n}\biggl[
							\sum\limits_{g=1}^{\ell_L}\sum\limits_{ 
								\substack{
									v_k,v_{k+1}, \ldots, v_L \in \N, 
									\\
									\forall \, w \in \N \cap [k,L] \colon 
									v_w \leq \ell_w
								}
							}
							2
							\indicator{ \{ i \} }( v_k )\indicator{\{g\}}(v_L)
							(\frac{\partial}{\partial x_g}\bbL_n)\bigl(\rell{\ell}{\theta}{\bbA_0}{v_{L}}
							( \bfX^m_n(\omega) ),
							Y^m_n (\omega)
							\bigr)
							\\
							& \quad\cdot
							\biggl(
							\textstyle
							\prod\limits_{ p= k+1}^L
							\bigl[\fw^{ p, \theta }_{ v_p, v_{ p- 1 } }
							(\scrd_h \mathbb A_0)
							\bigl(\relii{\ell}{p-1}{\theta}{\bbA_0}{v_{p-1}}(\bfX_n^m (\omega))\bigr)
							\bigr]\biggr)
							\biggr]\Biggr].
						\end{split}
				\end{equation}}
				\argument{\cref{def: inact};}{for all $k\in \{1,2,\dots,L-1\}$,  $i\in \{1,2,\dots,\ell_k\}$, $\theta\in \cJ_{k,i}$, $x\in[a,b]^{\ell_{0}}$ that
					\begin{equation}\llabel{eq1}
						\relii{\ell}{k}{\theta}{\bbA_0}{i}(x)\in (\alphaw,\betaw)\backslash S\dott
				\end{equation}}
				\argument{the assumption that $\sup_{ x \in ( \alphaw, \betaw )\backslash S } | \mathbb (\bbA_0)'(x)| = 0$; the assumption that $S$ is a finite set}{that for all $x\in (\alphaw,\betaw)\backslash S$ it holds that
                \begin{equation}\llabel{argt1}
                    (\scrd_h\bbA_0)(x)=0.
                \end{equation}}
				\argument{\lref{eq1};\lref{argt1}}{for all $k\in \{1,2,\dots,L-1\}$, $i\in \{1,2,\dots,\ell_k\}$, $\theta\in \cJ_{k,i}$, $x\in [a,b]^{\ell_0}$ that
					\begin{equation}\llabel{eq3}
						(\scrd_h \bbA_0)(\relii{\ell}{k}{\theta}{\bbA_0}{i}(x))=0\dott
				\end{equation}}
				\argument{\lref{eq3};\cref{def: inact}}{for all $k\in \{1,2,\dots,L-1\}$, $\theta\in \inact_k$, $l\in \{1,2,\dots,k\}$, $v_l\in \{1,2,\dots, \ell_l\}$, $v_{l+1}\in \{1,2,\dots,\ell_{l+1}\}$, $\dots$, $v_L\in \{1,2,\dots,\ell_L\}$, $x\in [a,b]^{\ell_0}$ that
					\begin{equation}\llabel{eq3'}
						\begin{split}
							\textstyle
							& \prod\limits_{ p = l + 1 }^{L}
							(\scrd_{\grad}\bbA_0)(\relii{\ell}{p-1}{\theta}{\bbA_0}{v_{p-1}}( x ))\\
							&=(\scrd_{\grad}\bbA_0)(\relii{\ell}{k}{\theta}{\bbA_0}{v_k}( x ))\biggl(\textstyle\prod\limits_{p\in \{l+1,l+2,\dots,L\}\backslash\{k+1\}}(\scrd_{\grad}\bbA_0)(\relii{\ell}{p-1}{\theta}{\bbA_0}{v_{p-1}}( x ))\biggr)=0.
						\end{split}
				\end{equation}}
				\argument{\lref{eq3'};}{for all $k\in \{1,2,\dots,L-1\}$, $\theta\in \inact_k$, $l\in\{1,2,\dots,k\}$, $v_l\in \{1,2,\dots,\ell_l\}$, $v_{l+1}\in\{1,2,\dots,\ell_{l+1}\}$, $\dots$, $v_{L}\in \{1,2,\dots,\ell_{L}\}$, $x\in [a,b]^{\ell_0}$ that
					\begin{equation}\llabel{eq3''}
						\begin{split}
							\textstyle
							&\prod\limits_{ p = l + 1 }^{L}
							\bigl(
							\fw^{ p,\theta }_{ v_p, v_{ p - 1 } }
							(\scrd_{\grad}\bbA_0)(\relii{\ell}{p-1}{\theta}{\bbA_0}{v_{p-1}}( x ))
							\bigr)\\
							&=\textstyle\biggl( \prod\limits_{ p = l + 1 }^{L}
							\fw^{ p,\theta }_{ v_p, v_{ p - 1 } }\biggr)\biggl(\prod\limits_{ p = l + 1 }^{L}(\scrd_{\grad}\bbA_0)(\relii{\ell}{p-1}{\theta}{\bbA_0}{v_{p-1}}( x ))\biggr)=0\dott
						\end{split}
				\end{equation}}
				\argument{\lref{eq3''};\lref{G_b};the fact that for all $m,n\in\N$, $\omega\in \Omega$ it holds that $\bfX_n^m(\omega)\in [a,b]^{\ell_0}$}{that for all $n\in \N_0$, $k\in \{1,2,\dots,L-1\}$,
					$\theta\in \inact_k$, $l\in \{1,2,\dots,k\}$, $ i \in \{1,2,\dots,\ell_{l}\} $, 
					$ j \in \{ 1,2, \ldots, \ell_{l-1} \} $, $\omega\in \Omega$
					it holds that
					\begin{align}
							&\cG^{ 
								( i - 1 ) \ell_{l-1} +j+\sum_{h=1}^{l-1}\ell_h(\ell_{h-1}+1)
							}_n( \theta,\omega ) \nonumber
							\\&=\frac{1}{M_n}\Biggl[
							\textstyle \sum\limits_{m=1}^{M_n}\biggl[\sum\limits_{g=1}^{\ell_L}
							\sum\limits_{ 
								\substack{
									v_l,v_{l+1}, \ldots, v_L \in \N, 
									\\
									\forall \, w \in \N \cap [l,L] \colon 
									v_w \leq \ell_w
								}
							}
							2 \Bigl(\bigl(\mathbb A_0\bigl(
							\relii{\ell}{l-1}{\theta}{\bbA_0}{j}( \bfX_n^m(\omega))\bigr)\bigr)\mathbbm 1_{(1,L]}(l)+\bfX_n^{m,j}(\omega)\mathbbm 1_{\{1\}}(l)\Bigr)\mathbbm 1_{\{i\}}(v_l) \nonumber\\
							&\quad \cdot 0\textstyle(\frac{\partial}{\partial x_g}\bbL_n)\bigl(
							\rell{\ell}{\theta}{\bbA_0}{v_{L}}( \bfX^m_n(\omega) ) 
							,
							Y^m_n(\omega) 
							\bigr)
							\biggr]\Biggr] =0
					\end{align}
					and
					\begin{equation}
						\llabel{eq5}
						\begin{split}
							& \cG^{ 
								\ell_{l}\ell_{l-1}+i+\sum_{h=1}^{l-1}\ell_h(\ell_{h-1}+1)
							}_n( \theta,\omega )\\
							&
							=\frac{1}{M_n}\Biggl[
							\textstyle \sum\limits_{m=1}^{M_n}\biggl[
							\sum\limits_{g=1}^{\ell_L}\sum\limits_{ 
								\substack{
									v_l,v_{l+1}, \ldots, v_L \in \N, 
									\\
									\forall \, w \in \N \cap [l,L] \colon 
									v_w \leq \ell_w
								}
							}
							2
							\indicator{ \{ i \} }( v_1 )\indicator{\{g\}}(v_L)0
							(\frac{\partial}{\partial x_g}\bbL_n)\bigl(\rell{\ell}{\theta}{\bbA_0}{v_L}
							( \bfX^m_n(\omega) ),
							Y^m_n (\omega)
							\bigr)
							\biggr]\Biggr]\\
							&=0.
						\end{split}
				\end{equation}}
				\argument{\lref{eq5};}{\lref{conclude}\dott}
			\end{aproof}
             \cfclear
			\begin{athm}{cor}{lem: estimate proba of nonconvergence: part 12}[Inactive neurons]
				Assume \cref{setting: SGD2} and let $k\in \{1,2,\dots,L-1\}$ \cfload. Then
				\begin{equation}\llabel{conclude}
					\{\omega\in \Omega\colon\Theta_0(\omega)\in \inact_k\}=
					\{\omega\in \Omega\colon (\forall\, n\in \N_0\colon \Theta_n(\omega)\in \inact_k)\}\ifnocf.
				\end{equation}
                \cfout[.]
			\end{athm}
			\begin{aproof}
            Throughout this proof for every $m\in \{1,2,\dots,L\}$ let $\bfd_m\in \N$ satisfy $\bfd_m=\sum_{i=1}^m\ell_i(\ell_{i-1}+1)$.
				\argument{\cref{vanishing gradient: very deep2};\cref{def: inact}}{that for all $m\in \N_0$, $\theta\in \inact_k$, $j\in \{1,2,\dots,\sum_{i=1}^k \ell_i(\ell_{i-1}+1)\}$, $\omega\in \Omega$ it holds that
					\begin{equation}\llabel{eqtg1}
						\cG^{j}_{m}(\theta,\omega)=0\dott
				\end{equation}}
				\argument{\lref{eqtg1};}{that for all $m\in \N$, $\theta_0,\theta_1,\dots,\theta_m\in \inact$, $j\in \{1,2,\dots,\sum_{i=1}^k \ell_i(\ell_{i-1}+1)\}$, $\omega\in \Omega$ it holds that
					\begin{equation}\llabel{NR_7}
						\sum_{v=0}^m |\cG^{j}_{v+1}(\theta_v,\omega)|=0\dott
				\end{equation}}
				\argument{the fact that for all $n\in\N$,
					$
					g =
					( 
					( g_{ i, j } )_{ j \in \{ 1, 2, \dots, \ffd(\ell) \} }
					)_{
						i \in \{ 1, 2, \dots, 2n\}
					}
					\in 
					(
					\R^{ 
						\ffd(\ell)
					}
					)^{ 2n }
					$, 
					$ 
					j \in \{1,2,\dots,\ffd(\ell)\}  
					$
					with $
					\sum_{ i = 1 }^{2n}
					\abs{ g_{ i, j } -g_{1,j}\mathbbm 1_{[1,n]}(i)}
					= 0
					$
					it holds that 
					$
					\Phi^{ j }_n( g ) = g_{n,j} 
					$;}{that for all $m\in\N$, $g=\allowbreak((g_{v,j}\allowbreak)_{j\in \{1,2,\dots,\ffd(\ell)\}}\allowbreak)_{v\in \{1,2,\dots,2m\}}\in (\R^{\ffd(\ell)})^{2m}$, $j\in \{1,2,\dots, \sum_{i=1}^k \ell_i(\ell_{i-1}+1)\}$ with $\sum_{v=1}^m|g_{v,j}-g_{1,j}|=0$ and $\sum_{v=m+1}^{2m}|g_{v,j}|=0$ it holds that
					\begin{equation}\llabel{eqtg3}
						\Phi^{j}_m(g)=g_{m,j}\dott
				\end{equation}}
				\argument{\lref{eqtg3};\lref{NR_7}}{that for all $m\in \N$, $\theta_0=(\theta_{0,1},\dots,\theta_{0,\ffd(\ell)})$, $\theta_1=(\theta_{1,1},\dots,\theta_{1,\ffd(\ell)})$, $\dots$, $\theta_{m-1}=(\theta_{m-1,1},\dots,\theta_{m-1,\ffd(\ell)})\in \inact_k$, $j\in \{1,2,\dots, \sum_{i=1}^k \ell_i(\ell_{i-1}+1)\}$, $\omega\in \Omega$ with $\sum_{i=0}^{m-1}|\theta_{i,j}-\theta_{0,j}|=0$ it holds that
					\begin{equation}\llabel{eqtg4}
						\Phi^{j}_m\bigl(\theta_0,\theta_1,\dots,\theta_{m-1},\cG_1(\theta_0,\omega), \cG_2(\theta_1,\omega),\dots, \cG_{m}(\theta_{m-1},\omega)\bigr)=\theta_{m-1,j}\dott
				\end{equation}}
				\argument{\lref{eqtg4};}{that for all $m\in \N$, $j\in \{1,2,\dots, \sum_{i=1}^k \ell_i(\ell_{i-1}+1)\}$, $\omega\in \Omega$ with $(\cup_{q=0}^{m-1} \{\Theta_q(\omega)\})\allowbreak\subseteq\inact_k$ and $\sum_{q=0}^{m-1} |\xTheta{q}{j}-\xTheta{0}{j}|=0$ it holds that
					\begin{equation}\llabel{eqtg6}
						\Phi^{ j }_m \bigl(\Theta_0,\Theta_1,\dots,\Theta_{m-1}, \cG_1 (  \Theta_0 ( \omega ), \omega ) 
						,
						\cG_2( 
						\Theta_1 ( \omega ), \omega 
						) 
						,
						\dots 
						,
						\cG_{m}( 
						\Theta_{m-1} ( \omega ), \omega 
						) \bigr)
						= \xTheta{m-1}{j} \dott
				\end{equation}}
                \argument{ \lref{eqtg6};\cref{recurrent2}} 
				{
					for all $m\in \N$,
					$j\in \{1,2,\dots,\sum_{i=1}^k \ell_i(\ell_{i-1}+1)\}$, $\omega\in\Omega$
					with 
					$
					( 
					\cup_{ q = 0 }^{m-1} \{ 
					\Theta_q ( \omega )
					\} 
					)
					\allowbreak\subseteq 
					\inact_k
					$ and $\sum_{q=0}^{m-1}|\xTheta{q}{j}(\omega)-\xTheta{0}{j}(\omega)|=0$
					that 
					\begin{equation}\llabel{arg4}
						\xTheta{ m}{j  }( \omega ) 
						=
						\xTheta{m-1}{j}( \omega ) 
						\dott
				\end{equation}}
				\argument{\lref{arg4};\cref{realization multi2};\cref{def: inact}} 
				{for all $m\in \N$, $\omega\in\Omega$
					with 
					$
					( 
					\cup_{ q = 0 }^{m-1} \{ 
					\Theta_q ( \omega )
					\} 
					)
					\subseteq 
					\inact_k
					$ and $\sum_{q=0}^{m-1} \sum_{j=1}^{\bfd_k}\allowbreak|\xTheta{q}{j}(\omega)-\xTheta{0}{j}(\omega)|=0$
					that 
					\begin{equation}\llabel{arg5}
						( 
						\cup_{ q = 0 }^{ m } \{ 
						\Theta_q ( \omega )
						\} 
						)
						\subseteq 
						\inact_k\qqandqq \textstyle\sum\limits_{q=0}^m\sum\limits_{j=1}^{\bfd_k}\allowbreak|\xTheta{q}{j}(\omega)-\xTheta{0}{j}(\omega)|=0
						\dott
				\end{equation}}
				\argument{\lref{arg5};induction} {\llabel{arg6} for all
					$m\in \N$, $\omega\in\Omega$ with 
					$ 
					\Theta_0 ( \omega ) 
					\in \inact_k
					$
					that
					$
					( 
					\cup_{ q= 0 }^m 
					\{ 
					\Theta_q ( \omega ) 
					\} 
					) 
					\subseteq 
					\inact_k
					$ and $\sum_{q=0}^m\sum_{j=1}^{\bfd_k}|\xTheta{q}{j}(\omega)-\xTheta{0}{j}(\omega)|=0$\dott}
				\argument{\lref{arg6}}{
					for all 
					$\omega\in \Omega$
					with 
					$ 
					\Theta_0 ( \omega ) 
					\in \inact_k
					$
					that 
					\begin{equation}\llabel{arg7}
						( 
						\cup_{ q= 0 }^{ \infty }
						\{ 
						\Theta_q ( \omega ) 
						\} 
						) 
						\subseteq 
						\inact_k
						\dott
				\end{equation}}
				\argument{\lref{arg7}; the fact that $\{\omega\in \Omega\colon\Theta_0(\omega)\in \inact_k\}\supseteq
					\{\omega\in \Omega\colon (\forall\, n\in \N_0\colon \Theta_n(\omega)\in \inact_k)\}$}{\lref{conclude}\dott}
			\end{aproof}
            \subsection{Non-optimal risks through inactive neurons}
             \cfclear
			\begin{athm}{lemma}{Estimate not convergence2}
				Assume \cref{setting: SGD2} and assume $\P\bigl(\inf_{\theta\in \R^{\ffd(\ell)}}\fL(\theta)<\inf_{z\in \mathcal \R^{\ell_L}}\allowbreak\int_{[a,b]^{\ell_0}\times\R^{\ell_L}}\smalll(z,y)\allowbreak\,\muu{\d x}{\d y}\bigr)\allowbreak=1$ \cfload. Then
				\begin{equation}\llabel{conclude}
					\textstyle
					\P\Bigl(\inf\limits_{ n\in \N_0 }  \cLnri{0}{0}(\NNelll_n)>\inf\limits_{\theta\in \R^{\ffd(\ell)}}\cLnri{0}{0}(\theta)\Bigr)\geq \P\bigl(\Theta_0\in (\cup_{k=1}^{L-1}\inact_k)\bigr)\ifnocf.
				\end{equation}
                \cfout[.]
			\end{athm}
			\begin{aproof}
				\argument{\cref{realization multi2};the fact that for all $x\in (\alphaw,\betaw)$ it holds that $\bbA(x)=\bbA(\alphaw)$}{that for all $k\in \{1,2,\dots,L-1\}$, $\theta\in \R^{\ffd(\ell)}$ with $\bigl(\forall\, x\in [a,b]^{\ell_0}, \, i\in \{1,2,\dots,\ell_k\}\colon \allowbreak \relii{\ell}{k}{\theta}{\bbA_0}{i}(x)\allowbreak\in (\alphaw,\betaw)\backslash S\bigr)$ it holds that
					\begin{equation}\llabel{eq1}
						\sup_{x\in [a,b]^{\ell_0}}\|\rel{\ell}{\theta}{\bbA_0}(x)-\rel{\ell}{\theta}{\bbA_0}(a,a,\dots,a)\|=0\dott
				\end{equation}}
				\argument{\lref{eq1};\cref{def: inact}}{that for all $\theta\in (\cup_{k=1}^{L-1}\inact_k)$ there exists $z\in \R^{\ell_L}$ such that
					\begin{equation}\llabel{eq2}
						\sup_{x\in [a,b]^{\ell_0}}\|\rel{\ell}{\theta}{\bbA_0}(x)-z\|=0\dott
				\end{equation}}
				\argument{\lref{eq2};the fact that for all $m\in \{1,2,\dots,M_0\}$, $\omega\in \Omega$ it holds that $X_0^m(\omega)\in [a,b]^{\ell_0}$}{that for all $\theta\in (\cup_{k=1}^{L}\inact_k)$, $\omega\in \Omega$ there exists $z\in \R^{\ell_L}$ such that for all $m\in\{1,2,\dots,M_0\}$ it holds that
					\begin{equation}\llabel{eq3}
						\rel{\ell}{\theta}{\bbA_0}(X_0^m(\omega))=z.
				\end{equation}}
				\argument{\lref{eq3};\cref{lem: estimate proba of nonconvergence: part 12}}{that for all $\omega\in \Omega$ with $\Theta_0(\omega)\in(\cup_{k=1}^{L-1}\inact_k)$ it holds that there exists $(c_n)\subseteq \R^{\ell_L}$ such that for all $n\in\N_0$, $m\in \{1,2,\dots,M_0\}$ it holds that
					\begin{equation}\llabel{eq4}
						\rel{\ell}{\Theta_n}{\bbA_0}(X_0^m(\omega))=c_n\dott
				\end{equation}}
				\argument{\lref{eq4};\cref{loss2}}{for all $\omega\in \Omega$ with $\Theta_0(\omega)\in(\cup_{k=1}^{L-1}\inact_k)$ it holds that
					\begin{equation}\llabel{eq5}
						\textstyle   \inf\limits_{n\in \N_0}\cL_0^0(\Theta_n(\omega))\geq \inf_{z\in \mathcal \R^{\ell_L}}\allowbreak\int_{[a,b]^{\ell_0}\times\R^{\ell_L}}\smalll(z,y)\allowbreak\,\muu{\d x}{\d y}\dott
				\end{equation}}
				\argument{\lref{eq5};the assumption that $\P\bigl(\inf_{\theta\in \R^{\ffd(\ell)}}\fL(\theta)<\inf_{z\in \mathcal \R^{\ell_L}}\allowbreak\int_{[a,b]^{\ell_0}\times\R^{\ell_L}}\smalll(z,y)\allowbreak\,\muu{\d x}{\d y}\bigr)=1$}{\lref{conclude}\dott}
			\end{aproof}
			\subsection{Lower bound for the probability of inactivity for the first hidden layer}\label{subsec: inactivity first hidden layer}
             \cfclear
			\begin{athm}{lemma}{estimate inactive2}
				Assume \cref{setting: SGD2}, assume that $\xTheta{0}{i}$, $i\in\{1,2,\dots,\ell_1+\ell_0\ell_1\}$, are independent, let $\eta\in \R$, $\zeta\in (\eta,\infty)$ satisfy
					\begin{equation}\llabel{def: gamma}
						(\eta,\zeta)\subseteq (\alphaw,\betaw)\backslash S,
				\end{equation}
                 and for every $i\in \{1,2,\dots,\ell_1\}$ let $\varrho_i\in \R$ satisfy
                \begin{equation}\llabel{def: varrho}
               \textstyle \varrho_i= \bigl[\P\bigl(\frac{3\eta+\zeta}{4}<\xTheta{0}{\ell_1\ell_0+i}<\frac{\eta+3\zeta}{4}\bigr)\bigr]\bigl[\prod_{j=1}^{\ell_0}\P\bigl(|\xTheta{ 0}{( i - 1 )\ell_0 + j }|<\frac{\zeta-\eta}{2\ell_0\max\{1,|a|,|b|\}}\bigr)\bigr]\ifnocf.  
                \end{equation}
                \cfload[.]
                Then 
				\begin{equation}\llabel{conclude}
					\P(\Theta_0\in \inact_1)\geq \textstyle\prod_{i=1}^{\ell_1}\varrho_i\ifnocf.
				\end{equation}
                \cfout[.]
			\end{athm}
			\begin{aproof}
			Throughout this proof for every $\theta\in \R^{\ffd(\ell)}$ let $\bfI^\theta\subseteq\R^{\ffd(\ell)}$ satisfy
				\begin{equation}\llabel{def: bfI}
					\bfI^\theta=\bigl\{ 
					i \in \{1,2,\dots,\ell_1\}
					\colon 
					\bigl(
					\forall \, x \in [a,b]^{\ell_0} \colon  
					\relii{\ell}{1}{\theta}{\bbA_0}{i}(x)
					\in (\eta,\zeta)
					\bigr)
					\bigr\}
				\end{equation}
				and for every $i\in \{1,2,\dots,\ell_1\}$  let $ \fU_{i } \subseteq \R^{\ffd(\ell)}$ satisfy 
				\begin{equation}
					\llabel{def: fU}
					\begin{split}
						&\fU_{i }=\{\theta\in \R^{\ffd(\ell)}\colon i\in \bfI^\theta\}\ifnocf.
					\end{split}
				\end{equation} 
                \cfload[.]
				\argument{\cref{realization multi2};\lref{def: bfI};\lref{def: fU}}{for all  $i\in \{1,2,\dots,\ell_1\}$ that
					\begin{multline}\llabel{NRZZZ}
							\fU_{i}= 
							\bigl\{  
							\theta = ( \theta_1, \dots \theta_{ \ffd(\ell)} ) 
							\in \R^{\ffd(\ell) } 
							\colon 
							\bigl(
							\forall \, x_1,x_2,\dots,x_{\ell_0} \in [a,b] \colon\\
							\eta<\theta_{ \ell_1 \ell_0 + i }  + \ssum_{j=1}^{\ell_0} \theta_{ ( i - 1 ) \ell_0 + j } x_j < \zeta
							\bigr)
							\bigr\}.
				\end{multline}}
				\argument{\lref{NRZZZ};\lref{def: varrho};the assumption that $\xTheta{0}{i}$, $i\in\{1,2,\dots,\ell_1+\ell_0\ell_1\}$, are independent}{that for all $i\in \{1,2,\dots,\ell_1\}$ it holds that
					\begin{equation}\llabel{eq1}
						\begin{split}
							&\P(\Theta_0\in \fU_i)\\
							&\textstyle=\P\bigl(\forall \, x_1,x_2,\dots,x_{\ell_0} \in [a,b] \colon 
							\eta<\xTheta{ 0}{\ell_1 \ell_0 + i }  + \ssum_{j=1}^{\ell_0} \xTheta{ 0}{( i - 1 ) \ell_0 + j } x_j < \zeta\bigr)\\
							&\textstyle\geq \P\bigl(\bigl\{\frac{3\eta+\zeta}{4}<\xTheta{0}{\ell_1\ell_0+i}<\frac{\eta+3\zeta}{4}\bigr\}\cap\bigl\{\forall\, x_1,x_2,\dots,x_{\ell_0}\in [a,b]\colon |\ssum_{j=1}^{\ell_0} \xTheta{ 0}{( i - 1 ) \ell_0 + j } x_j|<\frac{\zeta-\eta}{2}\bigr\}\bigr)\\
							&\textstyle\geq \P\bigl(\bigl\{\frac{3\eta+\zeta}{4}<\xTheta{0}{\ell_1\ell_0+i}<\frac{\eta+3\zeta}{4}\bigr\}\cap\bigl\{\ssum_{j=1}^{\ell_0} |\xTheta{ 0}{( i - 1 )\ell_0 + j } |<\frac{\zeta-\eta}{2\max\{1,|a|,|b|\}}\bigr\}\bigr)\\
							&\textstyle\geq \P\bigl(\bigl\{\frac{3\eta+\zeta}{4}<\xTheta{0}{\ell_1\ell_0+i}<\frac{\eta+3\zeta}{4}\bigr\}\cap\bigl\{\forall\,j\in \{1,2,\dots,\ell_0\}\colon |\xTheta{ 0}{( i - 1 )\ell_0 + j }|<\frac{\zeta-\eta}{2\ell_0\max\{1,|a|,|b|\}}\bigr\}\bigr)\\
							&\textstyle=\bigl[\P\bigl(\frac{3\eta+\zeta}{4}<\xTheta{0}{\ell_1\ell_0+i}<\frac{\eta+3\zeta}{4}\bigr)\bigr]\biggl[\prod\limits_{j=1}^{\ell_0}\P\bigl(|\xTheta{ 0}{( i - 1 )\ell_0 + j }|<\frac{\zeta-\eta}{2\ell_0\max\{1,|a|,|b|\}}\bigr)\biggr]\\
							&=\varrho_i\geq 0\dott
						\end{split}
				\end{equation}}
				\argument{\cref{def: inact};\lref{def: gamma};\lref{def: bfI};\lref{def: fU}}{that
					\begin{equation}\llabel{eq2}
						\bigl(\cap_{i=1}^{\ell_1}\fU_i\bigr)\subseteq \inact_1\dott
				\end{equation}}
				\argument{\lref{NRZZZ};the assumption that $\xTheta{0}{i}$, $i\in \{1,2,\dots,\ell_1+\ell_0\ell_1\}$, are independent}{that \llabel{arg1} $\{\Theta_0\in \fU_i\}$, $i\in \{1,2,\dots,\ell_1\}$, are independent\dott}
				\argument{\lref{arg1};\lref{eq1};\lref{eq2}}{that
					\begin{equation}
						\begin{split}
							&\P(\Theta_0\in \inact_1)\geq\P\bigl(\cap_{i=1}^{\ell_1}\{\Theta_0\in \fU_i\}\bigr)=\textstyle\prod_{i=1}^{\ell_1}\P(\Theta_0\in \fU_i)\geq \prod_{i=1}^{\ell_1}\varrho_i\dott
						\end{split}
				\end{equation}}
			\end{aproof}
            			\subsection{Lower bound for the probability of inactivity for the later hidden layers}\label{subsec: inactivity later hidden layer}
                         \cfclear
			\begin{athm}{lemma}{number of inactiven neuron}
				Assume \cref{setting: SGD2}, for every $k\in \{0,1,\dots,L\}$ let $\bfd_k\in \N$ satisfy $\bfd_k=\sum_{i=1}^{k}\ell_i(\ell_{i-1}\allowbreak+1)$, let $\bfA\in\R$ satisfy $\bfA\leq \inf_{x\in \R} \bbA_0(x)$, let $\gamma\in (-\infty,\min(S\cup\{\betaw\})]$, $\rho\in [-\infty,\infty)$ satisfy
				\begin{equation}\llabel{def: gamma}
				\rho=\begin{cases}
						-\infty&\colon \bfA\geq 0\\
						\textstyle\frac{1}{\bfA\max_{i\in\{1,2,\dots,L-2\}}\ell_i} &\colon \bfA<0, L>2\\
                        0 &\colon \bfA<0,L=2,
					\end{cases}
				\end{equation}
			 assume that $\xTheta{0}{i}$, $i\in \{1,2,\dots,\ffd(\ell)-\ell_L\ell_{L-1}-\ell_L\}$, are independent, and assume $\alphaw=-\infty$ \cfload.
				Then 
				\begin{equation}\llabel{conclude}
                \begin{split}
					&\P\bigl(\Theta_0\notin (\cup_{k=2}^{L-1}\inact_k)\bigr)\\
                    &\leq \textstyle \prod\limits_{k=2}^{L-1}\biggl[1-\biggl[\prod\limits_{j=1}^{\ell_{k}\ell_{k-1}}\P(\rho<\xTheta{0}{j+\bfd_{k-1}}<0)\biggr]\biggl[\prod\limits_{j=1}^{\ell_k}\P\bigl(\xTheta{0}{j+\ell_k\ell_{k-1}+\bfd_{k-1}}<\gamma-\mathbbm 1_{(-\infty,0)}(\bfA)\bigr)\biggr]\biggr]\ifnocf.
                     \end{split}
				\end{equation}
                \cfout[.]
			\end{athm}    
			\begin{aproof}
				Throughout this proof assume without loss of generality that $L\geq 3$ (otherwise \begin{equation}
                \begin{split}
				    &\P(\Theta_0\notin (\cup_{k=2}^{L-1}\inact_k))= 1\\
                    &=\textstyle \prod\limits_{k=2}^{L-1}\biggl[1-\biggl[\prod\limits_{j=1}^{\ell_{k}\ell_{k-1}}\P(\rho<\xTheta{0}{j+\bfd_{k-1}}<0)\biggr]\biggl[\prod\limits_{j=1}^{\ell_k}\P\bigl(\xTheta{0}{j+\ell_k\ell_{k-1}+\bfd_{k-1}}<\gamma-\mathbbm 1_{(-\infty,0)}(\bfA)\bigr)\biggr]\biggr]
                    \end{split}
				\end{equation}
                ), for every $k\in \{2,3,\dots,L-1\}$ let $\InActL{k}\subseteq \R^{\ffd(\ell)}$ satisfy
				\begin{equation}\llabel{def: J}
                \begin{split}
					\InActL{k} &= \bigl\{ \NNel = (\NNel_1,\dots,\NNel_{\ffd(\ell)}) \in \R^{\ffd(\ell)} \colon \big[ \forall\, j \in \{1,2,\dots,\ell_k\ell_{k-1}\} \colon \rho<\NNel_{j+\bfd_{k-1}}<0  \bigr] \\
                    &\quad \wedge \bigl[\forall\, j \in \{1,2,\dots,\ell_k\} \colon \NNel_{j+\ell_k\ell_{k-1}+\bfd_{k-1}}<\gamma-\mathbbm 1_{(-\infty,0)}(\bfA)\bigr]\bigr\}
                    \end{split}
				\end{equation}
				\argument{\lref{def: gamma};}{that \llabel{arg1} $\rho< 0$\dott}
				In the following we prove that for all $k\in\{2,3,\dots,L-1\}$, $\theta_1,\theta_2,\dots,\theta_{\ell_{k-1}+1}$, $y_1,y_2,\dots,y_{\ell_{k-1}}\in \R$ with $\forall\, i\in \{1,2,\dots,\ell_{k-1}\}\colon \rho<\theta_i<0$ and $\theta_{\ell_{k-1}+1}<\gamma-\mathbbm 1_{(-\infty,0)}(\bfA))$ it holds that
				\begin{equation}\llabel{eq1}
					\begin{split}
						\theta_{\ell_{k-1}+1}+\sum_{i=1}^{\ell_{k-1}} \theta_i\bbA_0(y_i)< \gamma.
					\end{split}
				\end{equation}
				In our proof of \lref{eq1} we distinguish between the case $\bfA\geq 0$ and the case $\bfA<0$. We first prove \lref{eq1} in the case
				\begin{equation}\llabel{case1}
					\bfA\geq 0.
					\end{equation}
					\startnewargseq
					\argument{\lref{def: gamma};\lref{case1};the fact that $\rho< 0$}{that for all $k\in\{2,3,\dots,L-1\}$, $\theta_1,\theta_2,\dots,\theta_{\ell_{k-1}+1}$, $y_1,y_2,\dots,y_{\ell_{k-1}}\in \R$ with $\forall\, i\in \{1,2,\dots,\ell_{k-1}\}\colon \rho<\theta_i<0$ and $\theta_{\ell_{k-1}+1}<\gamma$ it holds that
						\begin{equation}\llabel{case1 eq1}
							\begin{split}
								\theta_{\ell_{k-1}+1}+\sum_{i=1}^{\ell_{k-1}} \theta_i\bbA_0(y_i)\leq\theta_{\ell_{k-1}+1}<\gamma.
							\end{split}
					\end{equation}}
				\argument{\lref{case1 eq1};}{\lref{eq1} in the case $\bfA\geq 0$\dott}
				We now prove \lref{eq1} in the case
				\begin{equation}\llabel{case2}
					\bfA<0.
					\end{equation}
					\startnewargseq
	\argument{\lref{def: gamma};\lref{case2}; the assumption that for all $x\in \R$ it holds that $\bbA_0(x)\geq \bfA$; the assumption that for all $x\in \R$ it holds that $\bbA_0(x)\geq \bfA$;the fact that $\rho< 0$}{that for all $k\in\{2,3,\dots,L-1\}$, $\theta_1,\theta_2,\dots,\theta_{\ell_{k-1}+1}$, $y_1,y_2,\dots,y_{\ell_{k-1}}\in \R$ with $\forall\, i\in \{1,2,\dots,\ell_{k-1}\}\colon \rho<\theta_i<0$ and $\theta_{\ell_{k-1}+1}<\gamma-1$ it holds that
	\begin{equation}\llabel{case2 eq1}
		\begin{split}
			\theta_{\ell_{k-1}+1}+\sum_{i=1}^{\ell_{k-1}} \theta_i\bbA_0(y_i)<\gamma-1+ \sum_{i=1}^{\ell_{k-1}} \theta_i\bbA_0(y_i) \leq \gamma-1+\sum_{i=1}^{\ell_{k-1}}\frac{\bfA}{\bfA\max_{j\in \{1,2,\dots,L-2\}}\ell_j}\leq \gamma.
		\end{split}
\end{equation}}
\argument{\lref{case2 eq1};}{\lref{eq1} in the case $\bfA<0$\dott}
\startnewargseq
				\argument{\cref{realization multi2};\lref{eq1}}{that for all $k\in \{2,3,\dots,L-1\}$, $i\in\{1,2,\dots,\ell_{k}\}$, $\theta=(\theta_1,\dots,\theta_{\ffd(\ell)})$,  $x\in [a,b]^{\ell_0}$ with $\forall\,j \in \{1,2,\dots,\ell_{k-1}\} \colon \rho<\NNel_{j+(i-1)\ell_{k-1}+\bfd_{k-1}}<0 $ and $ \theta_{i+\ell_{k}\ell_{k-1}+\bfd_{k-1}}<\gamma-1$ it holds that
					\begin{equation}\llabel{eq2}
						\relii{\ell}{k}{\theta}{\bbA_0}{i}(x)<\gamma.
				\end{equation}}
			\argument{\lref{eq2};\cref{def: inact};the assumption that  $\gamma\leq \min(S\cup\{\betaw\})$}{that for all $k\in \{2,3,\dots,L-1\}$ it holds that
			\begin{equation}\llabel{eq3}
				\InActL{k}\subseteq \inact_k.
				\end{equation}}
			\argument{\lref{eq3};}{that
			\begin{equation}\llabel{eq4}
				\begin{split}
				& \P\bigl(\Theta_0\in (\cup_{k=2}^{L-1}\inact_k)\bigr)\geq \P\bigl(\Theta_0\in (\cup_{k=2}^{L-1}\InActL{k})\bigr)\\
				&=1-\P\bigl(\Theta_0\in(\Omega\backslash (\cup_{k=2}^{L-1}\InActL{k}))\bigr)=1- \P\bigl(\Theta_0\in (\cap_{k=2}^{L-1}[\Omega\backslash\InActL{k}])\bigr).
				\end{split}
				\end{equation}}
			\argument{\lref{def: J}; the assumption that $\xTheta{0}{i}$, $i\in \{1,2,\dots,\ffd(\ell)-\ell_L\ell_{L-1}-\ell_L\}$, are independent}{that the events $\{\Theta_0\in \InActL{k}\}$, $k\in \{2,3,\dots,L-1\}$, are \llabel{arg2} independent\dott}
			\argument{\lref{arg2};the assumption that $\xTheta{0}{i}$, $i\in \{1,2,\dots,\ffd(\ell)-\ell_L\ell_{L-1}-\ell_L\}$, are independent}{that \begin{equation}\llabel{eq5}
            \begin{split}
				&\textstyle	 \P\bigl(\Theta_0\in (\cap_{k=2}^{L-1}[\Omega\backslash\InActL{k}])\bigr)=\prod\limits_{k=2}^{L-1}[1-\P(\Theta_0\in\InActL{k})]
                =1\\
                &-\textstyle \prod\limits_{k=2}^{L-1}\biggl[1-\biggl[\prod\limits_{j=1}^{\ell_{k}\ell_{k-1}}\P(\rho<\xTheta{0}{j+\bfd_{k-1}}<0)\biggr]\biggl[\prod\limits_{j=1}^{\ell_k}\P\bigl(\xTheta{0}{j+\ell_k\ell_{k-1}+\bfd_{k-1}}<\gamma-\mathbbm 1_{(-\infty,0)}(\bfA)\bigr)\biggr]\biggr].
                \end{split}
			\end{equation}}
		\argument{\lref{eq5};\lref{eq4}}{\lref{conclude}\dott}
			\end{aproof}
            \subsection{Abstract lower bounds for the non-convergence probability}\label{subsec: general analysis final}
             \cfclear
			\begin{athm}{lemma}{Lemma confirm}
				Let $L\in \N\backslash\{1\}$, $\ell=(\ell_0,\ell_1,\dots,\ell_{L}) \in \N^{L+1}$, let $ ( \Omega, \mathcal{F}, \P) $ be a probability space, let $\bbA\colon \R\to\R$ be a function, 
				let $\smalll\colon \R^{ \ell_L } \times \R^{ \ell_L }\to\R$ be measurable, let $\mu\colon \mathcal B([a,b]^{\ell_0}\times\R^{\ell_L})\times \Omega\to [0,\infty]$ satisfy for all $\omega\in \Omega$ that $\mathcal B([a,b]^{\ell_0}\times\R^{\ell_L})\ni \set\mapsto \muu{\set}{\omega}\in [0,\infty]$ is a measure, assume for all $\set\in \mathcal B([a,b]^{\ell_0}\times\R^{\ell_L})$ that $\Omega\ni \omega \mapsto \muu{\set}{\omega}\in [0,\infty]$ is measurable, and assume for all $\theta\in \R^d$, $\omega\in \Omega$ that 
				\begin{equation}\llabel{assume}
					\int_{[a,b]^{\ell_0}\times \R^{\ell_L}}
					| \smalll(\rel{\ell}{\theta}{\bbA}(x),y)|\, \muuu {\mathrm{d}x}{\mathrm{d}y}{\omega}<\infty \ifnocf.
				\end{equation}
                \cfload[.]
				Then it holds for all $\omega\in \Omega$, $z\in \R^{\ell_L}$ that 
				\begin{equation}\llabel{conclude}
					\int_{\R^{\ell_0}\times\R^{\ell_L}}|\smalll(z,y)|\, \muuu {\mathrm{d}x}{\mathrm{d}y}{\omega}<\infty\ifnocf.
				\end{equation}
                \cfout[.]
			\end{athm}
			\begin{aproof}
				Throughout this proof for every $\omega\in \Omega$, $z=(z_1,\dots,z_{\ell_L})\in \R^{\ell_L}$ let $\vartheta_{\omega,z}\in \R^d$ satisfy
				\begin{enumerate}[label=(\roman*)]
					\item \llabel{item 1} it holds for all $i\in \bigl\{1,2,\dots,\ell_{L}\ell_{L-1}+\sum_{h=1}^{L-1} \ell_h(\ell_{h-1}+1)\bigr\}$ that $\vartheta_i=0$ and
					\item \llabel{item 2} it holds for all $i\in \{1,2,\dots,\ell_L\}$ that $\vartheta_{\ell_{L}\ell_{L-1}+i+\sum_{h=1}^{L-1} \ell_h(\ell_{h-1}+1)}=z_i$.
				\end{enumerate}
				\argument{\cref{realization multi2};\lref{item 1,item 2}}{that for all $\omega\in \Omega$, $z\in \R^{\ell_L}$, $x\in \R^{\ell_0}$ that
					\begin{equation}\llabel{eq1}
						\rel{\ell}{\vartheta_{\omega,z}}{\bbA}(x)=z\dott
				\end{equation}}
				\argument{\lref{eq1};\lref{assume}}{for all $\omega\in \Omega$, $z\in \R^{\ell_L}$ that
					\begin{equation}\llabel{eq2}
					\int_{\R^{\ell_0}\times\R^{\ell_L}}|\smalll(z,y)|\, \muuu {\mathrm{d}x}{\mathrm{d}y}{\omega}=\int_{[a,b]^{\ell_0}\times \R^{\ell_L}}
						| \smalll(\rel{\ell}{\vartheta_{\omega,z}}{\bbA}(x),y)|\, \muuu {\mathrm{d}x}{\mathrm{d}y}{\omega}<\infty\dott
				\end{equation}}
			\end{aproof}
             \cfclear
			\begin{athm}{prop}{main theorem general}
			Assume \cref{setting: SGD2}, for every $k\in \{0,1,\dots,L\}$ let $\bfd_k\in \N$ satisfy $\bfd_k=\sum_{i=1}^{k}\ell_i(\ell_{i-1}+1)$, assume $\P\bigl(\inf_{\theta\in \R^{\ffd(\ell)}}\fL(\theta)<\inf_{z\in \mathcal \R^{\ell_L}}\allowbreak\int_{[a,b]^{\ell_0}\times\R^{\ell_L}}\smalll(z,y)\allowbreak\,\muu{\d x}{\d y}\bigr)\allowbreak=1$ (cf.\ \cref{Lemma confirm}), let $\eta\in \R$, $\zeta\in (\eta,\infty)$ satisfy
					\begin{equation}\llabel{def: gamma}
						(\eta,\zeta)\subseteq (\alphaw,\betaw)\backslash S,
				\end{equation}
                 for every $i\in \{1,2,\dots,\ell_1\}$ let $\varrho_i\in \R$ satisfy
                \begin{equation}\llabel{def: varrho}
               \textstyle \varrho_i= \bigl[\P\bigl(\frac{3\eta+\zeta}{4}<\xTheta{0}{\ell_1\ell_0+i}<\frac{\eta+3\zeta}{4}\bigr)\bigr]\bigl[\prod_{j=1}^{\ell_0}\P\bigl(|\xTheta{ 0}{( i - 1 )\ell_0 + j }|<\frac{\zeta-\eta}{2\ell_0\max\{1,|a|,|b|\}}\bigr)\bigr].  
                \end{equation}
                let $\bfA\in \R$ satisfy $\bfA\leq \inf_{x\in\R}\bbA_0(x)$, let $\gamma\in (-\infty,\min(S\cup\{\betaw\})]$, $\rho\in [-\infty,\infty)$ satisfy
				\begin{equation}\llabel{def: gamma1}
				\rho=\begin{cases}
						-\infty&\colon \bfA\geq 0\\
						\textstyle\frac{1}{\bfA\max_{i\in\{1,2,\dots,L-2\}}\ell_i} &\colon \bfA<0, L>2\\
                        0 &\colon \bfA<0,L=2,
					\end{cases}
				\end{equation}
			let $\chi \in \N\cap [\ell_1+\ell_0\ell_1,\ffd(\ell)]$, and assume that $\xTheta{0}{i}$, $i\in \{1,2,\dots,\chi\}$, are independent \cfload.
				Then
				\begin{equation}\label{conclude: most general}
                \begin{split}
					&\textstyle
					\P\Bigl(\inf\limits_{ n\in \N_0 }  \fL(\Theta_n)>\inf\limits_{\theta\in \R^{\ffd(\ell)}}\fL(\theta)\Bigr)\\
                    &\geq \textstyle\max\biggl\{\prod\limits_{i=1}^{\ell_1}\varrho_i,\biggl[1-\textstyle \biggl(\prod\limits_{k=2}^{L-1}\biggl[1-\biggl[\prod\limits_{j=1}^{\ell_{k}\ell_{k-1}}\P(\rho<\xTheta{0}{j+\bfd_{k-1}}<0)\biggr]\\
                &\textstyle\cdot\biggl[\prod\limits_{j=1}^{\ell_k}\P\bigl(\xTheta{0}{j+\ell_k\ell_{k-1}+\bfd_{k-1}}<\gamma-\mathbbm 1_{(-\infty,0)}(\bfA)\bigr)\biggr]\biggr]\biggr)\biggr]\mathbbm 1_{\{-\infty\}}(\alphaw)\mathbbm 1_{[\ffd(\ell)-\ell_L\ell_{L-1}-\ell_L,\infty)}(\chi)\biggr\}\ifnocf.
                \end{split}
                	\end{equation}
                    \cfout[.]
			\end{athm}
			\begin{aproof}
				\argument{\cref{Estimate not convergence2};\cref{estimate inactive2};\cref{number of inactiven neuron}}{\cref{conclude: most general}\dott}
			\end{aproof}
			\section{Improved risks through ANNs with non-constant realization functions}\label{sec: improve risk}
In the main result of this section, \cref{verify maintheorem 5} in \cref{sucsec: improve risk for deep ANN} (which is a key new contribution of this work), we reveal a concrete sufficient condition (cf.\ \cref{assume1: a} in \cref{verify maintheorem 5}) which ensures that \ANNs\ with constant realization functions can not reach the optimal/infimal value of the risk (cf.\ \cref{conclude: main improve risk} in \cref{verify maintheorem 5}). In \cref{sec: application} we apply \cref{verify maintheorem 5} together with the main result of \cref{sec: general analysis}, \cref{main theorem general} in \cref{subsec: general analysis final} above, to establish in \cref{main cor: scientific general square} a general lower bound for the probability that the true risk of a general \SGD\ optimization process does \emph{not} converge to the optimal/infimal true risk value. \cref{main theorem1} and \cref{main theorem2} in the introduction can then be proved through an application of \cref{main cor: scientific general square} (cf.\ \cref{main cor: scientific square mean} in \cref{subsec: high probability} and \cref{main theo: scientific square mean error very deep} in \cref{subsec: high probability} for details).



			\subsection{Locally improved risks for shallow ANNs}
            Roughly speaking, in \cref{lem: globally risk} below we show under suitable assumptions (such as the hypothesis that the \ANN\ activation function is not a polynomial) that for every \ANN\ with a constant realization function there exists a certain shallow \ANN\ with a non-constant realization functions whose true risk is strictly smaller than the true risk of the \ANN\ with the constant realization function.
In our proof of \cref{lem: globally risk} we employ the property that the nonlinear function $\bbA\colon \R\to\R$ (which plays in \cref{lem: globally risk} the role of the activation function of the \ANN\ realization functions) is in a suitable sense \emph{discriminatory} for the probability distribution of the input datum $X\colon \Omega \to \R^d$ of the considered learning problem. For convenience of the reader we call this discriminatory property from the literature (see, \eg, \cite[Definition 3.7]{ArAd2024}) within the following notion, \cref{def:disc} below.

\begin{definition}
\label{def:disc}
Let $ d \in \N $, 
let $ K \subseteq \R^d $ be compact, 
let $ \mu \colon \cB( K ) \to [0,\infty] $ 
be a measure, 
and let 
$
  \bbA \colon \R \to \R 
$ 
be locally bounded and measurable. 
Then we say that $ \bbA $ is discriminatory for $ \mu $ 
if and only if it holds for all 
measurable $ \varphi \colon K \to \R $ 
with 
$
  \int_K | \varphi(x) | \, \mu( \d x ) < \infty
$
and 
\begin{equation}\label{eq1: def:disc}
  \sup_{
    w \in \R^d 
  } 
  \sup_{ 
    b \in \R 
  }
  \abs*{
    \int_K 
    \bbA( \spro{ w, x } + b ) 
    \, 
    \varphi( x ) 
    \,
    \mu( \d x ) }
  = 0
\end{equation}
that $ \int_K | \varphi( x ) | \, \mu( \d x ) = 0 $.
\end{definition}
\cref{def:disc} is essentially coincides with Definition 3.7 in \cite{ArAd2024}. We note that the discriminatory notion in \cref{eq1: def:disc} is slightly different from the standard notion of a discriminatory function in the literature (cf., \eg., \cite[Section 3]{Benth2023}, \cite[Section 2]{Cybenko1989}, 
\cite[Definition 2.1]{Guehring2022}, and \cite[Definition 4.1]{Capel2020_Approx} and the references therein).
             \cfclear
			\begin{athm}{prop}{lem: globally risk}
				Let $ ( \Omega, \mathcal{F}, \P) $ be a probability space, let $d,\delta\in \N$, $ a \in \R $, 
				$ b \in (a, \infty)  $,
				let 
				$ X \colon \Omega \to [a,b]^{d} $ and $Y\colon \Omega\to \R^\delta$
				be random variables, let $\smalll \colon  \R^\delta \times \R^\delta \to \R $ be measurable, let $U\subseteq \R$ be open, assume $0\in U$, let $y,\bfe\in \R^\delta$ satisfy for all $w\in \R^\delta$ that $(U\ni r\mapsto \smalll(y+r\bfe,w)\in \R)\in C^1(U,\R)$, let $\g\colon U\times\R^\delta\to\R$ satisfy for all $r\in U$, $w\in \R^\delta$ that $\g(r,w)=\frac{\d}{\d r} \smalll(y+r\bfe,w)$, assume 
				\begin{equation}\llabel{assume1}
					\textstyle \E\bigl[ | \smalll(y,Y) | + \sup_{ r \in U } | \g(r,Y) | \bigr] < \infty\qqandqq \E\bigl[|\E[\g(0,Y)|X]|\bigr]>0,
				\end{equation}
				let $\mathbb A\in C(\R,\R)$, and assume that $\bbA$ is not a polynomial\cfload.
				Then there exists $\varepsilon,\varrho\in (0,\infty)$ such that
				\begin{enumerate}[label=(\roman*)]
					\item\llabel{item 1} it holds that $\E\bigl[\sup_{r\in [-\varepsilon,\varepsilon]}\sup_{w\in [-\varrho,\varrho]^{1\times d}}\sup_{z\in [-\varrho,\varrho]}|\smalll(y+r\bfe\mathbb A(\W X + z),Y)|\bigr]<\infty$ and
					\item \llabel{item 2} it holds that
					\begin{equation}\llabel{conclude}
						\inf_{r\in [-\varepsilon,\varepsilon]}
						\inf_{ \W \in [-\varrho,\varrho]^{1\times d} }\inf_{z\in [-\varrho,\varrho]}
						\E\bigl[ \smalll(y+r\bfe\mathbb A(\W X + z),Y)  \bigr]<\E\bigl[\smalll(y,Y)\bigr]\ifnocf.
					\end{equation}
                    \cfout[.]
				\end{enumerate}
			\end{athm}
			\begin{aproof}
				Throughout this proof let $\mu\colon \mathcal B([a,b]^d)\to [0,1]$ satisfy for all $\set\in \mathcal B([a,b]^d)$ that 
				\begin{equation}\llabel{def: mu}
					\mu(\set)=\P(X^{-1}(\set)).
				\end{equation} 
				\argument{\lref{assume1};the assumption that $0\in U$}{that
					\begin{equation}\llabel{eqt0.1}
						\textstyle\E\bigl[|\smalll(y,Y)|+|\g(0,Y)|\bigr]<\infty.
				\end{equation}}
				\argument{\lref{eqt0.1};}{that
					\begin{equation}\llabel{eqt0.2}
						\textstyle
						\E\bigl[|\E[\g(0,Y)|X]|\bigr]\leq   \E\bigl[\E\bigl[|\g(0,Y)| \big|X\bigr]\bigr]=\E\bigl[|\g(0,Y)|\bigr]<\infty.
				\end{equation}}
				\argument{\lref{eqt0.1}; the factorization lemma for conditional expectations (see, \eg, \cite[Corollary 4.8]{HannibalJentzenThang2024})}{that there exists a measurable $\varphi\colon \R^d\to \R$ which satisfies $\P$-a.s.\ that
					\begin{equation}\llabel{def: varphi}
						\textstyle\varphi(X)=\E\bigl[\g(0,Y)|X\bigr].
				\end{equation}}
				In the following for every $\varrho\in (0,\infty)$ let $\kappa_\varrho\in [0,\infty]$ satisfy
				\begin{equation}\llabel{def: kappa_varrho}
					\kappa_\varrho=\E\bigl[|\varphi(X)|\bigr]+  \sup_{\W\in [-\varrho,\varrho]^{1\times d}}\sup_{z\in [-\varrho,\varrho]}\sup_{x\in [a,b]^d}|\mathbb A(\W x+z)|.
				\end{equation}
				\startnewargseq
				\argument{\lref{def: mu};\lref{eqt0.2};\lref{def: varphi};the change of variable formula for pushforward measures;the assumption that $\E\bigl[|\E[\g(0,Y)|X]|\bigr]>0$}{that 
					\begin{equation}\llabel{eqt1}
						\int_{[a,b]^d}|\varphi(x)|\,\mu(\d x)= \E\bigl[|\varphi(X)|\bigr]=\textstyle\E\bigl[|\E[\g(0,Y)|X]|\bigr] \in (0,\infty)\dott
				\end{equation}}
				\argument{the assumption that $\mathbb A\in C(\R,\R)$;}{that
					\begin{equation}\llabel{eqtt1}
						\sup_{\W\in [-\varrho,\varrho]^{1\times d}}\sup_{z\in [-\varrho,\varrho]}\sup_{x\in [a,b]^d}|\mathbb A(\W x+z)|<\infty\dott
				\end{equation}} 
				\argument{\lref{eqtt1};\lref{def: kappa_varrho};\lref{eqt1}}{that for all $\varrho\in (0,\infty)$ it holds that
					\begin{equation}\llabel{eqtt2}
						\kappa_\varrho\in (0,\infty)\dott
				\end{equation}}
				\argument{\lref{eqtt2};\lref{def: kappa_varrho};}{that for all $\varrho,\varepsilon\in (0,\infty)$, $\W\in [-\varrho,\varrho]^{1\times d}$, $z\in [-\varrho,\varrho]$, $r\in (-\varepsilon,\varepsilon)$, $x\in [a,b]^d$ it holds that \llabel{eqtt3}$|r\mathbb A(\W x+z)|\leq \kappa_\varrho \varepsilon<\infty$\dott}
				\argument{\lref{eqtt3};}{for all $\varrho,\varepsilon\in (0,\infty)$, $\W\in [-\varrho,\varrho]^{1\times d}$, $z\in [-\varrho,\varrho]$, $r\in (-\varepsilon,\varepsilon)$, $x\in [a,b]^d$ that
					\begin{equation}\llabel{NR2}
						-\infty<-\kappa_\varrho\varepsilon\leq  -r\mathbb A(\W x+z) \leq \kappa_\varrho \varepsilon<\infty.
				\end{equation}}
				\argument{the assumption that $\bbA$ is continuous;the assumption that $\bbA$ is not a polynomial}{that \llabel{arg2} there is no polynomial which agrees with $\bbA$ almost everywhere\dott}
				\argument{\lref{arg2};\cite[Lemma 3.10]{ArAd2024};}{that \llabel{arg3} $\mathbb A$ is discriminatory for $\mu$ (cf.\ \cref{def:disc})\dott}
				\argument{\lref{arg3};\cref{eq1: def:disc};\lref{eqt1}}{that there exist $\bfw\in \R^{1\times d}$, $\bfz\in \R$ which satisfy
					\begin{equation}\llabel{eq4}
						\int_{[a,b]^d} \varphi(x)\mathbb A(\bfw x + \bfz)\,\mu(\d x)\neq 0.
				\end{equation}}
				\argument{\lref{eqtt2}; the assumption that $U\subseteq\R$ is open; the assumption that $0\in U$;}{that for all $\varrho\in (0,\infty)$ there exists \llabel{argp1} $\varepsilon\in (0,\infty)$ such that $[-\kappa_\varrho\varepsilon,\kappa_\varrho \varepsilon]\subseteq U$\dott}
				\argument{\lref{argp1};}{that there exist $\varrho,\varepsilon\in (0,\infty)$ which satisfy 
					\begin{equation}\llabel{NR1}
						\bfw\in[-\varrho,\varrho]^{1\times d},\qquad \bfz\in [-\varrho,\varrho],\qqandqq [-\kappa_\varrho\varepsilon,\kappa_\varrho\varepsilon]\subseteq U
				\end{equation}}
				\startnewargseq
				\argument{\lref{NR2};\lref{NR1}}{that for all $\W\in [-\varrho,\varrho]^{1\times d}$, $z\in [-\varrho,\varrho]$, $r\in [-\varepsilon,\varepsilon]$, $x\in [a,b]^d$ it holds that
					\begin{equation}\llabel{NR3}
						-r\mathbb A(\W x+z)\in U\dott
				\end{equation}}
				\argument{\lref{NR3}; the assumption that for all $\free\in \R^\delta$ it holds that $
					(U\ni r\mapsto \smalll(y+r\bfe,W)\in \R)\in C^1(U,\R)$}{that for all $\W\in[-\varrho,\varrho]^{1\times d}$, $z\in [-\varrho,\varrho]$, $\free\in \R^\delta$, $\omega\in \Omega$ it holds that
					\begin{equation}\llabel{argp1.5}
						((-\varepsilon,\varepsilon)\ni r\mapsto \smalll(y-r\bfe\mathbb A(\W X(\omega)+z),\free)\in \R)\in C^1((-\varepsilon,\varepsilon),\R)\dott
				\end{equation}}
				\argument{the fundamental theorem of calculus;}{that for all $r \in (-\varepsilon,\varepsilon)$, $f \in C^1( (-\varepsilon,\varepsilon), \R )$ it holds that
					\begin{equation}\llabel{NR}
						f(r)=f(0)+f'(0)r+\int_{0}^1[f'(rs)-f'(0)]r\,\d s.
				\end{equation}}
				\argument{\lref{argp1.5}; the assumption that for all $r\in U$, $w\in \R^\delta$ it holds that $\g(r,w)=\frac{\d}{\d r} \smalll(y+r\bfe,w)$; \lref{NR}}{that for all $r\in (-\varepsilon,\varepsilon)$, $\W\in[-\varrho,\varrho]^{1\times d}$, $z\in [-\varrho,\varrho]$, $\free\in \R^\delta$, $\omega\in \Omega$ it holds that
					\begin{equation}\llabel{argp1.6}
						\begin{split}
							& \smalll( y-r\bfe\mathbb A(\W X(\omega) + z),Y(\omega))\\
							&\textstyle=\smalll(y,Y(\omega))-r\g(0,Y(\omega))\mathbb A(\W X(\omega) + z)\\
							&\textstyle+\int\limits_{0}^1\bigl[\g(0,Y(\omega))-\g(-rs\mathbb A(\W X(\omega) + z),Y(\omega))\bigr]r \mathbb A(\W X(\omega) + z)\, \d s.
						\end{split}
				\end{equation}}
				\argument{\lref{argp1.6};\lref{def: kappa_varrho}; the fact that for all $\omega\in \Omega$ it holds that $X(\omega)\in [a,b]^d$}{that for all $r\in (-\varepsilon,\varepsilon)$, $\W\in[-\varrho,\varrho]^{1\times d}$, $z\in [-\varrho,\varrho]$, $\free\in \R^\delta$, $\omega\in \Omega$ it hols that
					\begin{equation}\llabel{argp1.7}
						\begin{split}
							& |\smalll( y-r\bfe\mathbb A(\W X(\omega) + z),Y(\omega))|\\
							&\textstyle\leq |\smalll(y,Y(\omega))|+\kappa_\varrho|r\g(0,Y(\omega))|\\
							&\textstyle+\int\limits_{0}^1|r|\kappa_\varrho\bigl[|\g(0,Y(\omega))|+|\g(-rs\mathbb A(\W X(\omega) + z),Y(\omega))|\bigr]\, \d s.
						\end{split}
				\end{equation}}
				\argument{\lref{assume1};\lref{NR3};\lref{argp1.7}}{that 
					\begin{equation}\llabel{bounded moment}
						\textstyle  \E\biggl[\sup\limits_{r\in [-\varepsilon,\varepsilon]}\sup\limits_{\W\in [-\varrho,\varrho]^{1\times d}}\sup\limits_{z\in [-\varrho,\varrho]}|\smalll( y-r\bfe\mathbb A(\W X + z),Y)|\biggr]<\infty\dott
				\end{equation}}
				\argument{\lref{bounded moment}; \lref{argp1.6}}{that for all $r\in (-\varepsilon,\varepsilon)$, $\W\in [-\varrho,\varrho]^{1\times d}$, $z\in [-\varrho,\varrho]$ it holds that
					\begin{equation}\llabel{eq2pre}
						\begin{split}
							& \E\bigl[\smalll( y-r\bfe\mathbb A(\W X + z),Y) \bigr]\\
							&\textstyle=\E\bigl[\smalll(y,Y)\bigr]-r\E\bigl[\g(0,Y)\mathbb A(\W X + z)\bigr]\\
							&\textstyle+\biggl[\int\limits_{0}^1\E\bigl[\bigl[\g(0,Y)-\g(-rs\mathbb A(\W X + z),Y)\bigr]r \mathbb A(\W X + z)\bigr]\, \d s\biggr].
						\end{split}
				\end{equation}}
				\argument{\lref{def: varphi};the tower property for conditional expectations}{that for all $\W\in [-\varrho,\varrho]^{1\times d}$, $z\in [-\varrho,\varrho]$ it holds that
					\begin{equation}\llabel{eq2pre1}
						\begin{split}
							& \E\bigl[\g(0,Y)\mathbb A(\W X+z)\bigr]=\E\bigl[\E[\g(0,Y)\mathbb A(\W X+z)|X]\bigr]\\
							&=\E\bigl[\E[\g(0,Y)|X]\mathbb A(\W X+z)\bigr]=\E\bigl[\varphi(X)\mathbb A(\W X+z)\bigr]\dott
						\end{split}
				\end{equation}}
				\argument{\lref{eq2pre};\lref{eq2pre1};}{for all $r\in (-\varepsilon,\varepsilon)$, $\W\in [-\varrho,\varrho]^{1\times d}$, $z\in [-\varrho,\varrho]$ that
					\begin{equation}\llabel{eq2}
						\begin{split}
							& \E\bigl[\smalll( y-r\bfe\mathbb A(\W X + z),Y) \bigr]\\
							&\textstyle= \E\bigl[\smalll(y,Y)\bigr]- r\E\bigl[\varphi(X)\mathbb A(\W X + z)\bigr]\\
							&\textstyle+\biggl[\int\limits_{0}^1\E\bigl[\bigl[\g(0,Y)-\g(-rs\mathbb A(\W X + z),Y)\bigr]r \mathbb A(\W X + z)\bigr]\, \d s\biggr].
						\end{split}
				\end{equation}}
				\argument{\lref{def: mu};\lref{eq4}; the change of variable formula for pushforward measures}{that 
					\begin{equation}\llabel{eq5}
						\E\bigl[\varphi(X)\mathbb A(\bfw X + \bfz)\bigr]=\int_{[a,b]^d} \varphi(x)\mathbb A(\bfw x + \bfz)\,\mu(\d x)\neq 0\dott
				\end{equation}}
				In the following for every $\lambda \in \R$ let $\scrv_\lambda\in \R$ satisfy
				\begin{equation}\llabel{eq5.1}
					\scrv_\lambda=\lambda\E\bigl[\varphi(X)\mathbb A(\bfw X + \bfz)\bigr].
				\end{equation}
				\startnewargseq
				\argument{\lref{def: kappa_varrho};\lref{eq2};\lref{eq5.1}}{that for all $\lambda\in \R$ with $|\scrv_\lambda|\leq \varepsilon$ it holds that
					\begin{equation}\llabel{eq6}
						\begin{split}
							& \bigl|\E\bigl[\smalll( y-\scrv_\lambda\bfe\mathbb A(\bfw X + \bfz),Y) \bigr]-\E\bigl[\smalll(y,Y)\bigr]+\lambda \bigl(\E\bigl[\varphi(X)\mathbb A(\bfw X + \bfz)\bigr]\bigr)^2\bigl|\\
							&=\bigl|\E\bigl[\smalll( y-\scrv_\lambda\bfe\mathbb A(\bfw X + \bfz),Y) \bigr]-\E\bigl[\smalll(y,Y)\bigr]+\scrv_\lambda\E\bigl[\varphi(X)\mathbb A(\bfw X + y)\bigr]\bigl|\\
							&=\textstyle\biggl|\int\limits_{0}^1\E\bigl[\bigl(\g(0,Y)-\g(-\scrv_\lambda s\mathbb A(\bfw X + \bfz),Y)\bigr)\scrv_\lambda \mathbb A(\bfw X + \bfz)\bigr]\, \d s\biggr|\\
							&\textstyle\leq \int\limits_{0}^1\bigl|\lambda\E\bigl[\varphi(X)\mathbb A(\bfw X + \bfz)\bigr]\E\bigl[\bigl(\g(0,Y)-\g(-\scrv_\lambda s\mathbb A(\bfw X + \bfz),Y)\bigr) \mathbb A(\bfw X + \bfz)\bigr]\bigr|\, \d s\\
							&\textstyle\leq|\lambda|(\kappa_{\varrho})^3 \int\limits_{0}^1\E\bigl[|\g(0,Y)-\g(-\scrv_\lambda s\mathbb A(\bfw X + \bfz),Y)|\bigr]\, \d s.
						\end{split}
				\end{equation}}
				\argument{\lref{eq6};\lref{assume1}; Lebesgue's theorem of dominated convergence; the fact that for all $W\in \R^\delta$ it holds that $(U\ni r\mapsto \g(r,W)\in \R)\in C(\R,\R)$ }{that
					\begin{equation}\llabel{eq6.5}
						\begin{split}
							& \limsup\limits_{\lambda\searrow 0}\biggl(\frac{\bigl|\E\bigl[\smalll( y-\scrv_\lambda\bfe\mathbb A(\bfw X + \bfz),Y)\bigr]-\E[\smalll(y,Y)]+\lambda\bigl|\E\bigl[\varphi(X)\mathbb A(\bfw X + \bfz)\bigr]\bigr|^2\bigr|}{\lambda}\biggr)\\
							&\textstyle\leq (\kappa_\varrho)^3\biggl[\limsup\limits_{\lambda\searrow 0}\int\limits_{0}^1\E\bigl[|\g(0,Y)-\g(-\scrv_\lambda s\mathbb A(\bfw X + \bfz),Y)|\bigr]\, \d s\biggr]=0.
						\end{split}
				\end{equation}}
				\argument{\lref{eq6.5};}{that
					\begin{equation}\llabel{eq7}
						\limsup _{\lambda \searrow 0} \Bigl(\frac{\mathbb{E}\bigl[\smalll(y-\scrv_\lambda \mathbf{e} \mathbb{A}(\bfw X+\bfz), Y)\bigr]-\mathbb{E}\bigl[\smalll(y, Y)\bigr]}{\lambda}\Bigr)=-\bigl|\mathbb{E}\bigl[\varphi(X) \mathbb{A}(\bfw X+\bfz)\bigr]\bigr|^2<0.
				\end{equation}}
				\argument{\lref{eq7};}{that there exists $\lambda\in \R$ such that
					\begin{equation}\llabel{eq7.5}
						\sup_{\kappa\in (0,\lambda)}\Bigl(\frac{\mathbb{E}\bigl[\smalll(y-\scrv_\kappa \mathbf{e} \mathbb{A}(\bfw X+\bfz), Y)\bigr]-\mathbb{E}\bigl[\mathfrak{l}(y, Y)\bigr]}{\kappa}\Bigr)<0.
				\end{equation}}
				\argument{\lref{eq7.5};\lref{eq5.1}}{that there exists $r\in [-\varepsilon,\varepsilon]$ such that 
					\begin{equation}\llabel{eq8}
						\E\bigl[ \smalll(y-r\bfe\mathbb A(\bfw X + \bfz),Y)  \bigr]<\E\bigl[\smalll(y,Y)\bigr].
				\end{equation}}
				\argument{\lref{eq8};\lref{bounded moment}}{\lref{item 1, item 2}\dott}
			\end{aproof}
			\subsection{Globally improved risks for deep ANNs}\label{sucsec: improve risk for deep ANN}
             \cfclear
			\begin{athm}{lemma}{lem: not constant1}
				Let $\mathbb A\colon \R\to \R$ satisfy 
				$\sup_{x\in \R}|\bbA(x)-\bbA(0)|>0$\cfload.        
				Then there exist $a,b\in \R$ such that
				\begin{equation}\llabel{conclude}
					\sup_{x\in \R}|\bbA(a\bbA(x)+b)-\bbA(a\bbA(0)+b)|>0\ifnocf.
				\end{equation}
                \cfout[.]
			\end{athm}
			\begin{aproof}
				\argument{the assumption that $\sup_{x\in \R}|\bbA(x)-\bbA(0)|>0$}{that there exists $x\in \R$ which satisfies
					\begin{equation}\llabel{eq1}
						\bbA(x)\neq \bbA(0)\dott
				\end{equation}}
				\startnewargseq
				\argument{\lref{eq1};}{that
					\begin{equation}\llabel{eq2}
						\textstyle\bbA\bigl(\frac{x}{\bbA(x)-\bbA(0)}\bbA(x)-\frac{x\bbA(0)}{\bbA(x)-\bbA(0)}\bigr)=\bbA(x)\neq \bbA(0)=\bbA\bigl(\frac{x}{\bbA(x)-\bbA(0)}\bbA(0)-\frac{x\bbA(0)}{\bbA(x)-\bbA(0)}\bigr).
				\end{equation}}
				\argument{\lref{eq2};}{that
					\begin{equation}
						\textstyle \sup\limits_{y\in \R}\bigl|\bbA\bigl(\frac{x}{\bbA(x)-\bbA(0)}\bbA(y)-\frac{x\bbA(0)}{\bbA(x)-\bbA(0)}\bigr)-\bbA\bigl(\frac{x}{\bbA(x)-\bbA(0)}\bbA(0)-\frac{x\bbA(0)}{\bbA(x)-\bbA(0)}\bigr)\bigr|>0\ifnocf.
				\end{equation}}
				\argument{\lref{eq2};}{\lref{conclude}\dott}
			\end{aproof}
             \cfclear
			\begin{athm}{lemma}{lem: notconstant2}
				Let $L\in \N$, $\alphaw\in \R$, $\betaw\in (\alphaw,\infty)$, let $\mathbb A\colon \R\to\R$ satisfy $\sup_{ x \in ( \alphaw, \betaw ) } | \mathbb A(x) - \mathbb A( \alphaw ) | = 0<\sup_{x\in \R}|\mathbb A(x)-\mathbb A(0)|$, and for every $\theta=(\theta_1,\dots,\theta_{L})$, $\vartheta=(\vartheta_1,\dots,\vartheta_L)\in\R^{L}$ let $\scrN^{ v, \theta,\vartheta }\colon \R \to \R $, $v \in \Z$, satisfy for all $v\in \{0,1,\dots,L\}$, $x\in \R$ that
				\begin{equation}\llabel{def: scrN}
					\scrN^{v, \theta,\vartheta }( x ) = \mathbb A\bigl(x+(\theta_v\scrN^{v-1,\theta,\vartheta}(x)+\vartheta_{v}-x)\mathbbm 1_{\N}(v)\bigr)\cfload.
				\end{equation}
				Then there exist $\theta,\vartheta\in \R^L$ such that
				\begin{equation}\llabel{conclude}
					\sup_{x\in \R}|\scrN^{L, \theta,\vartheta }( x )-\scrN^{L, \theta,\vartheta }( 0 )|>0=\sup_{x\in (\alphaw,\betaw)}|\scrN^{L, \theta,\vartheta }( x )-\scrN^{L, \theta,\vartheta }( \alphaw )|\ifnocf.
				\end{equation}
                \cfout[.]
			\end{athm}
			\begin{aproof}
				Throughout this proof for every let $\scrM^{v,\theta,\vartheta}\colon \R\to\R$, $v\in \{0,1,\dots,L\}$, $\theta$, $\vartheta\in \R^{v+1}$, satisfy for all $v\in \{1,2,\dots,L\}$, $\theta=(\theta_0,\theta_1,\dots,\theta_v)$, $\vartheta=(\vartheta_0,\vartheta_1,\dots,\vartheta_v)\in \R^{v+1}$, $x\in \R$ that
				\begin{equation}\llabel{def: scrM}
					\scrM^{v, \theta,\vartheta }( x ) = \begin{cases}
						\mathbb A\bigl(\theta_{v}\scrM^{v-1,(\theta_0,\theta_1,\dots,\theta_{v-1}),(\vartheta_0,\vartheta_1,\dots,\vartheta_{v-1})}(x)+\vartheta_{v}\bigr), & v\geq 1\\
						\bbA(x), &v=0.
					\end{cases}
				\end{equation}
				In the following we prove that for all $v\in \{0,1,\dots,L\}$ there exist $\theta=(\theta_0,\theta_1,\dots,\theta_v)$, $\vartheta=(\vartheta_0,\vartheta_1,\dots,\vartheta_v)\in \R^{v+1}$ such that
				\begin{equation}\llabel{need to prove}
					\sup_{x\in \R}|\scrM^{v, \theta,\vartheta }( x )-\scrM^{v, \theta,\vartheta }( 0 )|>0=\sup_{x\in (\alphaw,\betaw)}|\scrM^{v, \theta,\vartheta }( x )-\scrM^{v, \theta,\vartheta }( \alphaw)|.
				\end{equation}
				We prove \lref{need to prove} by induction on $v\in \{0,1,\dots,L\}$. For the base case $v=0$ note that \lref{def: scrM} and the assumption that $\sup_{ x \in ( \alphaw, \betaw ) } | \mathbb A(x) - \mathbb A( \alphaw ) | = 0<\sup_{x\in \R}|\mathbb A(x)-\mathbb A(0)|$ imply that for all $\theta,\vartheta\in \R$ it holds that
				\begin{equation}\llabel{eq1}
					\begin{split}
						&\sup_{x\in \R}|\scrM^{0, \theta,\vartheta }( x )-\scrM^{0, \theta,\vartheta }( 0 )|=\sup_{x\in \R}|\bbA(x)-\bbA(x)|>0\\
						&=\sup_{x\in (\alphaw,\betaw)}|\bbA(x)-\bbA(\alphaw)|
						=\sup_{x\in (\alphaw,\betaw)}|\scrM^{0, \theta,\vartheta }( x )-\scrM^{0, \theta,\vartheta }( \alphaw)|\dott
					\end{split}
				\end{equation}
				This proves \lref{need to prove} in the base case $v=0$.
				For the induction step we assume that there exist $v\in \{0,1,\dots,L-1\}$, $\theta=(\theta_0,\theta_1,\dots,\theta_v)$, $\vartheta=(\vartheta_0,\vartheta_1,\dots,\vartheta_v)\in \R^{v+1}$ which satisfy
				\begin{equation}\llabel{induction}
					\sup_{x\in \R}|\scrM^{v, \theta,\vartheta }( x )-\scrM^{v, \theta,\vartheta }( 0 )|>0=\sup_{x\in (\alphaw,\betaw)}|\scrM^{v, \theta,\vartheta }( x )-\scrM^{v, \theta,\vartheta }( \alphaw)|.
				\end{equation}
				\startnewargseq
				\argument{\cref{lem: not constant1};\lref{induction}}{that there exists $a,b\in \R$ which satisfies
					\begin{equation}\llabel{eq2}
						\begin{split}
							& \sup_{x\in \R}|\bbA(a\scrM^{v,\theta,\vartheta}(x)+b)-\bbA(a\scrM^{v,\theta,\vartheta}(0)+b)|>0\\
							&=\sup_{x\in (\alphaw,\betaw)}|\bbA(a\scrM^{v,\theta,\vartheta}(x)+b)-\bbA(a\scrM^{v,\theta,\vartheta}(\alphaw)+b)|.
						\end{split}
				\end{equation}}
				\startnewargseq
				\argument{\lref{def: scrM};\lref{eq2}}{that
					\begin{equation}\llabel{eq3}
						\begin{split}
							& \sup_{x\in \R}|\scrM^{v+1, (\theta_0,
								\theta_1,\dots,\theta_v,a),(\vartheta_0,\vartheta_1,\dots,\vartheta_v,b)} ( x )-\scrM^{v, \theta,\vartheta }( 0 )|>0\\
							&= \sup_{x\in (\alphaw,\betaw)}|\scrM^{v+1, (\theta_0,
								\theta_1,\dots,\theta_v,a),(\vartheta_0,\vartheta_1,\dots,\vartheta_v,b)} ( x )-\scrM^{v, \theta,\vartheta }( \alphaw)|.
						\end{split}
				\end{equation}}
				\argument{\lref{eq3};\lref{eq1};induction}{\lref{need to prove}\dott}
				\startnewargseq
				\argument{\lref{def: scrN};\lref{def: scrM}}{that for all $\theta=(\theta_0,\theta_1,\dots,\theta_L), \vartheta=(\vartheta_0,\vartheta_1,\dots,\vartheta_L)\in \R^{L+1}$, $x\in \R$ it holds that
					\begin{equation}\llabel{eq4}
						\scrM^{L,\theta,\vartheta}(x)=\scrN^{L,(\theta_1,\theta_2,\dots,\theta_L),(\vartheta_1,\vartheta_2,\dots,\vartheta_L)}(x)\dott
				\end{equation}}
				\argument{\lref{need to prove};\lref{eq4}}{\lref{conclude}\dott}
			\end{aproof}
             \cfclear
			\begin{athm}{lemma}{ANN cover}
				Let $\alphaw\in \R$, $\betaw\in (\alphaw,\infty)$, $L\in\N\cap[3,\infty)$, $\ell=(\ell_0,\ell_1,\dots,\ell_L)\in \N^{L+1}$, let $\mathbb A\colon\R\to\R$ satisfy $\sup_{ x \in ( \alphaw, \betaw ) } | \mathbb A(x) - \mathbb A( \alphaw ) | = 0<\sup_{x\in \R}|\mathbb A(x)-\mathbb A(0)|$,
				for every $\theta=(\theta_1,\dots,\theta_{L-2})$, $\vartheta=(\vartheta_1,\dots,\vartheta_{L-2})\in\R^{L-2}$ let $\scrN^{ v, \theta,\vartheta }\colon \R \to \R $, $v \in \Z$, satisfy for all $v\in \{0,1,\dots,L-2\}$, $x\in \R$ that
				\begin{equation}\llabel{def: scrN}
					\scrN^{v, \theta,\vartheta }( x ) = \mathbb A\bigl(x+(\theta_v\scrN^{v-1,\theta,\vartheta}(x)+\vartheta_{v}-x)\indicator{\N}(v)\bigr),
				\end{equation}
				and let $r\in \R^{\ell_L}$, $\W\in \R^{1\times \ell_0}$, $z\in \R$, $\bftheta,\bfvartheta\in \R^{L-2}$, $y,\bfe\in \R^{\ell_L}$\cfload.
				Then
				\begin{enumerate}[label=(\roman*)]
					\item \label{item 1: ANN cover} there exist $\theta,\vartheta\in \R^{L-2}$ such that $\sup_{x\in (\alphaw,\betaw)}\allowbreak|\scrN^{L-2, \theta,\vartheta }( x )-\scrN^{L-2, \theta,\vartheta }( \alphaw )|=0< \sup_{x\in \R}\allowbreak|\scrN^{L-2, \theta,\vartheta }( x )-\scrN^{L-2, \theta,\vartheta }( 0 )|$ and
					\item \label{item 2: ANN cover} there exists $\theta\in \R^{\ffd(\ell)}$ such that for all $x\in \R^{\ell_0}$ it holds that
					$\rel{\ell}{\theta}{\bbA}(x)=y+r\bfe\scrN^{L-2, \bftheta,\bfvartheta }( 
					\W x+z)$\ifnocf.
				\end{enumerate}
                \cfout[.]
			\end{athm}
			\begin{aproof}
				\argument{\cref{lem: notconstant2};}{\cref{item 1: ANN cover}\dott}
				In the following let $W=(W_1,\dots,W_{\ell_0})$, $\Theta=(\Theta_1,\dots,\Theta_{L-2})\in \R^{L-2}$, $\varTheta=(\varTheta_1,\dots,\varTheta_{L-2})\in \R^{L-2}$, $\bfy=(\bfy_1,\dots,\bfy_{\ell_L})\in \R^{\ell_L}$, $\scre=(\scre_1,\dots,\scre_{\ell_L})\in \R^{\ell_L}$ satisfy
				\begin{equation}\llabel{def}
					W=\W^T,\qquad \Theta=\bftheta,\qquad \varTheta=\bfvartheta,\qquad \bfy=y, \qqandqq \scre=\bfe
				\end{equation}
				and let $\theta=(\theta_1,\dots,\theta_{\ffd(\ell)})\in \R^{\ffd(\ell)}$ satisfy
				\begin{enumerate}[label=(\Roman*)]
					\item \llabel{item1} it holds for all $i\in \{1,2,\dots,\ell_0\}$ that $\theta_i=W_i$,
					\item \llabel{item2} it holds that $\theta_{\ell_1\ell_0+1}=z$,
					\item \llabel{item3} it holds for all $i\in \{1,2,\dots,L-2\}$ that $\theta_{1+\sum_{h=1}^{i}\ell_h(\ell_{h-1}+1)}=\Theta_i$,
					\item \llabel{item4} it holds for all $i\in \{1,2,\dots,L-2\}$ that $\theta_{\ell_{i+1}\ell_{i}+1+\sum_{h=1}^{i}\ell_h(\ell_{h-1}+1)}=\varTheta_i$,
					\item \llabel{item5} it holds for all $i\in \{1,2,\dots,\ell_L\}$ that $\theta_{1+(i-1)\ell_{L-1}\sum_{h=1}^{L-1}\ell_h(\ell_{h-1}+1)}=r\scre_i$,
					\item \llabel{item6} it holds for all $i\in \{1,2,\dots,\ell_L\}$ that $\theta_{\ell_{L}\ell_{L-1}+i+\sum_{h=1}^{L-1}\ell_h(\ell_{h-1}+1)}=\bfy_i$, and
					\item \llabel{item7} it holds for all $i\in\bigl(\{1,2,\dots,\ffd(\ell)\}\backslash\bigl( \cup_{j=1}^{\ell_0}\cup_{k=1}^{L-2}\cup_{l=1}^{\ell_L}\bigl\{j,\ell_1\ell_0+1,1+\sum_{h=1}^{k}\ell_h(\ell_{h-1}+1),1+\ell_{k+1}\ell_k+\sum_{h=1}^{L-1}\ell_h(\ell_{h-1}+1),1+(l-1)\ell_L+\sum_{h=1}^{L-1}\ell_h(\ell_{h-1}+1),\ell_L\ell_{L-1}+l\}\bigr)\bigr)$ that $\theta_i=0$.
				\end{enumerate}
				\startnewargseq
				In the following we prove that for all $v\in \{0,1,\dots,L-3\}$, $i\in \{1,2,\dots,\ell_{v+2}\}$, $x\in \R^{\ell_0}$ it holds that
				\begin{equation}\llabel{need to prove}
					\relii{\ell}{1}{\theta}{\bbA}{i}(x)=\bigl(\Theta_{v}\scrN^{v,\Theta,\varTheta}(W x+z)+\varTheta_{v}\bigr)\mathbbm 1_{\{1\}}(i).
				\end{equation}
				We prove \lref{need to prove} by induction on $v\in\{0,1,\dots,L-3\}$. For the base case $v=0$ note that \cref{realization multi2} and \lref{item1,item2,item7} imply that for all $i\in \{1,2,\dots,\ell_1\}$, $x=(x_1,\dots,x_{\ell_0})\in \R^{\ell_0}$ it holds that 
				\begin{equation}\llabel{base1}
					\begin{split}
						\textstyle \relii{\ell}{1}{\theta}{\bbA}{i}(x)=\theta_{\ell_1\ell_0+1}+\biggl(\sum\limits_{j=1}^{\ell_1}\theta_{(i-1)\ell_1+j}x_j\biggr)=(W x+z)\mathbbm 1_{\{1\}}(i).      
					\end{split}
				\end{equation}
				Combining \lref{def: scrN} and \lref{item3,item4,item7},  hence implies for all $i\in \{1,2,\dots,\ell_2\}$, $x\in \R^{\ell_0}$ that
				\begin{equation}\llabel{base}
					\relii{\ell}{2}{\theta}{\bbA}{i}(x)=\bigl(\Theta_1\bbA\bigl(\relii{\ell}{1}{\theta}{\bbA}{1}(x)\bigr)+\varTheta_1\bigr)\mathbbm 1_{\{1\}}(i)=\bigl(\Theta_{1}\scrN^{0,\Theta,\varTheta}(W x+z)+\varTheta_{1}\bigr)\mathbbm 1_{\{1\}}(i).
				\end{equation}
				This implies \lref{need to prove} in the base case $v=0$. For the induction step we assume that there exists $v\in\{0,1,\dots,L-4\}$ which satisfies for all $i\in \{1,2,\dots,\ell_{v+2}\}$, $x\in \R^{\ell_0}$ that
				\begin{equation}\llabel{assume induction}
					\relii{\ell}{v+2}{\theta}{\bbA}{i}(x)=\bigl(\Theta_{v}\scrN^{v,\Theta,\varTheta}(W x+z)+\varTheta_{v}\bigr)\mathbbm 1_{\{1\}}(i).
				\end{equation}
				\startnewargseq
				\argument{\cref{realization multi2};\lref{def: scrN};\lref{assume induction};\lref{item3,item4}}{that for all $i\in \{1,2,\dots,\allowbreak\ell_{v+3}\allowbreak\}$, $x\in \R^{\ell_0}$ it holds that
					\begin{equation}\llabel{eq2}
						\begin{split}
							& \relii{\ell}{v+3}{\theta}{\bbA}{i}(x)=\bigl(\Theta_{v+1}\bbA\bigl(\relii{\ell}{v+2}{\theta}{\bbA}{1}(x)\bigr)+\varTheta_{v+1}\bigr)\mathbbm 1_{\{1\}}(i)\\
							&=\bigl(\Theta_{v+1}\bbA\bigl(\Theta_{v}\scrN^{v,\Theta,\varTheta}(W x+z)+\varTheta_{v})+\varTheta_{v+1}\bigr)\mathbbm 1_{\{1\}}(i)\\
							&=\bigl(\Theta_{v+1}\scrN^{v+1,\Theta,\varTheta}(W x+z)+\varTheta_{v+1}\bigr)\mathbbm 1_{\{1\}}(i).
						\end{split}
				\end{equation}}
				\argument{\lref{eq2};\lref{base};induction}{\lref{need to prove}\dott}
				\startnewargseq
				\argument{\cref{realization multi2};\lref{def: scrN};\lref{need to prove};\lref{item5,item6,item7}}{that for all  $x\in \R^{\ell_0}$ it holds that
					\begin{equation}\llabel{eq3}
						\begin{split}
							&\rel{\ell}{\theta}{\bbA}(x)=y+r\scre \bbA\bigl(\relii{\ell}{L-1}{\theta}{\bbA}{1}(x)\bigr)=y+r\scre \bbA\bigl(\Theta_{L-3}\scrN^{L-3,\Theta,\varTheta}(W x+z)+\varTheta_{L-3}\bigr)\\
							&=y+r\scre \scrN^{L-2,\Theta,\varTheta}(W x+z).
						\end{split}
				\end{equation}}
				\argument{\lref{eq3};}{\cref{item 2: ANN cover}\dott}
			\end{aproof}
             \cfclear
			\begin{athm}{lemma}{Shallow ANN cover}
				Let $\ell=(\ell_0,\ell_1,\ell_2)\in \N^3$, let $\mathbb A\colon\R\to\R$ be a function, 
				and let $r\in \R^{\ell_2}$, $\W\in \R^{1\times \ell_0}$, $z\in \R$, $y,\bfe\in \R^{\ell_2}$\cfload.
				Then there exists $\theta\in \R^{\ffd(\ell)}$ which satisfies for all $x\in \R^{\ell_0}$ that
				\begin{equation}\llabel{conclude}
					\rel{\ell}{\theta}{\bbA}(x)=y+r\bfe\bbA( 
					\W x+z)\ifnocf.
				\end{equation}
               \cfout[.]
			\end{athm}
			\begin{aproof}
				Throughout this proof let $W=(W_1,\dots,W_{\ell_0})$, $\Theta=(\Theta_1,\dots,\Theta_{L-2})\in \R^{L-2}$, $\varTheta=(\varTheta_1,\dots,\varTheta_{L-2})\in \R^{L-2}$, $\bfy=(\bfy_1,\dots,\bfy_{\ell_L})\in \R^{\ell_L}$, $\scre=(\scre_1,\dots,\scre_{\ell_2})\in \R^{\ell_L}$ satisfy
				\begin{equation}\llabel{def}
					W=\W^T,\qquad \Theta=\bftheta,\qquad \varTheta=\bfvartheta,\qquad \bfy=y, \qqandqq \scre=\bfe
				\end{equation}
				and let $\theta=(\theta_1,\dots,\theta_{\ffd(\ell)})\in \R^{\ffd(\ell)}$ satisfy that
				\begin{enumerate}[label=(\roman*)]
					\item \llabel{item1} it holds for all $i\in \{1,2,\dots,\ell_0\}$ that $\theta_i=W_i$,
					\item \llabel{item2} it holds that $\theta_{\ell_1\ell_0+1}=z$,
					\item \llabel{item3} it holds for all $i\in \{1,2,\dots,\ell_2\}$ that $\theta_{1+(i-1)\ell_{1}+\ell_1(\ell_{0}+1)}=r\scre_i$,
					\item \llabel{item4} it holds for all $i\in \{1,2,\dots,\ell_L\}$ that $\theta_{\ell_{2}\ell_{1}+i+\ell_1(\ell_{0}+1)}=\bfy_i$, and
					\item \llabel{item5} it holds for all $i\in\bigl(\{1,2,\dots,\ffd(\ell)\}\backslash\bigl( \cup_{j=1}^{\ell_0}\cup_{k=0}^{\ell_2}\bigl\{\ell_1\ell_{2}+\max\{k,1\},\ell_0\ell_1+\ell_1 k+1,j\bigr\}\bigr)\bigr)$ that $\theta_i=0$.
				\end{enumerate}
				\argument{\cref{realization multi2};\lref{item1,item2,item5}}{that for all $i\in \{1,2,\dots,\ell_1\}$, $x\in \R^{\ell_0}$ it holds that
					\begin{equation}\llabel{eq2}
						\relii{\ell}{1}{\theta}{\bbA}{i}(x)= (W x+z)\mathbbm 1_{\{1\}}(i)\dott
				\end{equation}}
				\argument{\cref{realization multi2};\lref{eq2};\lref{item3,item4,item5}}{for all $i\in \{1,2,\dots,\ell_2\}$, $x\in \R^{\ell_0}$ that
					\begin{equation}\llabel{eq3}
						\reli{\ell}{2}{\theta}{\bbA}(x)=y+r\bfe\bbA\bigl(\relii{\ell}{1}{\theta}{\bbA}{1}(x)\bigr)=y+r\bfe \bbA(W x+z)\dott
				\end{equation}}
				\argument{\lref{eq3};}{\lref{conclude}\dott}
			\end{aproof}
             \cfclear
			\begin{athm}{cor}{verify maintheorem 5}
				Let $L\in \N\backslash\{1\}$, $\ell=(\ell_0,\ell_1,\dots,\ell_{L}) \in \N^{L+1}$, let $ ( \Omega, \mathcal{F}, \P) $ be a probability space, let
				$ a \in \R $, 
				$ b \in (a, \infty)  $,
				let 
				$ X \colon \Omega \to [a,b]^{\ell_0} $
				and 
				$ Y \colon \Omega \to \R^{\ell_{L}} $
				be random variables, let $\mathbb A\in C(\R,\R)$, $\alphaw\in \R$, $\betaw\in (\alphaw,\infty)$ satisfy $\sup_{ x \in ( \alphaw, \betaw ) } | \mathbb A(x) - \mathbb A( \alphaw ) | = 0<\sup_{x\in \R}|\mathbb A(x)-\mathbb A(0)|$,
				let $\smalll \colon  \R^\delta \times \R^\delta \to \R $ be measurable, let $U\subseteq \R$ be open, assume $0\in U$, let $y,\bfe\in \R^\delta$ satisfy for all $w\in \R^\delta$ that $(U\ni r\mapsto \smalll(y+r\bfe,w)\in \R)\in C^1(U,\R)$,
				let $\g\colon U\times\R^\delta\to\R$ satisfy for all $r\in U$, $w\in \R^\delta$ that $\g(r,w)=\frac{\d}{\d r} \smalll(y+r\bfe,w)$, assume 
				\begin{equation}\label{assume1: a}
					\textstyle \E\bigl[ | \smalll(y,Y) | + \sup_{ r \in U } | \g(r,Y) | \bigr] < \infty,
				\end{equation}
				assume $\E\bigl[|\E[\g(0,Y)|X]|\bigr]>0$, and let $\cP\subseteq\R^{\ffd(\ell)}$ satisfy $\cP=\bigl\{\theta\in \R^{\ffd(\ell)}\colon \E\bigl[|\smalll(\rel{\ell}{\theta}{\bbA}(X),Y)|\bigr]<\infty\bigr\}$ \cfload.
				Then
				\begin{equation}\label{conclude: main improve risk}
					\inf_{\theta \in \cP}\E\bigl[\smalll(\rel{\ell}{\theta}{\bbA}(X),Y)\bigr]<\E\bigl[\smalll(y,Y)\bigr]\ifnocf.
				\end{equation}
                \cfout[.]
			\end{athm}
			\begin{aproof}
				In our proof of \cref{conclude: main improve risk} we distinguish between the case $L=2$ and the case $L\geq 3$. We first prove \cref{conclude: main improve risk} in the case
				\begin{equation}\llabel{case 1}
					L=2.
				\end{equation}
				\argument{the assumption that $\sup_{ x \in ( \alphaw, \betaw ) } | \mathbb A(x) - \mathbb A( \alphaw ) | = 0<\sup_{x\in \R}|\mathbb A(x)-\mathbb A(0)|$}{that \llabel{arg2.5} $\bbA$ is not a polynomial\dott}
				\argument{\lref{arg2.5};\cref{lem: globally risk} (applied with $d\curvearrowleft \ell_0$, $\delta \curvearrowleft \ell_L$, $a\curvearrowleft a$, $b\curvearrowleft b$, $X\curvearrowleft X$, $Y\curvearrowleft Y$, $\smalll\curvearrowleft \smalll$, $U\curvearrowleft U$, $y\curvearrowleft y$, $\bfe\curvearrowleft \bfe$, $\g\curvearrowleft\g$, $\mathbb A\curvearrowleft\mathbb A$  in the notation of \cref{lem: globally risk});}{that there exist $\varepsilon,\varrho\in (0,\infty)$ which satisfy
					\begin{equation}\llabel{eq1}
						\textstyle\E\biggl[\sup\limits_{r\in [-\varepsilon,\varepsilon]}\sup\limits_{\W\in [-\varrho,\varrho]^{1\times \ell_0}}\sup\limits_{z\in [-\varrho,\varrho]}|\smalll(y+r\bfe\mathbb A(\W X + z),Y)|\biggr]<\infty
					\end{equation}
					and  
					\begin{equation}\llabel{eq2}
						\inf_{r\in [-\varepsilon,\varepsilon] }
						\inf_{\W\in [-\varrho,\varrho]^{1\times \ell_0}}
						\inf_{z\in [-\varrho,\varrho]}
						\E\bigl[ \smalll(y+r\bfe\mathbb A(\W X + z),Y)  \bigr]<\E\bigl[\smalll(y,Y)\bigr].
				\end{equation}}
				\argument{\cref{Shallow ANN cover};}{for all $ r\in [-\varepsilon,\varepsilon],\,\W\in [-\varrho,\varrho]^{1\times \ell_0},\, z\in [-\varrho,\varrho]$ that there exists $\theta\in \R^{\ffd(\ell)}$ such that
					\begin{equation}\llabel{eq2.5}
						\reli{\ell}{2}{\theta}{\bbA}(X)=y+r\bfe \bbA(\W X+z)\dott
				\end{equation}}
				\argument{\lref{eq2.5};\lref{eq1}}{that
					\begin{multline}
						\bigl\{\theta\in \R^{\ffd(\ell)}\colon \bigl(\exists\, r\in [-\varepsilon,\varepsilon],\,\W\in [-\varrho,\varrho]^{1\times \ell_0},\, z\in [-\varrho,\varrho]\colon\\ \rel{\ell}{\theta}{\bbA}(X)=y+r\bfe \bbA(\W X+z)\bigr)\bigr\}\subseteq \cP
					\end{multline}
					and
					\begin{equation}\llabel{eq3}
						\begin{split}
							\inf_{\theta\in \cP} \E\bigl[\smalll(\reli{\ell}{2}{\theta}{\bbA}(X),Y)\bigr]\leq \inf_{r\in [-\varepsilon,\varepsilon] }
							\inf_{\W\in [-\varrho,\varrho]^{1\times \ell_0}}
							\inf_{z\in [-\varrho,\varrho]}
							\E\bigl[ \smalll(y+r\bfe\mathbb A(\W X + z),Y)  \bigr].
						\end{split}
				\end{equation}}
				\argument{\lref{eq3};\lref{eq2}}{\cref{conclude: main improve risk} in the case $L=2$\dott}
				Next we prove \cref{conclude: main improve risk} in the case
				\begin{equation}\llabel{case2}
					L\geq 3.
				\end{equation}
				In the following for every $\theta=(\theta_1,\dots,\theta_{L-2})$, $\vartheta=(\vartheta_1,\dots,\vartheta_L)\in\R^{L-2}$ let $\scrN^{ v, \theta,\vartheta }\colon \R \to \R $, $v \in \{0,1,\dots,L-2\}$, satisfy for all $v\in \{0,1,\dots,L-2\}$, $x\in \R$ that
				\begin{equation}\llabel{def: scrN}
					\scrN^{v, \theta,\vartheta }( x ) = \mathbb A\bigl((\theta_v\scrN^{v-1,\theta,\vartheta}(x)+\vartheta_{v})\indicator{\N}(v)+x\indicator{\{0\}}(v)\bigr).
				\end{equation}
				\startnewargseq
				\argument{\cref{item 1: ANN cover} in \cref{ANN cover};}{that there exist $\Theta,\varTheta\in \R^{L-2}$ which satisfy
					\begin{equation}\llabel{eqq1}
						\sup_{x\in \R}|\scrN^{L-2, \theta,\vartheta }( x )-\scrN^{L-2, \theta,\vartheta }( 0 )|>0=\sup_{x\in (\alphaw,\betaw)}|\scrN^{L-2, \theta,\vartheta }( x )-\scrN^{L-2, \theta,\vartheta }( \alphaw )|.
				\end{equation}}
				\startnewargseq
				\argument{\lref{eqq1};}{that \llabel{arg2} $\R\ni x\mapsto \scrN^{L-2, \theta,\vartheta }(x)\in \R$ is not a polynomial\dott}
				\argument{\cref{lem: globally risk} (applied with $d\curvearrowleft \ell_0$, $\delta \curvearrowleft \ell_L$, $a\curvearrowleft a$, $b\curvearrowleft b$, $X\curvearrowleft X$, $Y\curvearrowleft Y$, $\smalll\curvearrowleft \smalll$, $U\curvearrowleft U$, $y\curvearrowleft y$, $\bfe\curvearrowleft \bfe$, $\g\curvearrowleft\g$, $\mathbb A\curvearrowleft\mathbb \scrN^{L-2,\Theta,\varTheta}$ in the notation of \cref{lem: globally risk});\lref{arg2}}{that there exist $\varepsilon,\varrho\in (0,\infty)$ which satisfy
					\begin{equation}\llabel{eqq2}
						\textstyle\E\biggl[\sup\limits_{r\in [-\varepsilon,\varepsilon]}\sup\limits_{\W\in [-\varrho,\varrho]^{1\times \ell_0}}\sup\limits_{z\in [-\varrho,\varrho]}|\smalll(y+r\bfe\scrN^{L-2,\Theta,\varTheta}(\W X + z),Y)|\biggr]<\infty
					\end{equation}
					and  
					\begin{equation}\llabel{eqq3}
						\inf_{r\in [-\varepsilon,\varepsilon] }
						\inf_{\W\in [-\varrho,\varrho]^{1\times \ell_0}}
						\inf_{z\in [-\varrho,\varrho]}
						\E\bigl[ \smalll(y+r\bfe\scrN^{L-2,\Theta,\varTheta}(\W X + z),Y)  \bigr]<\E\bigl[\smalll(y,Y)\bigr].
				\end{equation}}
				\startnewargseq
				\argument{\cref{item 2: ANN cover} in \cref{ANN cover};}{for all $ r\in [-\varepsilon,\varepsilon],\,\W\in [-\varrho,\varrho]^{1\times \ell_0},\, z\in [-\varrho,\varrho]$ that there exists $\theta\in \R^{\ffd(\ell)}$ such that
					\begin{equation}\llabel{eqq2.5}
						\rel{\ell}{\theta}{\bbA}(X)=y+r\bfe \scrN^{L-2,\Theta,\varTheta}(\W X+z)\dott
				\end{equation}}
				\argument{\lref{eqq2.5};\lref{eqq2}}{that
					\begin{multline}
						\bigl\{\theta\in \R^{\ffd(\ell)}\colon \bigl(\exists\, r\in [-\varepsilon,\varepsilon],\,\W\in [-\varrho,\varrho]^{1\times \ell_0},\, z\in [-\varrho,\varrho]\colon\\ \rel{\ell}{\theta}{\bbA}(X)
						=y+r\bfe \scrN^{L-2,\Theta,\varTheta}(\W X+z)\bigr)\bigr\}\subseteq \cP
					\end{multline}
					and
					\begin{equation}\llabel{eqq4}
						\begin{split}
							\inf_{\theta\in \cP} \E\bigl[\smalll(\rel{\ell}{\theta}{\bbA}(X),Y)\bigr]\leq \inf_{r\in [-\varepsilon,\varepsilon] }
							\inf_{\W\in [-\varrho,\varrho]^{1\times \ell_0}}
							\inf_{z\in [-\varrho,\varrho]}
							\E\bigl[ \smalll(y+r\bfe\scrN^{L-2,\Theta,\varTheta}(\W X + z),Y)  \bigr].
						\end{split}
				\end{equation}}
				\argument{\lref{eqq4};\lref{eqq3}}{\cref{conclude: main improve risk} in the case $L\geq 3$\dott}
			\end{aproof}
            \section{Concrete lower bounds for the non-convergence probability}\label{sec: application}
            In this section we combine the main result of \cref{sec: general analysis}, \cref{main theorem general} in \cref{subsec: general analysis final}, with the main result of \cref{sec: improve risk}, \cref{verify maintheorem 5} in \cref{sucsec: improve risk for deep ANN} to establish in \cref{main cor: scientific general square} a general lower bound for the probability that the true risk (cf.\ \cref{loss: scientific} in \cref{setting: scientific learning}) of a general \SGD\ optimization process (cf.\ \cref{SGD: scientific} in \cref{setting: scientific learning}) does not converge to the optimal/infimal true risk value (cf.\ \cref{conclude: estimate1} in \cref{main cor: scientific general square}). Furthermore, in \cref{sec: application} we apply \cref{main cor: scientific general square} to establish the concrete non-convergence results in 
            \begin{enumerate}[label=(\roman*)]
            \item
            \cref{main cor: scientific square mean} (non-convergence with strictly positive probability for the standard mean squared error loss; see \cref{def: bbLa} and \cref{conclude: squaremean1}, \cref{main cor: scientific general loss very deep} (non-convergence with strictly positive probability for a class of general loss functions; see \cref{assume1a} and \cref{conclude: other loss1})),
            \item \cref{main theo: scientific square mean error very deep} (non-convergence with high probability for the standard mean squared error loss; see \cref{conclude: mean square loss2}), and
            \item  \cref{main theo: scientific other error very deep} (non-convergence with high probability for a class of general loss functions; see \cref{def: bbLaa}, \cref{assume1aa}, and \cref{concludeaaa}).
            \end{enumerate}
            \cref{main theorem1} in \cref{subsec: main theorem 1} in the introduction is a direct consequence of \cref{main cor: scientific square mean} and \cref{main theorem2} in \cref{subsec: main theorem 2} in the introduction is direct consequence of \cref{main theo: scientific square mean error very deep}.

            \subsection{General lower bounds for the non-convergence probability}\label{subsec: lower bound}
            \subsubsection{Mathematical framework for the training of deep ANNs}
            \cfclear
			\begin{setting}\label{setting: scientific learning}
				Let $L\in \N\backslash\{1\}$, $\ell =(\ell_0,\ell_1,\dots,\ell_{L}) \in \N^{L+1}$, let $ ( \Omega, \mathcal{F}, \P) $ be a probability space, let
				$ a \in \R $, 
				$ b \in [a, \infty)  $, for every $ m, n \in \N_0 $ 
				let 
				$ X^m_n \colon \Omega \to [a,b]^{\ell_0} $
				and 
				$ Y^m_n \colon \Omega \to \R^{\ell_{L}} $
				be random variables, let $S\subseteq \R$ be finite, for every $r\in \N_0$ let $\mathbb A_r\in C^{\min\{r,1\}}(\R,\R)$, let $\grad\colon \R\to\R$ satisfy for all $x\in \R$ that there exists $R\in \N$ such that $ \restr{\mathbb A_0}{\R \backslash S} \in C^1(\R \backslash S,\R)$, $\restr{\grad}{\R \backslash S}=(\restr{\mathbb A_0}{\R \backslash S})'$, and
				\begin{equation}\llabel{def: g_r}
					\textstyle
					\sum_{r =R}^\infty\bigl(|\mathbb A_r(x)-\mathbb A_0(x)|+|(\mathbb A_r)'(x)-\grad(x)|\bigr)=0,
				\end{equation}
				 let $\betaw \in \R$, $\alphaw \in [-\infty,\betaw)$ satisfy $\sup_{ x \in ( \alphaw, \betaw )\backslash S } | \grad(x)| = 0<\sup_{x\in\R\backslash S}|\grad(x)|$, for every $n\in \N_0$
				let $\bbL_n\in C^{\min\{n,1\}}(\R^{ \ell_L } \times \R^{ \ell_L },\R)$, $M_n\in \N$, let $\fL\colon \R^{\ffd(\ell)}\to \R$ satisfy for all $\theta\in \R^{\ffd(\ell)}$ that $\E\bigl[|\bbL_0(\rel{\ell}{\theta}{\bbA_0}(X^0_0),Y^0_0)|\bigr]\allowbreak<\infty$ and
				\begin{equation}\label{loss: scientific}
					\fL(\theta)=\E\bigl[\bbL_0(\rel{\ell}{\theta}{\bbA_0}(X^0_0),Y^0_0)\bigr],
				\end{equation}
				 for every $r,n\in \N_0$ let
				$ 
				\cLnri{n}{r} \colon \R^{ \ffd(\ell) } \times \Omega \to \R 
				$
				satisfy for all 
				$ \theta \in \R^{ \ffd(\ell) }$
				that
				\begin{equation}\label{objective: scientific}
					\displaystyle
					\cLnri{n}{r}( \theta) 
					= 
					\frac{ 1 }{ M_n } 
					\biggl[ \textstyle
					\sum\limits_{ m = 1 }^{ M_n} 
					\bbL_n(\rel{\ell}{\theta}{\bbA_r}(X_n^m),Y_n^m)
					\biggr]
					,
				\end{equation}
				for every $n\in \N_0$ let 
				$ 
				\cG_n  
				\colon \R^{ \ffd(\ell)} \times \Omega \to \R^{ \ffd(\ell) } 
				$ 
				satisfy for all $\omega\in \Omega$, $\theta\in \{\vartheta\in \R^{\ffd(\ell)}\colon (\nabla_{\vartheta} \cLnri{n}{r}(\vartheta,\omega))_{r\in \N}$ is convergent$\}$
				that
				\begin{equation}\label{gradient: scientific}
					\cG_n( \theta,\omega) 
					= 
					\lim_{r\to\infty}\bigl[\nabla_\theta \cLnri{n}{r}(\theta,\omega)\bigr]
				\end{equation}
				and let 
				$
				\Theta_n=(\xTheta{n}{1},\dots,\xTheta{n}{\ffd(\ell)}) 
				\colon \Omega  \to \R^{\ffd(\ell) }
				$
				be a random variable, and for every $n\in \N$ let 
				$
				\Phi_n 
				= 
				( 
				\Phi^{ 1 }_n, \dots, 
				\Phi^{ \ffd(\ell) }_n 
				)
				\colon 
				\allowbreak
				( \R^{ \ffd(\ell)} )^{ 2n }
				\allowbreak
				\to 
				\R^{ \ffd(\ell) }
				$ 
				satisfy 
				for all 
				$
				g =
				( 
				( g_{ i, j } )_{ j \in \{ 1, 2, \dots, \ffd(\ell) \} }
				)_{
					i \in \{ 1, 2, \dots, 2n\}
				}
				\in 
				(
				\R^{ 
					\ffd(\ell)
				}
				)^{ 2n }
				$, 
				$ 
				j \in \{1,2,\dots,\ffd(\ell)\}  
				$
				with $
				\sum_{ i = 1 }^{2n}
				\abs{ g_{ i, j } -g_{1,j}\mathbbm 1_{[1,n]}(i)}
				= 0
				$
				that 
				$
				\Phi^{ j }_n( g ) = g_{n,j} 
				$ and
				assume 
				\begin{equation}
					\label{SGD: scientific}
					\Theta_{ n  } 
					=  
					\Phi_{n}\bigl(\Theta_0,\Theta_1,\dots,\Theta_{n-1},
					\cG_1( \Theta_0  ) ,
					\cG_2( \Theta_1  ) ,
					\dots ,
					\cG_n ( \Theta_{n-1} )
					\bigr)\ifnocf.
				\end{equation} 
                \cfload[.]
			\end{setting}
            \subsubsection{General lower bounds for the non-convergence probability}
             \cfclear
            \begin{athm}{prop}{main cor: scientific verify}
                Assume \cref{setting: scientific learning}, let $\psi\in C^1([0,\infty),\R)$ be strictly increasing, assume for all $x,y \in \R^{\ell_L}$ that 
				\begin{equation}\llabel{def: bbL}
					\bbL_0(x,y)=\psi(\|x-y\|^2),
				\end{equation}
				and let $\varepsilon\in (0,\infty)$ satisfy for all $y\in \R^{\ell_L}$ that
				\begin{equation}\llabel{assume1}
					\sup_{x\in \R^{\ell_L},\, \|x-y\|\leq \varepsilon}  \E[\|(x-Y_0^0) \psi'(\|x-Y_0^0\|^2)\|]<\infty
				\end{equation}
				and
				\begin{equation}\llabel{assume}
					\E\bigl[\bigl\|\E[(y-Y_0^0) \psi'(\|y-Y_0^0\|^2) |X_0^0]\bigr\|^2\bigr]>0\ifnocf.
				\end{equation}
                \cfload[.]
				Then
				\begin{equation}\llabel{conclude}
				    \inf_{\theta\in \R^{\ffd(\ell)}}\fL(\theta)<\inf_{y\in \R^{\ell_L}}\E\bigl[\bbL_0(y,Y_0^0)\bigr]\ifnocf.
				\end{equation}
                \cfout[.]
            \end{athm}
            \begin{aproof}
                Throughout this proof for every $\omega\in \Omega$ let $\mu_\omega\colon \mathcal F\to [0,\infty]$ satisfy for all $A\in \mathcal F$ that
				\begin{equation}\llabel{def: mu}
					\mu_\omega(A)=\P(A)
				\end{equation}
				and let $\cX=(\cX_1,\dots,\cX_{\ell_0})\colon \Omega\to \R^{\ell_0}$ and $\cY=(\cY_1,\dots,\cY_{\ell_L})\colon \Omega\to \R^{\ell_L}$ satisfy $\cX=X_0^0$ and $\cY=Y_0^0$.
				\argument{\lref{def: bbL}; the assumption that $\psi$ is strictly increasing; Fatou's lemma}{that for all $c\in \R^{\ell_L}$ it holds that
					\begin{equation}\llabel{eq1}
						\liminf_{\|y\|\to \infty}\E[\bbL_0(y,Y)]=\liminf_{\|y\|\to \infty}\E[\psi(\|y-Y\|^2)] \geq \sup_{x\in \R} \psi(x)\geq \E[\psi(\|c-Y\|^2)]=\E[\bbL_0(c,Y)].
				\end{equation}}
				\argument{\lref{eq1};}{that there exists $c=(c_1,\dots,c_{\ell_L})\in \R^{\ell_L}$ which satisfies
					\begin{equation}\llabel{def: c}
					\E[\bbL_0(c,\cY)]=\inf_{y\in \R^{\ell_L}}\E[\bbL_0(y,\cY)].
				\end{equation}}
				\startnewargseq
				\argument{\lref{assume};}{that there exists $i\in 
					\{1,2,\dots,\ell_L\}$ which satisfies 
					\begin{equation}\llabel{def: i}
						\E\bigl[\bigl|\E[(c_i-\cY_i) \psi'(\|c-\cY\|^2) |\cX]\bigr|\bigr]>0.
				\end{equation}}
				In the following let $\bfe\in \R^{\ell_L}$ satisfy for all $x=(x_1,x_2,\dots,x_{\ell_L})\in \R^{\ell_L}$ that
				\begin{equation}\llabel{def: bfe}
					\spro{\bfe, x}=x_i
				\end{equation}
				and let $\g\colon (-\varepsilon,\varepsilon)\times\R^{\ell_L}\to\R$ satisfy for all $r\in (-\varepsilon,\varepsilon)$, $w\in \R^{\ell_L}$ that
				\begin{equation}\llabel{def: g}
					\g(r,w)=\frac{\d}{\d r} \bbL_0(c+r\bfe,w).
				\end{equation}
				\startnewargseq
				\argument{\lref{def: bbL};\lref{def: bfe};\lref{def: g}}{that for all $r\in (-\varepsilon,\varepsilon)$ that
					\begin{equation}\llabel{eq2}
						\g(r,\cY)=2(c_i+r-\cY_i)\psi'(\|c+r\bfe-\cY\|^2).
				\end{equation}}
				\argument{\lref{eq2};\lref{assume}}{that
					\begin{equation}\llabel{verify 1}
						\E[|\g(0,\cY)|\cX|]>0.
				\end{equation}}
				\argument{\cref{loss: scientific};}{that \llabel{arg1} $\E\bigl[ | \bbL_0(c,\cY) |]<\infty$\dott}
				\argument{\lref{arg1};\lref{assume1};\lref{eq2}}{that
					\begin{equation}\llabel{verify 2}
						\textstyle \E\bigl[ | \bbL_0(c,\cY) | + \sup_{ r \in (-\varepsilon,\varepsilon)} | \g(r,\cY) | \bigr] < \infty,
				\end{equation}}
				\argument{\cref{verify maintheorem 5} (applied with $\fd\curvearrowleft \ffd(\ell)$, $L\curvearrowleft L$, $(\ell_i)_{i\in \{0,1,\dots,L\}}\curvearrowleft (\ell_i)_{i\in \{0,1,\dots,L\}}$, $\Omega\curvearrowleft \Omega$, $\mathcal F\curvearrowleft \mathcal F$, $\P\curvearrowleft\P$, $a\curvearrowleft a$, $b\curvearrowleft b$, $X\curvearrowleft\cX$, $Y\curvearrowleft \cY$, $\bbA\curvearrowleft\bbA$, $\alphaw\curvearrowleft\alphaw$, $\betaw\curvearrowleft\betaw$, $(\reli{\ell}{v}{\theta}{\bbA_0})_{v\in \{0,1,\dots,L\},\,\theta\in \R^\fd}\curvearrowleft(\reli{\ell}{v}{\theta}{\bbA_0})_{v\in \{0,1,\dots,L\},\,\theta\in \R^{\ffd(\ell)}}$, $\smalll\curvearrowleft\bbL_0$, $U\curvearrowleft(-\varepsilon,\varepsilon)$, $y\curvearrowleft c$, $\bfe\curvearrowleft\bfe$, $\g\curvearrowleft\g$ in the notation of \cref{verify maintheorem 5}) ;\cref{loss: scientific};\lref{def: c};\lref{verify 1};\lref{verify 2}}{that for all $\theta\in \R^{\ffd(\ell)}$ it holds that $\E\bigl[|\bbL_0(\rel{\ell}{\theta}{\bbA_0}(\cX),\cY)|\bigr]\allowbreak<\infty$ and
					\begin{equation}\llabel{eq7}
						\inf_{\vartheta \in \R^{\ffd(\ell)}}\E\bigl[\bbL_0(\rel{\ell}{\vartheta}{\bbA_0}(\cX),\cY)\bigr]<\E\bigl[\bbL_0(c,\cY)\bigr].
				\end{equation}}
                \argument{\cref{loss: scientific};\lref{eq7};\lref{def: c}}{\lref{conclude}\dott}
            \end{aproof}
             \cfclear
			\begin{athm}{theorem}{main cor: scientific general square}
				Assume \cref{setting: scientific learning}, for every $k\in \{0,1,\dots,L\}$ let $\bfd_k\in \N$ satisfy $\bfd_k=\sum_{i=1}^{k}\ell_i(\ell_{i-1}+1)$, let $\psi\in C^1([0,\infty),\R)$ be strictly increasing, assume for all $x,y \in \R^{\ell_L}$ that 
				\begin{equation}\llabel{def: bbL}
					\bbL_0(x,y)=\psi(\|x-y\|^2),
				\end{equation}
				let $\varepsilon\in (0,\infty)$ satisfy for all $y\in \R^{\ell_L}$ that
				\begin{equation}\llabel{assume1}
					\sup_{x\in \R^{\ell_L},\, \|x-y\|\leq \varepsilon}  \E[\|(x-Y_0^0) \psi'(\|x-Y_0^0\|^2)\|]<\infty
				\end{equation}
				and
				\begin{equation}\llabel{assume}
					\E\bigl[\bigl\|\E[(y-Y_0^0) \psi'(\|y-Y_0^0\|^2) |X_0^0]\bigr\|^2\bigr]>0,
				\end{equation}
				let $\eta\in \R$, $\zeta\in (\eta,\infty)$ satisfy
						$(\eta,\zeta)\subseteq (\alphaw,\betaw)\backslash S$,
                for every $i\in \{1,2,\dots,\ell_1\}$ let $\varrho_i\in \R$ satisfy
                \begin{equation}\llabel{def: varrho}
               \textstyle \varrho_i= \bigl[\P\bigl(\frac{3\eta+\zeta}{4}<\xTheta{0}{\ell_1\ell_0+i}<\frac{\eta+3\zeta}{4}\bigr)\bigr]\bigl[\prod_{j=1}^{\ell_0}\P\bigl(|\xTheta{ 0}{( i - 1 )\ell_0 + j }|<\frac{\zeta-\eta}{2\ell_0\max\{1,|a|,|b|\}}\bigr)\bigr],  
                \end{equation}
                let $\bfA\in \R$ satisfy $\bfA\leq \inf_{x\in \R}\bbA_0(x)$, let $\gamma\in (-\infty,\min(S\cup\{\betaw\})]$, $\rho\in [-\infty,\infty)$ satisfy
				\begin{equation}\llabel{def: gamma1}
				\rho=\begin{cases}
						-\infty&\colon \bfA\geq 0\\
						\textstyle\frac{1}{\bfA\max_{i\in\{1,2,\dots,L-2\}}\ell_i} &\colon \bfA<0, L>2\\
                        0 &\colon \bfA<0,L=2,
					\end{cases}
				\end{equation}
			 let $\chi\in \N\cap [\ell_1+\ell_0\ell_1,\ffd(\ell)]$, and assume that $\xTheta{0}{i}$, $i\in \{1,2,\dots,\chi\}$, are independent \cfload.
				Then
				\begin{equation}\label{conclude: estimate1}
                \begin{split}
					&\textstyle
					\P\Bigl(\inf\limits_{ n\in \N_0 }  \fL(\NNelll_n)>\inf\limits_{\theta\in \R^{\ffd(\ell)}}\fL(\theta)\Bigr)\\
                   &\geq \textstyle\max\biggl\{\prod\limits_{i=1}^{\ell_1}\varrho_i,\biggl[1-\textstyle \biggl(\prod\limits_{k=2}^{L-1}\biggl[1-\biggl[\prod\limits_{j=1}^{\ell_{k}\ell_{k-1}}\P(\rho<\xTheta{0}{j+\bfd_{k-1}}<0)\biggr]\\
                &\textstyle\cdot\biggl[\prod\limits_{j=1}^{\ell_k}\P\bigl(\xTheta{0}{j+\ell_k\ell_{k-1}+\bfd_{k-1}}<\gamma-\mathbbm 1_{(-\infty,0)}(\bfA)\bigr)\biggr]\biggr]\biggr)\biggr]\mathbbm 1_{\{-\infty\}}(\alphaw)\mathbbm 1_{[\ffd(\ell)-\ell_L\ell_{L-1}-\ell_L,\infty)}(\chi)\biggr\}\ifnocf.
                    \end{split}
				\end{equation}
                \cfout[.]
			\end{athm}
			\begin{aproof}
				\argument{\cref{main theorem general};\cref{main cor: scientific verify}}{\cref{conclude: estimate1}\dott}
			\end{aproof}
\subsubsection{Lower bounds for the non-convergence probability for one-dimensional outputs}
 \cfclear
            \begin{athm}{lemma}{conditional expectation verify}
				Let $(\Omega,\cF,\P)$ be a probability space, let $d\in \N$, let $X\colon \Omega\to \R^d$ and $Y\colon \Omega\to \R$ random variables, let $\varphi\colon \R^d\to\R$ be measurable, assume that $\varphi(X)$ and $Y-\varphi(X)$ are independent, assume for all $x\in \R$ that $\P(\varphi(X)=x)<1$, let $\psi\in C^1([0,\infty),\R)$ be strictly increasing, assume that $(0,\infty)\ni x\mapsto \psi'(x)\sqrt{x}\in \R$ is strictly increasing, and assume for all $y\in \R$ that $\E\bigl[|(y-Y) \psi'(|y-Y|^2)|\bigr]<\infty$ \cfload. Then it holds for all $y\in \R$ that 
				\begin{equation}\llabel{conclude}
					\E\bigl[|\E[(y-Y) \psi'(|y-Y|^2) |X]|^2\bigr]>0\ifnocf.
				\end{equation}
			\end{athm}
            \cfout[.]
			\begin{aproof}
            Throughout this proof let $\cP\subseteq \R$ satisfy 
            \begin{equation}\llabel{def: P}
                \cP=\bigl\{x\in \R\colon \E\bigl[|(y-\xi-x)\psi'(|y-\xi-x|^2)|\bigr]<\infty\bigr\}
            \end{equation} 
            and let $Z\colon \Omega\to \R$ and $\xi\colon \Omega\to \R$ satisfy $Z=\varphi(X)$ and $\xi=Y-Z$.
				We prove \lref{conclude} by contradiction. In the following we thus assume that there exists $y\in \R$ which satisfies
				\begin{equation}\llabel{assume}
					\E\bigl[|\E[(y-Y) \psi'(|y-Y|^2) |X]|^2\bigr]=0.
				\end{equation}
				\startnewargseq
				\argument{\lref{assume};}{that 
					\begin{equation}\llabel{eq1}
						\E\bigl[|\E[(y-Y) \psi'(|y-Y|^2) |Z]|^2\bigr]=0.
				\end{equation}}
                In the following let $\Psi\colon\R\to \R$ satisfy for all $x\in \R$ that
                \begin{equation}\llabel{def: Psi}
                \Psi(x)=
                \begin{cases}
                \E[(y-\xi-x)\psi'(|y-\xi-x|^2)],& x\in \cP\\
                    0,&x\notin \cP 
                \end{cases}
                \end{equation}
                \startnewargseq
				\argument{\cite[Corollary 2.9]{ArBeArPhi2018};\lref{def: P};\lref{def: Psi};the assumption that for all $y\in \R$ it holds that $\E\bigl[|(y-Y) \psi'(|y-Y|^2)|\bigr]<\infty$;the fact that $\xi$ and $Z$ are independent;}{that 
                \begin{equation}\llabel{eq2'}
                   \P\bigl( Z \in \cP\bigr)=1 
                \end{equation}
                and it holds $\P$-a.s.\ that
					\begin{equation}\llabel{eq2}
						\begin{split}
							\E[(y-Y)\psi'(|y-Y|^2)|Z]= \E[(y-\xi-Z)\psi'(|y-\xi-Z|^2)|Z]=\Psi(Z).
						\end{split}
				\end{equation}}
				\argument{\lref{eq2};\lref{def: Psi};the change of variable formula for pushforward measures}{that 
					\begin{equation}\llabel{eq3}
						\begin{split}
							&\E\bigl[|\E[(y-Y)\psi'(|y-Y|^2)|Z]|\bigr]=\E\bigl[|\Psi(Z)|\bigr]\\
				&=\int_{\R}|\Psi(x)|\,\P_{Z}(\d x)=\int_{\R}| \E[(y-\xi-x)\psi'(|y-\xi-x|^2)]|\,\P_{Z}(\d x).
						\end{split}
				\end{equation}}
                \argument{\lref{def: P};the assumption that $(0,\infty)\ni x\mapsto \psi'(x)\sqrt{x}\in \R$ is strictly increasing;}{that
                \begin{equation}\llabel{eq4}
                    \#\cP\leq 1\dott
                \end{equation}}
               \argument{the fact that for all $c\in \R$ it holds that $\P(Z=c)<1$;\lref{eq3};\lref{eq4}}{that
                \begin{equation}\llabel{eq5}
                    0=\E\bigl[|\E[(y-Y)\psi'(|y-Y|^2)|Z]|\bigr]=\int_{\R}| \E[(y-\xi-x)\psi'(|y-\xi-x|^2)]|\,\P_{Z}(\d x)>0\dott
                \end{equation}}
                This contradiction shows \lref{conclude}\dott
			\end{aproof}
             \cfclear
                        \begin{athm}{cor}{scientific 1 dimension ouput general loss}
                Assume \cref{setting: scientific learning}, for every $k\in \{0,1,\dots,L\}$ let $\bfd_k\in \N$ satisfy $\bfd_k=\sum_{i=1}^{k}\ell_i(\ell_{i-1}+1)$, assume $\ell_L=1$, let $\psi\in C^1([0,\infty),\R)$ be strictly increasing, assume for all $x,y \in \R$ that 
				\begin{equation}\llabel{def: bbL}
					\bbL_0(x,y)=\psi(|x-y|^2),
				\end{equation}
             assume that $(0,\infty)\ni x\mapsto \psi'(x)\sqrt{x}\in \R$ is strictly increasing, let $\varphi\colon \R^{\ell_0}\to\R$ be measurable, assume that $\varphi(X_0^0)$ and $Y_0^0-\varphi(X_0^0)$ are independent, 
				assume for all $x\in \R$ that 
				\begin{equation}\llabel{assume1}
					\P(\varphi(X_0^0)=x)<1\qqandqq \E\bigl[|Y_0^0|\psi'((|Y_0^0|+|x|)^2)|\bigr]<\infty,
				\end{equation}
				 let $\eta\in \R$, $\zeta\in (\eta,\infty)$ satisfy
						$(\eta,\zeta)\subseteq (\alphaw,\betaw)\backslash S$,
               for every $i\in \{1,2,\dots,\ell_1\}$ let $\varrho_i\in \R$ satisfy
                \begin{equation}\llabel{def: varrho}
               \textstyle \varrho_i= \bigl[\P\bigl(\frac{3\eta+\zeta}{4}<\xTheta{0}{\ell_1\ell_0+i}<\frac{\eta+3\zeta}{4}\bigr)\bigr]\bigl[\prod_{j=1}^{\ell_0}\P\bigl(|\xTheta{ 0}{( i - 1 )\ell_0 + j }|<\frac{\zeta-\eta}{2\ell_0\max\{1,|a|,|b|\}}\bigr)\bigr], 
                \end{equation}
                let $\bfA\in \R$ satisfy $\bfA\leq \inf_{x\in \R}\bbA_0(x)$, let $\gamma\in (-\infty,\min(S\cup\{\betaw\})]$, $\rho\in [-\infty,\infty)$ satisfy
				\begin{equation}\llabel{def: gamma1}
				\rho=\begin{cases}
						-\infty&\colon \bfA\geq 0\\
						\textstyle\frac{1}{\bfA\max_{i\in\{1,2,\dots,L-2\}}\ell_i} &\colon \bfA<0, L>2\\
                        0 &\colon \bfA<0,L=2,
					\end{cases}
				\end{equation}
			  let $\chi\in \N\cap [\ell_1+\ell_0\ell_1,\ffd(\ell)]$, and assume that $\xTheta{0}{i}$, $i\in \{1,2,\dots,\chi\}$, are independent \cfload.
				Then
				\begin{equation}\llabel{conclude}
                \begin{split}
					&\textstyle
					\P\Bigl(\inf\limits_{ n\in \N_0 }  \fL(\NNelll_n)>\inf\limits_{\theta\in \R^{\ffd(\ell)}}\fL(\theta)\Bigr)\\
                   &\geq \textstyle\max\biggl\{\prod\limits_{i=1}^{\ell_1}\varrho_i,\biggl[1-\textstyle \biggl(\prod\limits_{k=2}^{L-1}\biggl[1-\biggl[\prod\limits_{j=1}^{\ell_{k}\ell_{k-1}}\P(\rho<\xTheta{0}{j+\bfd_{k-1}}<0)\biggr]\\
                &\textstyle\cdot\biggl[\prod\limits_{j=1}^{\ell_k}\P\bigl(\xTheta{0}{j+\ell_k\ell_{k-1}+\bfd_{k-1}}<\gamma-\mathbbm 1_{(-\infty,0)}(\bfA)\bigr)\biggr]\biggr]\biggr)\biggr]\mathbbm 1_{\{-\infty\}}(\alphaw)\mathbbm 1_{[\ffd(\ell)-\ell_L\ell_{L-1}-\ell_L,\infty)}(\chi)\biggr\}\ifnocf.
                    \end{split}
				\end{equation}
                \cfout[.]
            \end{athm}
            \begin{aproof}
            \argument{\lref{assume1};the assumption that $\psi\in C^1(\R,\R)$}{that for all $x\in \R$ it holds that
            \begin{equation}\llabel{eqq1}
                \begin{split}
                    &\E\bigl[(|Y_0^0|+|x|)\psi'((|Y_0^0|+|x|)^2)\bigr]\\
                    &= \E\bigl[|Y_0^0|\psi'((|Y_0^0|+|x|)^2)\bigr]+|x|\E\bigl[\psi'((|Y_0^0|+|x|)^2)\bigr]\\
                    &=\E\bigl[|Y_0^0|\psi'((|Y_0^0|+|x|)^2)\bigr]+|x|\E\bigl[\psi'((|Y_0^0|+|x|)^2)\mathbbm 1_{[1,\infty)}(|Y_0^0|)\bigr]\\
                    &+|x|\E\bigl[\psi'((|Y_0^0|+|x|)^2)\mathbbm 1_{[0,1]}(|Y_0^0|)\bigr]\\
                    &\leq (1+|x|)\E\bigl[|Y_0^0|\psi'((|Y_0^0|+|x|)^2)\bigr]+|x|\E\bigl[\psi'((|Y_0^0|+|x|)^2)\mathbbm 1_{[0,1]}(|Y_0^0|)\bigr]\\
                    &<\infty.
                \end{split}
            \end{equation}}
            \argument{\lref{eqq1};the assumption that $(0,\infty)\ni x\mapsto \psi'(x)\sqrt{x}\in \R$ is strictly increasing}{that for all $\varepsilon\in (0,\infty)$, $y\in \R$ it holds that
            \begin{equation}\llabel{eq1}
                \sup_{x\in [y-\varepsilon,y+\varepsilon]}  \E[|(x-Y_0^0) \psi'(|x-Y_0^0|^2)|]<\infty.
            \end{equation}}
                \argument{\lref{eq1};\cref{main cor: scientific general square};\cref{conditional expectation verify}}{\lref{conclude}\dott}
            \end{aproof}
            \subsection{Non-convergence with strictly positive probability}\label{subsec: positive probability}
            \subsubsection{Mean squared error loss function}\label{subsubsec: positive probability mean square error}
             \cfclear
            \begin{athm}{lemma}{Integrability of Y}
                Assume \cref{setting: scientific learning} and assume for all $x,y \in \R^{\ell_L}$ that 
				\begin{equation}\llabel{def: bbL}
					\bbL_0(x,y)=\|x-y\|^2\ifnocf.
				\end{equation}
                \cfload[.]
                Then 
                \begin{equation}\llabel{conclude}
                    \E\bigl[\|Y_0^0\|\bigr]<\infty \ifnocf.
                \end{equation}
                \cfout[.]
            \end{athm}
            \newcommand{\frakw}{\mathfrak{w}}
            \begin{aproof}
            Throughout this proof let $\mu\colon \mathcal B([a,b]^{\ell_0}\times\R^{\ell_L})\times \Omega\to [0,\infty]$ satisfy for all $\set\in \mathcal B([a,b]^{\ell_0}\times\R^{\ell_L})$, $\frakw\in \Omega$ that
            \begin{equation}\llabel{def: mu}
                \mu(\set,\frakw)=\P((X_0^0,Y_0^0)\in A).
            \end{equation}
                \argument{\cref{Lemma confirm} (applied with $\fd\curvearrowleft\ffd(\ell)$, $L\curvearrowleft L$, $(\ell_i)_{i\in \{1,2,\dots,L\}}\curvearrowleft(\ell_i)_{i\in \{1,2,\dots,L\}}\curvearrowleft$, $\Omega\curvearrowleft\Omega$, $\cF\curvearrowleft\cF$, $\P\curvearrowleft \P$, $\bbA\curvearrowleft\bbA_0$, $(\reli{\ell}{v}{\theta}{\bbA})_{v\in\{0,1,\dots,L\},\, \theta\in \R^{\ffd(\ell)}}\curvearrowleft(\reli{\ell}{v}{\theta}{\bbA_0})_{v\in\{0,1,\dots,L\},\, \theta\in \R^{\ffd(\ell)}}$, $\smalll\curvearrowleft \bbL_0$, $\mu\curvearrowleft \mu$, $z\curvearrowleft 0$ in the notation of \cref{Lemma confirm});\cref{loss: scientific};\lref{def: bbL};\lref{def: mu};the change of variable formula for pushforward measures}{that for all $\frakw\in \Omega$ it holds that
                \begin{equation}\llabel{arg1} 
                \begin{split}    \E[\|Y_0^0\|^2]=&\int_{[a,b]^{\ell_0}\times\R^{\ell_L}}\|y\|^2\,\P_{(X_0^0,Y_0^0)}(\d x,\d y)\\
        &=\int_{[a,b]^{\ell_0}\times\R^{\ell_L}}|\bbL_0(0,y)|\,\P_{(X_0^0,Y_0^0)}(\d x,\d y)\\
            &=\int_{[a,b]^{\ell_0}\times\R^{\ell_L}}|\bbL(0,y)|\,\mu(\d x,\d y,\frakw)<\infty.
            \end{split}
                \end{equation}}
               \argument{\lref{arg1};}{\lref{conclude}\dott}
            \end{aproof}
             \cfclear
			\begin{athm}{cor}{main cor: scientific square mean}
				Assume \cref{setting: scientific learning}, assume for all $x,y \in \R^{\ell_L}$ that 
				\begin{equation}\label{def: bbLa}
					\bbL_0(x,y)=\|x-y\|^2,
				\end{equation}
				assume $\P(\E[Y_0^0|X_0^0]=\E[Y_0^0])<1$ (cf.\ \cref{Integrability of Y}),
				for every $i\in\N$ let $ \Dens_i \colon \R \to (0,\infty) $ be measurable, 
and assume for all 
$x_1,x_2,\dots,x_{\ffd(\ell)} \in \R$
that
\begin{equation}\llabel{assume}
  \P \bigl( \cap_{i=1}^{\ell_1+\ell_0\ell_1}
    \big\{
    \xTheta{0}{i} 
    < x_i\big\}
  \bigr)
  =
  \prod\limits_{i=1}^{\ell_1+\ell_0\ell_1}\biggl[\int_{ - \infty }^{x_i} \Dens_i(y) \, \d y\biggr]\ifnocf.
\end{equation} 
\cfload[.]
				Then
				\begin{equation}\label{conclude: squaremean1}
					\textstyle
					\P\Bigl(\inf\limits_{ n\in \N_0 }  \fL(\NNelll_n)>\inf\limits_{\theta\in \R^{\ffd(\ell)}}\fL(\theta)\Bigr)>0 \ifnocf.
				\end{equation}
                \cfout[.]
			\end{athm}
			\begin{aproof}
				\argument{the fact that $\E[\|Y_0^0\|]<\infty$; the triangle inequality}{that for all $\varepsilon\in (0,\infty)$, $y\in \R^{\ell_L}$ it holds that
					\begin{equation}\llabel{evidence 1}
						\sup_{x\in \R^{\ell_L},\, \|x-y\|\leq \varepsilon} \E[\|x-Y_0^0\|]\leq \|y\|+\varepsilon +\E[\|Y_0^0\|]<\infty.
				\end{equation}}
				In the following we prove that for all $y\in \R^{\ell_L}$ it holds that
				\begin{equation}\llabel{evidence 2}
					\E[\|\E[y-Y_0^0|X_0^0]\|]>0\dott
				\end{equation}
				\startnewargseq
				We prove \lref{evidence 2} by contradiction. In the following we thus assume that there exists $y\in \R^{\ell_L}$ which satisfies
				\begin{equation}\llabel{assume evidence 2}
					\E[\|\E[y-Y_0^0|X_0^0]\|]=0.
				\end{equation}
				\argument{\lref{assume evidence 2};}{that \llabel{arg1} $\E[y-Y_0^0]=0$\dott}
				\argument{\lref{arg1};}{that
					\begin{equation}\llabel{eq1}
						y=\E[Y_0^0].
				\end{equation}}
				\argument{\lref{eq1};\lref{assume evidence 2};the assumption that $\P(\E[Y_0^0|X_0^0]=\E[Y_0^0])<1$}{ that
					\begin{equation}\llabel{eq2}
						0<\E[\|\E[Y_0^0|X_0^0]-\E[Y_0^0]\|]=\E[\|\E[y-Y_0^0|X_0^0]\|]=0\dott
				\end{equation}}
				This contradiction show \lref{evidence 2}\dott
				\startnewargseq
                \argument{\lref{assume};the fact that for all $i\in \N$, $x\in \R$ it holds that $\Dens_i(x)>0$;}{that for all $i\in \{1,2,\dots,\ell_1+\ell_0\ell_1\}$, $x\in \R$, $y\in (x,\infty)$ it holds that
                \begin{equation}\llabel{eqc1}
                    \P(\xTheta{0}{i}\in (x,y))>0\dott
                \end{equation}}
                \argument{\lref{assume};}{that \llabel{arggg1} $\xTheta{0}{i}$, $i\in \{1,2,\dots,\ell_1+\ell_0\ell_1\}$, are independent\dott}
				\argument{\cref{main cor: scientific general square};\lref{arggg1};\lref{evidence 1};\lref{evidence 2};\lref{eqc1}}{\cref{conclude: squaremean1}\dott}
			\end{aproof}
            \begin{athm}{cor}{main cor: scientific square mean self-contained}
                Let $L,\scrd\in \N\backslash\{1\}$, $\ell =(\ell_0,\ell_1,\dots,\ell_{L}) \in \N^{L+1}$ satisfy $\scrd=\ffd(\ell)$,  let $ ( \Omega, \mathcal{F}, \P) $ be a probability space, let
				$ a \in \R $, 
				$ b \in [a, \infty)  $, for every $ m, n \in \N_0 $ 
				let 
				$ X^m_n \colon \Omega \to [a,b]^{\ell_0} $
				and 
				$ Y^m_n \colon \Omega \to \R^{\ell_{L}} $
				be random variables, let $S\subseteq \R$ be finite, for every $r\in \N_0$ let $\mathbb A_r\in C^{\min\{r,1\}}(\R,\R)$, let $\grad\colon \R\to\R$ satisfy for all $x\in \R$ that there exists $R\in \N$ such that $ \restr{\mathbb A_0}{\R \backslash S} \in C^1(\R \backslash S,\R)$, $\restr{\grad}{\R \backslash S}=(\restr{\mathbb A_0}{\R \backslash S})'$, and
				\begin{equation}\label{main cor1: eq1}
					\textstyle
					\sum_{r =R}^\infty\bigl(|\mathbb A_r(x)-\mathbb A_0(x)|+|(\mathbb A_r)'(x)-\grad(x)|\bigr)=0,
				\end{equation}
				 let $\alphaw \in \R$, $\betaw \in (\alphaw,\infty)$ satisfy $\sup_{ x \in ( \alphaw, \betaw )\backslash S } | \grad(x)| = 0<\sup_{x\in\R\backslash S}|\grad(x)|$,
				 let $\fL\colon \R^{\scrd}\to \R$ satisfy for all $\theta\in \R^{\scrd}$ that 
				\begin{equation}\label{main cor1: eq2}
					\fL(\theta)=\E\bigl[\|\bfN_{\bbA_0}^{\ell,\theta}(X^0_0)-Y^0_0\|^2\bigr],
				\end{equation}
                assume $\P(\E[Y_0^0|X_0^0]=\E[Y_0^0])<1$, 
				for every $n\in \N_0$ let $ M_n \in  \N $, for every $r,n\in \N_0$ let
				$ 
				\cLnri{n}{r} \colon \R^{ \scrd } \times \Omega \to \R 
				$
				satisfy for all 
				$ \theta \in \R^{ \scrd }$
				that
				\begin{equation}\label{maincor1: eq3}
					\displaystyle
					\cLnri{n}{r}( \theta) 
					= 
					\frac{ 1 }{ M_n } 
					\biggl[ \textstyle
					\sum\limits_{ m = 1 }^{ M_n} 
					\|\bfN_{\bbA_r}^{\ell,\theta}(X_n^m)-Y_n^m\|^2
					\biggr]
					,
				\end{equation}
				for every $n\in \N_0$ let
				$ 
				\cG_n  
				\colon \R^{ \scrd} \times \Omega \to \R^{ \scrd } 
				$ 
				satisfy for all $\omega\in \Omega$, $\theta\in \{\vartheta\in \R^{\scrd}\colon (\nabla_{\vartheta} \cLnri{n}{r}(\vartheta,\omega))_{r\in \N}$ is convergent$\}$
				that
				\begin{equation}\label{main cor1: eq4}
					\cG_n( \theta,\omega) 
					= 
					\lim_{r\to\infty}\bigl[\nabla_\theta \cLnri{n}{r}(\theta,\omega)\bigr]
				\end{equation}
				and let 
				$
				\Theta_n=(\xTheta{n}{1},\dots,\xTheta{n}{\scrd} 
				\colon \Omega  \to \R^{\scrd}
				$
				be a random variable, for every $n\in \N$ let 
				$
				\Phi_n 
				= 
				( 
				\Phi^{ 1 }_n, \dots, 
				\Phi^{\scrd}_n 
				)
				\colon 
				\allowbreak
				( \R^{\scrd} )^{ 2n }
				\allowbreak
				\to 
				\R^{ \scrd}
				$ 
				satisfy 
				for all 
				$
				g =
				( 
				( g_{ i, j } )_{ j \in \{ 1, 2, \dots, \scrd \} }
				)_{
					i \in \{ 1, 2, \dots, 2n\}
				}
				\in 
				(
				\R^{ 
					\scrd
				}
				)^{ 2n }
				$, 
				$ 
				j \in \{1,2,\dots,\scrd\}  
				$
				with $
				\sum_{ i = 1 }^{2n}
				\abs{ g_{ i, j } -g_{1,j}\mathbbm 1_{[1,n]}(i)}
				= 0
				$
				that 
				$
				\Phi^{ j }_n( g ) = g_{n,j} 
				$ and
				assume 
				\begin{equation}
					\label{maincor1: eq5}
					\Theta_{ n  } 
					=  
					\Phi_{n}\bigl(\Theta_0,\Theta_1,\dots,\Theta_{n-1},
					\cG_1( \Theta_0  ) ,
					\cG_2( \Theta_1  ) ,
					\dots ,
					\cG_n ( \Theta_{n-1} )
					\bigr),
				\end{equation} 
				 and let $\sigma,\mu\in \R$ satisfy that $\sigma\Theta_0+\mu$ is standard normal   \cfload.
				Then 
				\begin{equation}\label{main cor1: conclude}
					\textstyle
					\P\Bigl(\inf\limits_{ n\in \N_0 }  \fL(\NNelll_n)>\inf\limits_{\theta\in \R^{\scrd}}\fL(\theta)\Bigr)>0 \ifnocf.
				\end{equation}
                \cfout[.]
            \end{athm}
            \begin{aproof}
                \argument{\cref{main cor: scientific square mean};}{\cref{main cor1: conclude}\dott}
            \end{aproof}
            In \cref{conclude: squaremean1} in \cref{main cor: scientific square mean} and \cref{main cor1: conclude} in \cref{main cor: scientific square mean self-contained} we reveal that the probability that the infimum over the training steps of the true risk of the \SGD\ optimization process $\inf_{ n \in \N_0 } \fL( \Theta_n )$ is strictly larger than the optimal/infimal true risk value $\inf_{ \theta \in \R^{\fd(\ell)} } \fL( \theta )$ is strictly positive. This fact directly implies that it does not hold that the true risk of the \SGD\ optimization process $\fL(\Theta_n)$ converges in probability to the optimal/infimal true risk value $\inf_{ \theta \in \R^{\fd(\ell)} } \fL( \theta )$ as $n$ tends to infinity (cf.\ \cref{NR1} in \cref{subsec: main theorem 1} above). For the sake of completeness we briefly illustrate this implication in the follow elementary and well-known result, \cref{lem: equivalent high probability}.
                         \begin{athm}{lemma}{lem: equivalent high probability}[Non-convergence in probability]
               Let $(\Omega,\mathcal F,\P)$ be a probability space. Then
               \begin{enumerate}[label=(\roman*)]
                   \item \label{item 1: support lemma} it holds for all $\delta\in (0,\infty]$ and all random variables $X_n\colon \Omega\to[0,\infty]$, $n\in \N$, with $\P\bigl(\liminf_{n\to\infty}X_n>0\bigr)>0$ that  \begin{equation}\llabel{conclude1}
                   \liminf_{n\to\infty}\E\bigl[\min\{\delta,X_n\}\bigr]>0.
               \end{equation}
               and
                   \item  \label{item 2: support lemma} it holds for all $\delta\in (0,\infty]$ and all random variables $X_n\colon \Omega \to[0,\infty]$, $n\in \N$, with $\P\bigl(\inf_{n\in \N}X_n>0\bigr)>0$ that
                   \begin{equation}\llabel{conclude 2}
                       \liminf_{n\to\infty}\E\bigl[\min\{\delta,X_n\}\bigr]\geq \inf_{n\in \N}\E\bigl[\min\{\delta,X_n\}\bigr]>0.
                   \end{equation}
               \end{enumerate}
            \end{athm}
            \begin{aproof}
                \argument{the Lebesgue's dominated convergence theorem}
               {that for all stochastic process $X\colon \N\times \Omega\to[0,\infty]$ with $\P\bigl(\liminf_{n\to\infty}X_n>0\bigr)$ it holds that
              \begin{equation}\llabel{eq0}
\textstyle\liminf_{N\to\infty} \P\bigl(\inf_{n\in \N\cap[N,\infty)} X_n>0\bigr)=\P\bigl(\liminf_{n\to\infty} X_n>0\bigr)>0.
\end{equation}}
\argument{\lref{eq0}; the Lebesgue's dominated convergence theorem}
               {that for all stochastic process $X\colon \N\times\Omega\to[0,\infty]$ with $\P\bigl(\liminf_{n\to\infty}X_n>0\bigr)$ it holds that there exist $N_X\in \N$, $\varepsilon_X\in (0,\infty)$ which satisfy
               \begin{equation}\llabel{def: varepsilon}
                  \textstyle \P\bigl(\inf_{n\in \N\cap[N_X,\infty)} X_n>\varepsilon_X\bigr)>0.
               \end{equation}
               and for all stochastic process $X\colon \N\times\Omega\to[0,\infty]$ with $\P\bigl(\inf_{n\in \N}X_n>0\bigr)$ it holds that there exists $\xi_X\in (0,\infty)$ which satisfies
               \begin{equation}\llabel{def: xi}
                   \textstyle \P\bigl(\inf_{n\in \N} X_n>\xi_X\bigr)>0.
               \end{equation}}
               In the following for every $X\colon \N\times\Omega\to[0,\infty]$ with $\P\bigl(\liminf_{n\to\infty}X_n>0\bigr)$ let $A_X\in \mathcal F$ satisfy
                \begin{equation}\llabel{def: A}
                   \textstyle A=\bigl\{\omega\in \Omega\colon \inf_{n\in \N\cap[N_X,\infty)} X_n(\omega)>\varepsilon_X\bigr\}
                \end{equation}
                and for every $X\colon \N\times\Omega\to[0,\infty]$ with $\P\bigl(\inf_{n\in\N}X_n>0\bigr)$ let $B_X\in \mathcal F$ satisfy
                \begin{equation}\llabel{def: B}
                   \textstyle B=\bigl\{\omega\in \Omega\colon \inf_{n\in \N} X_n(\omega)>\xi_X\bigr\}.
                \end{equation}
                \startnewargseq
                 \argument{\lref{def: varepsilon};\lref{def: xi};\lref{def: A};\lref{def: B}}{that for all stochastic process $X\colon \N\times\Omega\to[0,\infty]$ and all stochastic process $Y\colon \N\times\Omega\to[0,\infty]$ with $\P\bigl(\liminf_{n\to\infty}X_n>0\bigr)$ and $\P\bigl(\inf_{n\in \N}Y_n>0\bigr)$ it holds that
                 \begin{equation}\llabel{arg1}
                     \P(A_X)>0 \qqandqq\P(B_Y)>0.
                 \end{equation}}
                \argument{\lref{arg1};\lref{def: A}}{that for all stochastic process $X\colon \N\times\Omega\to[0,\infty]$ with $\P\bigl(\allowbreak\liminf_{n\to\infty}\allowbreak X_n>0\bigr)$ and for all $n\in \N\cap[N_X,\infty)$, $\delta\in (0,\infty]$ it holds that
                \begin{equation}\llabel{eq1}
                    \E\bigl[\min\{\delta,X_n\}\bigr]\geq  \E\bigl[\min\{\delta,X_n\}\mathbbm 1_{A_X}\bigr]\geq \E[\min\{\delta,\varepsilon_X\}\mathbbm 1_{A_X}]= \min\{\delta,\varepsilon_X\}\P(A_X)>0. 
                \end{equation}}
                \argument{\lref{eq1};}[verbs=ep]{\cref{item 1: support lemma}\dott}\startnewargseq
                \argument{\lref{def: B};\lref{arg1}}{that for all stochastic process $X\colon \N\times\Omega\to[0,\infty]$ with $\P\bigl(\inf_{n\in \N}X_n>0\bigr)$ and for all $n\in \N$, $\delta\in (0,\infty]$ it holds that
                \begin{equation}\llabel{eq2}
                     \E\bigl[\min\{\delta,X_n\}\bigr]\geq  \E\bigl[\min\{\delta,X_n\}\mathbbm 1_{B_X}\bigr]\geq \E[\min\{\delta,\xi_X\}\mathbbm 1_{B_X}]= \min\{\delta,\xi_X\}\P(B_X)>0. 
                \end{equation}}
                \argument{\lref{eq2};the fact that for all $(x_n)_{n\in \N}\subseteq\R$ it holds that $\liminf_{n\to\infty}x_n\geq \inf_{n\in \N}x_n$}{\cref{item 2: support lemma}\dott}
            \end{aproof}
            We note that \cref{main theorem1: conclude} in \cref{main theorem1} in the introduction and \cref{item 2: support lemma} in \cref{lem: equivalent high probability} (applied with $( X_n )_{ n \in \N } \curvearrowleft ( | \fL( \Theta_n ) - \inf_{ \theta\in \R^\scrd}  \fL( \theta ) | )_{ n \in \N_0}$ in the notation of \cref{lem: equivalent high probability}) establish \cref{NR1} in the introduction. Moreover, we observe that in the setup of \cref{main cor: scientific square mean} we have that \cref{conclude: squaremean1} implies that there exists $\varepsilon \in (0,\infty)$ such that 
            \begin{equation}\label{eq: NR2}
               \textstyle  \E\biggl[ \Bigl|\inf\limits_{n\in \N_0} \fL(\Theta_n) - \inf
               \limits_{ \theta\in \R^{\fd(\ell)} } \fL(\theta) \Bigr|\biggr] >\varepsilon \qqandqq \P\Bigl(\inf\limits_{n\in \N_0} \fL(\Theta_n) >\varepsilon+ \inf\limits_{ \theta\in \R^{\fd(\ell)} } \fL(\theta)\Bigr)>0.
            \end{equation}
            For completeness we also briefly illustrate this implication in the next elementary and well-known lemma.
             \cfclear
              \begin{athm}{lemma}{conclusion equivalent}
				Let $(\Omega, \mathcal F, \P)$ be a probability space and let $X \colon \Omega \to [0,\infty]$ be a random variable \cfload. Then the following five
				statements are equivalent:
				\begin{enumerate}[label=(\roman*)]
                \item \label{item 3a} It holds that $\P(X>0)>0$ \cfload.
                 \item \label{item 5a} It holds that there exists $\varepsilon\in (0,\infty)$ such that $\P(X>\varepsilon)>0$\cfload.
                 \item \label{item 4a} It holds that $\P(X=0)<1$\cfload.
                 \item \label{item 1a} It holds that $\E[X]>0$ \cfload.
                \item \label{item 2a} It holds for all $\delta\in (0,\infty)$ that $\E[\min\{\delta,X\}]>0$ \cfout.
                \end{enumerate}
			\end{athm}
            \begin{aproof}
            \argument{\cref{lem: equivalent high probability};}{that
                (\ref{item 3a}$\rightarrow$\ref{item 2a})\dott}
                \startnewargseq
                \argument{the fact that for all $\delta\in \R$, $x\in [-\infty,\infty]$ it holds that $\min \{\delta,x\}\leq x$;}{that (\ref{item 2a}$\rightarrow$\ref{item 1a})\dott}
                \startnewargseq
                \argument{the fact that for every random variable $Y\colon \Omega\to [-\infty,\infty]$ with $\P(Y=0)=1$ it holds that $\E[Y]= 0$;}{that (\ref{item 1a}$\rightarrow$\ref{item 4a})\dott}
                \startnewargseq
                \argument{the fact that $\P(X\geq 0)=1$ that; Lebesgue's dominated convergence theorem}{that
                \begin{equation}\llabel{eq1}
                    \liminf_{\varepsilon\searrow 0}\P(X> \varepsilon)=\P(X>0)=1-\P(X=0).
                \end{equation}}
                \argument{\lref{eq1};}{that (\ref{item 4a}$\rightarrow$\ref{item 5a})\dott}
                \startnewargseq
                \argument{the fact that for all $\varepsilon>0$ it holds that $\P(X>\varepsilon)\leq\P(X>0) $}{that (\ref{item 5a}$\rightarrow$\ref{item 3a})\dott}
            \end{aproof}
            Note that \cref{conclusion equivalent} (applied with $X\curvearrowleft (\Omega\ni \omega\mapsto \inf_{ n\in \N_0 }  \fL(\NNelll_n(\omega))\allowbreak-\inf_{\theta\in \R^{\ffd(\ell)}}\fL(\theta)\in [0,\infty])$ in the notation of \cref{conclusion equivalent}) and \cref{conclude: squaremean1} in \cref{main cor: scientific square mean} establish \cref{eq: NR2}.
            \subsubsection{General loss functions}\label{subsec: positive probability other loss}
             \cfclear
         \begin{athm}{cor}{main cor: scientific general loss very deep}
              Assume \cref{setting: scientific learning}, assume $\ell_L=1$,  
                let $\varphi\colon \R^{\ell_0}\to\R$ be measurable, assume that $\varphi(X_0^0)$ and $Y_0^0-\varphi(X_0^0)$ are independent, let $\psi\in C^1([0,\infty),\R)$ be strictly increasing,
				assume for all $x,y\in \R$ that
				\begin{equation}\label{assume1a}
					\bbL_0(x,y)=\psi(|x-y|^2),\quad\P(\varphi(X_0^0)=x)<1,\qandq \E\bigl[|Y_0^0|\psi'((|Y_0^0|+|x|)^2)|\bigr]<\infty,
				\end{equation}
                assume that $(0,\infty)\ni x\mapsto \psi'(x)\sqrt{x}\in \R$ is strictly increasing,
				for every $i\in \N$ let $ \Dens_i \colon \R \to (0,\infty) $ be measurable, 
and assume for all 
$x_1,x_2,\dots,x_{\ffd(\ell)} \in \R$
that
\begin{equation}\llabel{assume}
  \P \bigl( \cap_{i=1}^{\ell_1+\ell_0\ell_1}
    \big\{
   \xTheta{0}{i} 
    < x_i\big\}
  \bigr)
  =
  \prod\limits_{i=1}^{\ell_1+\ell_0\ell_1}\biggl[\int_{ - \infty }^{x_i} \Dens_i(y) \, \d y\biggr] \ifnocf.
\end{equation}
\cfload[.]
Then 
                 \begin{equation}\label{conclude: other loss1}
					\textstyle
					\P\Bigl(\inf\limits_{ n\in \N_0 }  \fL(\NNelll_n)>\inf\limits_{\theta\in \R^{\ffd(\ell)}}\fL(\theta)\Bigr)>0 \ifnocf.
				\end{equation}
                \cfout[.]
         \end{athm}
         \begin{aproof}
         Throughout this proof let $\xi\colon \Omega\to\R$ satisfy
         \begin{equation}\llabel{def: varphi}
					\xi=Y_0^0-\varphi(X
                    _0^0).
				\end{equation}
          \argument{\lref{assume};the fact that for all $i\in \N$, $x\in \R$ it holds that $\Dens_i(x)>0$;}{that for all $i\in \{1,2,\dots,\ell_1+\ell_0\ell_1\}$, $x\in \R$, $y\in (x,\infty)$ it holds that
                \begin{equation}\llabel{eqc1}
                    \P(\xTheta{0}{i}\in (x,y))>0\dott
                \end{equation}}
                \argument{\lref{assume};}{that \llabel{arggg1} $\xTheta{0}{i}$, $i\in \{1,2,\dots,\ell_1+\ell_0\ell_1\}$, are independent\dott}
             \argument{\lref{arggg1};\cref{conditional expectation verify};\cref{scientific 1 dimension ouput general loss};\lref{eqc1}}{\cref{conclude: other loss1}\dott}
         \end{aproof}
            \subsection{Non-convergence with high probability for very deep ANNs}\label{subsec: high probability}
            \subsubsection{Mathematical framework for the training of deep ANNs}
            \cfclear
            \begin{setting}\label{setting: scientific learning very deep}
				For every $k\in \N$ let $l_k,L_k\in \N\backslash\{1\}$, $\xell{k} =(\xxell{0}{k},\xxell{1}{k},\dots,\xxell{L_k}{k}) \in \N\times \{1,2,\dots,l\}^{L_k-1}\times \N$, let $ ( \Omega, \mathcal{F}, \P) $ be a probability space, let
				$ a \in \R $, 
				$ b \in [a, \infty)  $, for every $ k,m, n \in \N_0 $ 
				let 
				$ X^m_{k,n} \colon \Omega \to [a,b]^{\xxell{0}{k}} $
				and 
				$ Y^m_{k,n} \colon \Omega \to \R^{\xxell{L_k}{k}} $
				be random variables, let $S\subseteq \R$ be finite, for every $r\in \N_0$ let $\mathbb A_r\in C^{\min\{r,1\}}(\R,\R)$, let $\grad\colon \R\to\R$ satisfy for all $x\in \R$ that there exists $R\in \N$ such that $ \restr{\mathbb A_0}{\R \backslash S} \in C^1(\R \backslash S,\R)$, $\restr{\grad}{\R \backslash S}=(\restr{\mathbb A_0}{\R \backslash S})'$, and
				\begin{equation}\llabel{def: g_r}
					\textstyle
					\sum_{r =R}^\infty\bigl(|\mathbb A_r(x)-\mathbb A_0(x)|+|(\mathbb A_r)'(x)-\grad(x)|\bigr)=0,
				\end{equation}
				 let  $\betaw\in \R$ satisfy $\sup_{ x \in ( -\infty, \betaw )\backslash S } | \grad(x)| = 0<\sup_{x\in\R\backslash S}|\grad(x)|$, assume $\inf_{x\in \R} \bbA_0(x)>-\infty$,
				for every $k,n\in \N_0$ let $\bbL_{k,n}\in C^{\min\{n,1\}}(\R^{\xxell{L_k}{k}}\times \R^{\xxell{L_k}{k}},\R)$, $M_n^k\in \N$, for every $k\in \N_0$ let $\fL_k\colon \R^{\ffd(\xell{k})}\to \R$ satisfy for all $\theta\in \R^{\ffd(\xell{k})}$ that $\E\bigl[|\bbL_{k,0}(\rel{\xell{k}}{\theta}{\bbA_{0}}(X^0_{k,0}),Y^0_{k,0})|\bigr]\allowbreak<\infty$ and
				\begin{equation}\label{loss: scientific very deep}
			\fL_k(\theta)=\E\bigl[\bbL_{k,0}(\rel{\xell{k}}{\theta}{\bbA_0}(X^0_{k,0}),Y^0_{k,0})\bigr],
				\end{equation}
				 for every $r,k,n\in \N_0$ let
				$ 
				\cL_{k,n}^r \colon \R^{ \ffd(\xell{k}) } \times \Omega \to \R 
				$
				satisfy for all 
				$ \theta \in \R^{ \ffd(\xell{k}) }$
				that
				\begin{equation}\label{objective: scientific very deep}
					\displaystyle
					\cL_{k,n}^r( \theta) 
					= 
					\frac{ 1 }{ M_n^k } 
					\biggl[ \textstyle
					\sum\limits_{ m = 1 }^{ M_n^k} 
					\bbL_{k,n}(\rel{\xell{k}}{\theta}{\bbA_r}(X_{k,n}^m),Y_{k,n}^m)
					\biggr]
					,
				\end{equation}
				for every $k,n\in \N_0$ let 
				$ 
				\cG_n ^k 
				\colon \R^{ \ffd(\xell{k})} \times \Omega \to \R^{ \ffd(\xell{k}) } 
				$ 
				satisfy for all $\omega\in \Omega$, $\theta\in \{\vartheta\in \R^{\ffd(\xell{k})}\colon (\nabla_{\vartheta} 
                \cL_{k,n}^r(\vartheta,\allowbreak\omega))_{r\in \N}$ is convergent$\}$
				that
				\begin{equation}\label{gradient: scientific very deep}
					\cG_n^k( \theta,\omega) 
					= 
					\lim_{r\to\infty}\bigl[\nabla_\theta \cL_{k,n}^r(\theta,\omega)\bigr]
				\end{equation}
				and let 
				$
				\Theta_n^k =(\Theta_{n}^{k,1},\dots,\Theta_{n}^{k,\ffd(\xell{k})})
				\colon \Omega  \to \R^{\ffd(\xell{k}) }
				$
				be a random variable, and for every $k,n\in \N$ let 
				$
				\Phi_n ^k
				= 
				( 
				\Phi^{ k,1 }_n, \dots, 
				\Phi^{ k,\ffd(\xell{k}) }_n 
				)
				\colon 
				\allowbreak
				( \R^{ \ffd(\xell{k})} )^{ 2n }
				\allowbreak
				\to 
				\R^{ \ffd(\xell{k}) }
				$ 
				satisfy 
				for all 
				$
				g =
				( 
				( g_{ i, j } )_{ j \in \{ 1, 2, \dots, \ffd(\xell{k}) \} }
				)_{
					i \in \{ 1, 2, \dots, 2n\}
				}
				\in 
				(
				\R^{ 
					\ffd(\xell{k})
				}
				)^{ 2n }
				$, 
				$ 
				j \in \{1,2,\dots,\ffd(\xell{k})\}  
				$
				with $
				\sum_{ i = 1 }^{2n}
				\abs{ g_{ i, j } -g_{1,j}\mathbbm 1_{[1,n]}(i)}
				= 0
				$
				that 
				$
				\Phi^{ j }_n( g ) = g_{n,j} 
				$ and
				assume 
				\begin{equation}
					\label{SGD: scientific very deep}
					\Theta_{ n  } ^k
					=  
					\Phi_{n}^k\bigl(\Theta_0^k,\Theta_1^k,\dots,\Theta_{n-1}^k,
					\cG_1^k( \Theta_0^k  ) ,
					\cG_2^k( \Theta_1^k  ) ,
					\dots ,
					\cG_n^k ( \Theta_{n-1}^k )
					\bigr) \ifnocf.
				\end{equation}
                \cfload[.]
			\end{setting}
                        \subsubsection{Mean squared error loss function}
            \begin{athm}{lemma}{very deep: estimate probability}
                Assume \cref{setting: scientific learning very deep}, for every $n\in \N$, $k\in\{1,2,\dots,L\}$ let $\bfd_k^n\in \N$ satisfy $\bfd_k^n=\sum_{i=1}^k\xxell{i-1}{n}(\xxell{i-1}{n}+1)$, let $\bfA\in\R$ satisfy $\bfA\leq \inf_{x\in \R} \bbA_0(x)$, let $\gamma\in (-\infty,\min(S\cup\{\betaw\})]$, for every $k\in \N$ let $\rho_{k}\in [-\infty,\infty)$ satisfy
				\begin{equation}\llabel{def: gamma}
					\rho_k=\begin{cases}
						-\infty&\colon \bfA\geq 0\\
						\textstyle\frac{1}{\bfA l_k} &\colon \bfA<0, L>2\\
                        0 &\colon \bfA<0,L=2,
                        \end{cases}
				\end{equation}
                let $ \Dens \colon \R \to (0,\infty) $ be measurable, let $(\cki_{k,i})_{(k,i)\in \N^2}\allowbreak\subseteq(0,\infty)$ 
satisfy for all $k\in \N
$, $\chi\in \{1,2,\dots,\allowbreak\bfd_{L_k-1}^k\}$,
$x_1,x_2,\dots,x_{\chi} \in \R$
that
  \begin{equation}\llabel{assume}
      \P \bigl( \cap_{i=1}^{\chi}
    \big\{
    \Theta_{0}^{k,i} 
    < x_i\big\}
  \bigr)
  =
  \prod_{i=1}^{\chi}\biggl[\int_{ - \infty }^{\cki_{k,i}x_i} \Dens(y) \, \d y\biggr],
   \end{equation} 
    let $\bfc\in \R$ satisfy $\bfc=\limsup_{k\to\infty} \max_{i\in \{1,2,\dots,\ffd(\xell{k})\}} ((\cki_{k,i})^{-1}+\cki_{k,i})$, let $\fp\in[0,\infty]$, $p,\varepsilon\in (0,\infty)$ satisfy $\fp= \inf_{x\in [-\varepsilon,0]} \Dens(x)$ and
    \begin{equation}\llabel{def: p}
   p<\min\biggl\{\fp\varepsilon\mathbbm 1_{(-\infty,0)}(\bfA)+\mathbbm 1_{[0,\infty)}(\bfA)\int_{-\infty}^{0}\Dens(y)\,\d y,\int_{-\infty}^{\min\{0,\bfc(\gamma-\mathbbm 1_{(-\infty,0)}(\bfA))\}}\Dens(y)\,\d y\biggr\},
    \end{equation}
 and assume $\liminf_{k\to\infty}\bigl(L_k\bigl(\frac{p\varepsilon}{\varepsilon- l_k\bfc\min\{\bfA,0\}}\bigr)^{l_k(l_k+1)}\bigr)=\infty$
    \cfload. Then
                \begin{multline}\llabel{conclude}
\textstyle\textstyle\limsup\limits_{n\to\infty}\biggl(\prod\limits_{k=2}^{L_n-1}\biggl[1\\ 
\textstyle-\biggl[\prod\limits_{j=1}^{\xxell{k}{n}\xxell{k-1}{n}}\P(\rho_k<\xTheta{0}{n,j+\bfd_{k-1}^n}<0)\biggr]
\textstyle\biggl[\prod\limits_{j=1}^{\xxell{k}{n}}\P\bigl(\xTheta{0}{n,j+\xxell{k}{n}\xxell{k-1}{n}+\bfd_{k-1}^n}<\gamma-\mathbbm 1_{(-\infty,0)}(\bfA)\bigr)\biggr]\biggr]\biggr)=0\ifnocf.
\end{multline}
            \end{athm}
            \begin{aproof}
               \argument{\lref{assume}; \cite[item (ii) in Lemma 2.5]{HannibalJentzenThang2024};}{that for all $n\in \N$, $i\in\{1,2,\dots,\ffd(\xell{n})-\xxell{L_n}{n}(\xxell{L_n-1}{n}+1)\}$, $x\in \R$ it holds that
               \begin{equation}\llabel{eq1}
                   \P(\Theta_{0}^{n,i}<x)=\int_{-\infty}^{\cki_{n,i}x}\Dens(y)\,\d y.
               \end{equation}} 
               \argument{the assumption that $\liminf_{k\to\infty}\bigl(L_k\bigl(\frac{p\varepsilon}{\varepsilon-l_k\bfc\min\{\bfA,0\}}\bigr)^{l_k(l_k+1)}\bigr)=\infty$}{that \llabel{arg1} $\liminf_{n\to\infty}L_n=\infty$\dott}
               \argument{\lref{arg1};\lref{def: gamma}}{that there exists $N\in \N$ such that for all $k\in \N\cap[N,\infty)$ it holds that
               \begin{equation}\llabel{eq2}
                   \rho_k<0.
               \end{equation}}
               \argument{\lref{eq1};\lref{eq2}; the assumption that $\bfc=\limsup_{k\to\infty} \max_{i\in \{1,2,\dots,\ffd(\xell{k})\}} ((\cki_{k,i})^{-1}+\cki_{k,i})$}{that for all $i\in\{1,2,\dots,\ffd(\xell{n})-\xxell{L_n}{n}(\xxell{L_n-1}{n}+1)\}$ it holds that
               \begin{equation}\llabel{eq3}
                   \liminf_{n\to\infty}\min_{i\in\{1,2,\dots,\bfd_{L_n-1}^n\}}\biggl(\frac{\P(\rho_n<\xTheta{0}{n,i}<0)}{\int_{\frac{\rho_n}{\bfc}}^{0}\Dens(y)\,\d y}\biggr)\geq 1
                   \end{equation}
                   and
                   \begin{equation}\llabel{eq4}
                   \liminf_{n\to\infty}\min_{i\in\{1,2,\dots,\bfd_{L_n-1}^n\}}\biggl(\frac{\P(\xTheta{0}{n,i}<\gamma-1)}{\int_{-\infty}^{\max\{0,\bfc(\gamma-\mathbbm 1_{(-\infty,0)}(\bfA))\}}\Dens(y)\,\d y}\biggr)\geq 1.
               \end{equation}}
               In the proof of \lref{conclude} we distinguish between the case $\bfA\geq 0$ and the case $\bfA<0$. We first prove \lref{conclude} in the case
               \begin{equation}\llabel{case 1}
                   \bfA\geq 0.
               \end{equation}
               \startnewargseq
               \argument{\lref{def: gamma};\lref{def: p}; \lref{eq3};\lref{eq4};\lref{case 1}}{that 
               \begin{equation}\llabel{case1: eq1'}
                   \liminf_{n\to\infty}\min_{i\in\{1,2,\dots,\bfd_{L_n-1}^n\}}\P(\rho_n<\xTheta{0}{n,i}<0)\geq p
                   \end{equation}
                   and
                   \begin{equation}\llabel{case1: eq1}
                  \liminf_{n\to\infty}\min_{i\in\{1,2,\dots,\bfd_{L_n-1}^n\}}\P(\xTheta{0}{n,i}<\gamma)\geq p\dott
               \end{equation}}
               \argument{\lref{case1: eq1};the fact that $\liminf_{k\to\infty}\bigl(L_k p^{l_k(l_k+1)}\bigr)=\infty$}{that
               \begin{equation}\llabel{case1: eq2}
                   \begin{split}
                       &\textstyle
\limsup\limits_{n\to\infty}\biggl(\prod\limits_{k=2}^{L_n-1}\biggl[1-\biggl[\prod\limits_{j=1}^{\xxell{k}{n}\xxell{k-1}{n}}\P(\rho_k<\xTheta{0}{n,j+\bfd_{k-1}^n}<0)\biggr]\biggl[\prod\limits_{j=1}^{\xxell{k}{n}}\P(\xTheta{0}{n,j+\xxell{k}{n}\xxell{k-1}{n}+\bfd_{k-1}^n}<\gamma)\biggr]\biggr]\biggr)\\
&\textstyle\leq \limsup_{n\to\infty}\biggl(\prod_{k=2}^{L_n-1}\bigl[1-p^{\xxell{n}{k}(\xxell{n}{k}+1)}\bigr]\biggr)\leq \textstyle\limsup_{n\to\infty}[1-p^{l_n(l_n+1)}]^{L_n-2}=0.
                   \end{split}
               \end{equation}}
               \argument{\lref{case1: eq2};}{\lref{conclude} in the case $\bfA\geq 0$\dott}
               We now prove \lref{conclude} in the case
               \begin{equation}\llabel{case 2}
                   \bfA<0.
               \end{equation}
\startnewargseq
\argument{\lref{def: gamma};\lref{def: p};\lref{case 2}}{that for all $n\in \N$ with $L_n>2$ and $
|\bfA|l_n \bfc>\varepsilon$ it holds that 
\begin{equation}\llabel{case2: eqp1}
     \int_{\frac{\rho_n}{c}}^0\Dens(y)\,\d y\geq \int_{-\varepsilon}^0 \Dens(y)\,\d y\geq \fp\varepsilon\geq p
\end{equation}
and for all $n\in \N$ with  $L_n>2$ and
$|\bfA|l_n \bfc\leq \varepsilon$ it holds that
\begin{equation}\llabel{case2: eq1}
    \int_{\frac{\rho_n}{c}}^0\Dens(y)\,\d y\geq \frac{\varepsilon}{|\bfA l_n\bfc|}.
\end{equation}}
\argument{\lref{case2: eq1};the fact that $0<p<1$}{that for all $n\in \N$ with $L_n>2$ it holds that
\begin{equation}\llabel{case2: eq1t}
    \int_{\frac{\rho_n}{c}}^0\Dens(y)\,\d y\geq\frac{p}{1-\frac{\bfA l_n\bfc}{\varepsilon}}.
\end{equation}}
\argument{\lref{def: p};\lref{eq3};\lref{eq4};\lref{case2: eq1t};the fact that $\liminf_{n\to\infty}L_n=\infty$}{that
\begin{equation}\llabel{case2: eq2}
                   \liminf_{n\to\infty}\min_{i\in\{1,2,\dots,\bfd_{L_n-1}^n\}}\biggl(\frac{\P(\rho_n<\xTheta{0}{n,i}<0)}{\frac{p}{1-\frac{\bfA l_n\bfc}{\varepsilon}}}\biggr)\geq 1
                   \end{equation}
                   and
                   \begin{equation}\llabel{case2: eq3}
                   \liminf_{n\to\infty}\min_{i\in\{1,2,\dots,\bfd_{L_n-1}^n\}}\biggl(\frac{\P(\xTheta{0}{n,i}<\gamma-1)}{\frac{p}{1-\frac{\bfA l_n\bfc}{\varepsilon}}}\biggr)\geq 1.
               \end{equation}}
                \argument{\lref{case2: eq3};the fact that $\liminf_{k\to\infty}\bigl(L_k\bigl(\frac{p}{1-\frac{\bfA l_k\bfc}{\varepsilon}}\bigr)^{l_k(l_k+1)}\bigr)=\infty$}{that
               \begin{equation}\llabel{case2: eq4}
                   \begin{split}
                       &\textstyle
\limsup\limits_{n\to\infty}\biggl(\prod\limits_{k=2}^{L_n-1}\biggl[1-\biggl[\prod\limits_{j=1}^{\xxell{k}{n}\xxell{k-1}{n}}\P(\rho_k<\xTheta{0}{n,j+\bfd_{k-1}^n}<0)\biggr]\biggl[\prod\limits_{j=1}^{\xxell{k}{n}}\P(\xTheta{0}{n,j+\xxell{k}{n}\xxell{k-1}{n}+\bfd_{k-1}^n}<\gamma-1)\biggr]\biggr]\biggr)\\
&\textstyle\leq \limsup_{n\to\infty}\biggl(\prod_{k=2}^{L_n-1}\bigl[1-\bigl(\frac{p}{1-\frac{\bfA l_n\bfc}{\varepsilon}}\bigr)^{\xxell{n}{k}(\xxell{n}{k}+1)}\bigr]\biggr)\\
&\textstyle\leq \textstyle\limsup_{n\to\infty}\bigl[1-\bigl(\frac{p}{1-\frac{\bfA l_n\bfc}{\varepsilon}}\bigr)^{l_n(l_n+1)}\bigr]^{L_n-2}=0.
\end{split}
\end{equation}}
\argument{\lref{case2: eq4};}{\lref{conclude} in the case $\bfA<0$\dott}
            \end{aproof}
             \cfclear
            \begin{athm}{cor}{main theo: scientific square mean error very deep}
                 Assume \cref{setting: scientific learning very deep}, let $\bfA\in\R$ satisfy $\bfA\leq \inf_{x\in \R} \bbA_0(x)$, let $\gamma\in (-\infty,\min(S\allowbreak\cup\{\betaw\})]$, assume for all $k\in \N$, $x,y \in \R^{\xxell{L_k}{k}}$ that $\bbL_{k,0}(x,y)=\|x-y\|^2$, assume for all $k\in \N$ that $\P(\E[Y_{k,0}^0|X_{k,0}^0]=\E[Y_{k,0}^0])<1$, (cf.\ \cref{Lemma confirm}), let $ \Dens \colon \R \to (0,\infty) $ be measurable, let $(\cki_{k,i})_{(k,i)\in \N^2}\allowbreak\subseteq(0,\infty)$ 
satisfy for all $k\in \N
$, $\chi\in \N\cap[1,\ffd(\xell{k})-\xxell{L_k}{k}\xxell{L_k-1}{k}-\xxell{L_k}{k}]$,
$x_1,x_2,\dots,x_{\chi} \in \R$
that
  \begin{equation}\llabel{assume}
      \P \bigl( \cap_{i=1}^{\chi}
    \big\{
    \Theta_{0}^{k,i} 
    < x_i\big\}
  \bigr)
  =
  \prod_{i=1}^{\chi}\biggl[\int_{ - \infty }^{\cki_{k,i}x_i} \Dens(y) \, \d y\biggr],
   \end{equation} 
   let $\bfc\in \R$ satisfy $\bfc=\limsup_{k\to\infty} \max_{i\in \{1,2,\dots,\ffd(\xell{k})\}} ((\cki_{k,i})^{-1}+\cki_{k,i})$, let $\fp\in[0,\infty]$, $p,\varepsilon\in (0,\infty)$ satisfy $\fp= \inf_{x\in [-\varepsilon,0]} \Dens(x)$ and
    \begin{equation}\llabel{def: p}
   p<\min\biggl\{\fp\varepsilon\mathbbm 1_{(-\infty,0)}(\bfA)+\mathbbm 1_{[0,\infty)}(\bfA)\int_{-\infty}^{0}\Dens(y)\,\d y,\int_{-\infty}^{\min\{0,\bfc(\gamma-\mathbbm 1_{(-\infty,0)}(\bfA))\}}\Dens(y)\,\d y\biggr\},
    \end{equation}
 and assume $\liminf_{k\to\infty}\bigl(L_k\bigl(\frac{p\varepsilon}{\varepsilon-l_k\bfc\min\{\bfA,0\}}\bigr)^{l_k(l_k+1)}\bigr)=\infty$
    \cfload. Then
                \begin{equation}\label{conclude: mean square loss2}
                    \textstyle
\liminf\limits_{k\to\infty}\P\Bigl(\inf\limits_{ n\in \N_0 }  \fL_k(\NNelll_n^k)>\inf\limits_{\theta\in \R^{\ffd(\xell{k})}}\fL_k(\theta)\Bigr)=1 \ifnocf.
                \end{equation}
                \cfout[.]
            \end{athm}
            \begin{aproof}
                \argument{\cite[item (iii) in Lemma 2.5]{HannibalJentzenThang2024};\lref{assume}}{that for all $k\in\N$ it holds that \llabel{arggg1} $\Theta_{0}^{k,i}$, $i\in \{1,2,\dots,\ffd(\xell{k})-\xxell{L_k}{k}(\xxell{L_k-1}{k}+1)\}$, are independent\dott}
                \argument{\lref{arggg1};\cref{main cor: scientific general square};\cref{very deep: estimate probability}}{\cref{conclude: mean square loss2}\dott}
            \end{aproof}
            \begin{athm}{cor}{main theo: scientific square mean error very deep self-contained}
            Let $d,\delta\in \N$, for every $k\in \N_0$ let $l_k,L_k\in \N\backslash\{1\}$, $\xell{k} =(\xxell{0}{k},\xxell{1}{k},\dots,\xxell{L_k}{k}) \in \{d\}\times \{1,2,\dots,l_k\}^{L_k-1}\allowbreak\times \{\delta\}$, let $ ( \Omega, \mathcal{F}, \P) $ be a probability space, let
				$ a \in \R $, 
				$ b \in [a, \infty)  $, for every $ m, n \in \N_0 $ 
				let 
				$ X^m_{n} \colon \Omega \to [a,b]^{d} $
				and 
				$ Y^m_{n} \colon \Omega \to \R^{\delta} $
				be random variables, let $S\subseteq \R$ be finite, for every $r\in \N_0$ let $\mathbb A_r\in C^{\min\{r,1\}}(\R,\R)$, let $\grad\colon \R\to\R$ satisfy for all $x\in \R$ that there exists $R\in \N$ such that $ \restr{\mathbb A_0}{\R \backslash S} \in C^1(\R \backslash S,\R)$, $\restr{\grad}{\R \backslash S}=(\restr{\mathbb A_0}{\R \backslash S})'$, and
				\begin{equation}\label{main cor2: eq1}
					\textstyle
					\sum_{r =R}^\infty\bigl(|\mathbb A_r(x)-\mathbb A_0(x)|+|(\mathbb A_r)'(x)-\grad(x)|\bigr)=0,
				\end{equation}
				 let  $\betaw\in \R$ satisfy $\sup_{ x \in ( -\infty, \betaw )\backslash S } | \grad(x)| = 0<\sup_{x\in\R\backslash S}|\grad(x)|$, 
				for every $k\in \N_0$ let $\fL_k\colon \R^{\ffd(\xell{k})}\to \R$ satisfy for all $\theta\in \R^{\ffd(\xell{k})}$ that 
				\begin{equation}\label{main cor2: eq2}
			\fL_k(\theta)=\E\bigl[\|\rel{\xell{k}}{\theta}{\bbA_0}(X^0_{0})-Y^0_{0}\|^2\bigr],
				\end{equation}
                assume $\P(\E[Y_{0}^0|X_{0}^0]=\E[Y_{0}^0])<1$,
				for every $k,n\in \N_0$ let $ M_n ^k\in  \N $, for every $r,k,n\in \N_0$ let
				$ 
				\cL_{k,n}^r \colon \R^{ \ffd(\xell{k}) } \times \Omega \to \R 
				$
				satisfy for all 
				$ \theta \in \R^{ \ffd(\xell{k}) }$
				that
				\begin{equation}\label{main cor2: eq3}
					\displaystyle
					\cL_{k,n}^r( \theta) 
					= 
					\frac{ 1 }{ M_n^k } 
					\biggl[ \textstyle
					\sum\limits_{ m = 1 }^{ M_n^k} 
					\|\rel{\xell{k}}{\theta}{\bbA_r}(X_{n}^m)-Y_{n}^m\|^2
					\biggr]
					,
				\end{equation}
				for every $k,n\in \N_0$  let
				$ 
				\cG_n ^k 
				\colon \R^{ \ffd(\xell{k})} \times \Omega \to \R^{ \ffd(\xell{k}) } 
				$ 
				satisfy for all $\omega\in \Omega$, $\theta\in \{\vartheta\in \R^{\ffd(\xell{k})}\colon (\nabla_{\vartheta} 
                \cL_{k,n}^r(\vartheta,\omega)\allowbreak)_{r\in \N}$ is convergent$\}$
				that
				\begin{equation}\label{main cor2: eq4}
					\cG_n^k( \theta,\omega) 
					= 
					\lim_{r\to\infty}\bigl[\nabla_\theta \cL_{k,n}^r(\theta,\omega)\bigr]
				\end{equation}
				and let 
				$
				\Theta_n^k 
				\colon \Omega  \to \R^{\ffd(\xell{k}) }
				$
				be a random variable, for every $k,n\in \N$ let 
				$
				\Phi_n ^k
				= 
				( 
				\Phi^{ k,1 }_n, \dots, 
				\Phi^{ k,\ffd(\xell{k}) }_n 
				)
				\colon 
				\allowbreak
				( \R^{ \ffd(\xell{k})} )^{ 2n }
				\allowbreak
				\to 
				\R^{ \ffd(\xell{k}) }
				$ 
				satisfy 
				for all 
				$
				g =
				( 
				( g_{ i, j } )_{ j \in \{ 1, 2, \dots, \ffd(\xell{k}) \} }
				)_{
					i \in \{ 1, 2, \dots, 2n\}
				}
				\in 
				(
				\R^{ 
					\ffd(\xell{k})
				}
				)^{ 2n }
				$, 
				$ 
				j \in \{1,2,\dots,\allowbreak\ffd(\xell{k})\}  
				$
				with $
				\sum_{ i = 1 }^{2n}
				\abs{ g_{ i, j } -g_{1,j}\mathbbm 1_{[1,n]}(i)}
				= 0
				$
				that 
				$
				\Phi^{ j }_n( g ) = g_{n,j} 
				$ and
				assume 
				\begin{equation}
					\label{main cor2: eq5}
					\Theta_{ n  } ^k
					=  
					\Phi_{n}^k\bigl(\Theta_0^k,\Theta_1^k,\dots,\Theta_{n-1}^k,
					\cG_1^k( \Theta_0^k  ) ,
					\cG_2^k( \Theta_1^k  ) ,
					\dots ,
					\cG_n^k ( \Theta_{n-1}^k )
					\bigr),
				\end{equation}
				 let $\sigma,\mu\in \R$ satisfy for all $k\in \N$ that $\sigma\Theta_0^k+\mu$ is standard normal, 
assume $\inf_{x\in \R}\bbA_0(x)\geq 0$, and assume $\limsup_{p\searrow 0}\liminf_{k\to\infty}\bigl(\allowbreak p^{l_k(l_k+1)}\allowbreak L_k\bigr)=\infty$ \cfload. Then
                \begin{equation}\label{main cor2: conclude}
                    \textstyle
\liminf\limits_{k\to\infty}\P\Bigl(\inf\limits_{ n\in \N_0 }  \fL_k(\NNelll_n^k)>\inf\limits_{\theta\in \R^{\ffd(\xell{k})}}\fL_k(\theta)\Bigr)=1\ifnocf.
                \end{equation}
                \cfout[.]
                \end{athm}
                \begin{aproof}
                    \argument{\cref{main theo: scientific square mean error very deep};}{\cref{main cor2: conclude}\dott}
                \end{aproof}
            \subsubsection{General loss functions}
             \cfclear
             \begin{athm}{cor}{main theo: scientific other error very deep}
                 Assume \cref{setting: scientific learning very deep}, let $\bfA\in\R$ satisfy $\bfA\leq \inf_{x\in \R} \bbA_0(x)$, let $\gamma\in (-\infty,\min(S\allowbreak\cup\{\betaw\})]$, let $\psi\in C^1([0,\infty),\R)$ be strictly increasing, assume for all $k\in \N$, $x,y \in \R$ that $\xxell{L_k}{k}=1$ and
				\begin{equation}\label{def: bbLaa}
					\bbL_{k,0}(x,y)=\psi(|x-y|^2),
				\end{equation}
                assume that $(0,\infty)\ni x\mapsto \psi'(x)\sqrt{x}\in \R$ is strictly increasing,
               for every $k\in\N$ let $\varphi_k\colon \R^{\xxell{0}{k}}\to\R$ be measurable, assume for all $k\in \N$ that $\varphi_k(X_{k,0}^0)$ and $Y_{k,0}^0-\varphi_k(X_{k,0}^0)$ are independent,
				let $\varepsilon\in (0,\infty)$ satisfy for all $k\in \N$, $x\in \R$ that
				\begin{equation}\label{assume1aa}
					\P(\varphi_k(X_{k,0}^0)=x)<1\qqandqq \E\bigl[|Y_{k,0}^0|\psi'((|Y_{k,0}^0|+|x|)^2)|\bigr]<\infty,
				\end{equation} 
                let $ \Dens \colon \R \to (0,\infty) $ be measurable, let $(\cki_{k,i})_{(k,i)\in \N^2}\allowbreak\subseteq(0,\infty)$ 
satisfy for all $k\in \N
$, $\chi\in \N\cap[1,\ffd(\xell{k})-\xxell{L_k}{k}\xxell{L_k-1}{k}-\xxell{L_k}{k}]$,
$x_1,x_2,\dots,x_{\chi} \in \R$
that
  \begin{equation}\label{assumeaa}
      \P \bigl( \cap_{i=1}^{\chi}
    \big\{
    \Theta_{0}^{k,i} 
    < x_i\big\}
  \bigr)
  =
  \prod_{i=1}^{\chi}\biggl[\int_{ - \infty }^{\cki_{k,i}x_i} \Dens(y) \, \d y\biggr],
   \end{equation} 
    let $\bfc\in \R$ satisfy $\bfc=\limsup_{k\to\infty} \max_{i\in \{1,2,\dots,\ffd(\xell{k})\}} ((\cki_{k,i})^{-1}+\cki_{k,i})$, let $\fp\in[0,\infty]$, $p,\varepsilon\in (0,\infty)$ satisfy $\fp= \inf_{x\in [-\varepsilon,0]} \Dens(x)$ and
    \begin{equation}\llabel{def: p}
   p<\min\biggl\{\fp\varepsilon\mathbbm 1_{(-\infty,0)}(\bfA)+\mathbbm 1_{[0,\infty)}(\bfA)\int_{-\infty}^{0}\Dens(y)\,\d y,\int_{-\infty}^{\min\{0,\bfc(\gamma-\mathbbm 1_{(-\infty,0)}(\bfA))\}}\Dens(y)\,\d y\biggr\},
    \end{equation}
 and assume $\liminf_{k\to\infty}\bigl(L_k\bigl(\frac{p\varepsilon}{\varepsilon-l_k\bfc\min\{\bfA,0\}}\bigr)^{l_k(l_k+1)}\bigr)=\infty$
    \cfload. Then
                \begin{equation}\label{concludeaaa}
                    \textstyle
\liminf\limits_{k\to\infty}\P\Bigl(\inf\limits_{ n\in \N_0 }  \fL_k(\NNelll_n)>\inf\limits_{\theta\in \R^{\ffd(\xell{k})}}\fL_k(\theta)\Bigr)=1 \ifnocf.
                \end{equation}
                \cfout[.]
            \end{athm}
            \begin{aproof}
                \argument{\cite[item (iii) in Lemma 2.5]{HannibalJentzenThang2024};\cref{assumeaa}}{that for all $k\in\N$ it holds that \llabel{arggg1} $\Theta_{0}^{k,i}$, $i\in \{1,2,\dots,\ffd(\xell{k})-\xxell{L_k}{k}(\xxell{L_k-1}{k}+1)\}$, are independent\dott}
                \argument{\lref{arggg1};\cref{scientific 1 dimension ouput general loss};\cref{very deep: estimate probability}}{\cref{concludeaaa}\dott}
            \end{aproof}
            \subsection*{Acknowledgements}
This work has been partially funded by the National Science Foundation of China (NSFC) under grant number 12250610192. Moreover, we gratefully acknowledge the Cluster of Excellence EXC 2044-390685587, Mathematics Münster: Dynamics-Geometry-Structure funded by the Deutsche Forschungsgemeinschaft (DFG, German Research Foundation). Benno Kuckuck is gratefully acknowledged for several useful comments.

			\bibliographystyle{acm}
			\bibliography{bibfile}
		\end{document}